\documentclass{article}

\usepackage[preprint]{neurips_2026}

\usepackage[utf8]{inputenc} % allow utf-8 input
\usepackage[T1]{fontenc}    % use 8-bit T1 fonts
\usepackage{hyperref}       % hyperlinks
\usepackage{url}            % simple URL typesetting
\usepackage{booktabs}       % professional-quality tables
\usepackage{amsfonts}       % blackboard math symbols
\usepackage{nicefrac}       % compact symbols for 1/2, etc.
\usepackage{microtype}      % microtypography
\usepackage{xcolor}         % colors

\usepackage{graphicx}
\usepackage{latexsym}
\usepackage{caption}
\usepackage[font=small,labelfont=bf]{subcaption}
\usepackage{booktabs}
\usepackage{enumitem}
\usepackage{multirow}
\usepackage{color}
\usepackage{hyperref}
\usepackage{amsmath}
\usepackage{amsfonts}
\usepackage{amsthm}
\usepackage{scalefnt}
\usepackage{xspace}
\usepackage{fontawesome}
\usepackage{tcolorbox}
\usepackage{rotating}
\usepackage{subcaption}
\usepackage{dashrule}
\usepackage{afterpage}
\usepackage{makecell}
\usepackage{mathtools}
\usepackage{url}
\usepackage[utf8]{inputenc}
\usepackage[T1]{fontenc}
\usepackage{tipa}
\usepackage{tcolorbox}
\usepackage{wrapfig}
\usepackage{amssymb}
\usepackage{tablefootnote}
\usepackage{threeparttable}
\usepackage{textcomp}

\usepackage{CJKutf8}
\newcommand{\cn}[1]{\begin{CJK*}{UTF8}{gbsn}#1\end{CJK*}}

\usepackage{tikz}
\usepackage{pgfplots}
\usepgfplotslibrary{groupplots,polar}
\pgfplotsset{compat=1.18}
\usetikzlibrary{arrows.meta,calc,positioning,fit,backgrounds,plotmarks,patterns}

\usepackage{tabularx}
\newcolumntype{C}{>{\centering\arraybackslash}X}

\usepackage{cleveref}
\crefname{section}{\S\!}{\S\S\!}
\crefname{table}{Tab.}{Tabs.}
\crefname{figure}{Fig.}{Figs.}
\crefname{algorithm}{Alg.}{Algs.}
\crefname{appendix}{App.}{Apps.}
\crefname{lemma}{Lemma}{}
\Crefname{theorem}{Theorem}{}
\crefname{proposition}{Proposition}{}
\crefname{hypothesis}{Hypothesis}{}
\crefname{deduction}{Deduction}{}
\crefname{intuition}{\textbf{Intuition}}{\textbf{Intuitions}}
\crefname{observation}{\textbf{Observation}}{\textbf{Observations}}
\crefname{finding}{\textbf{Finding}}{\textbf{Findings}}
\crefname{cor}{Corollary}{Corollaries}
\newcommand{\appref}[1]{\hyperref[#1]{App.~\ref*{#1}}}
\newcommand{\Appref}[1]{\hyperref[#1]{App.~\ref*{#1}}}
\newcommand{\apprefs}[2]{Apps.~\hyperref[#1]{\ref*{#1}} and \hyperref[#2]{\ref*{#2}}}

\crefname{equation}{Eq.}{Eqs.} % 你的原有定义
\creflabelformat{equation}{#2#1#3} % 去掉单个公式引用的小括号 (例如: Eq. 1)
\crefrangelabelformat{equation}{#3#1#4--#5#2#6} % (可选) 如果你会用到范围引用 \cref{eq1,eq2,eq3}，建议顺便把范围引用的格式也改掉
\newcommand{\Mspec}{\pi_{\theta_1}}
\newcommand{\Mzero}{\pi_{\theta_0}}
\newcommand{\Mdist}{\pi_{\theta'_0}}

\newcommand{\IMspec}{\pi_{\phi_1}}
\newcommand{\IMzero}{\pi_{\phi_0}}

\usepackage[table]{xcolor}
\definecolor{tablerowcolor}{gray}{0.9}

\newcommand{\alternaterowcolors}[1][2]{
    \def\startrow{#1}  % The starting row is set by the optional parameter (default is 2)

    \rowcolors{\startrow}{tablerowcolor}{white}
}

\definecolor{ForestGreen}{HTML}{009B55}
\definecolor{OrangeRed}{HTML}{ED135A}
\definecolor{CadetBlue}{HTML}{74729A}
\definecolor{SkyBlue}{HTML}{46C5DD}
\definecolor{SeaGreen}{HTML}{3FBC9D}
\definecolor{Peach}{HTML}{F7965A}
\definecolor{Goldenrod}{HTML}{FFDF42}
\definecolor{NavyBlue}{HTML}{006EB8}
\definecolor{Periwinkle}{HTML}{7977B8}
\definecolor{Orchid}{HTML}{AF72B0}
\definecolor{BlueViolet}{HTML}{473992}
\definecolor{SpringGreen}{HTML}{C6DC67}
\definecolor{ETHGray}{RGB}{0, 94, 184}	% gray

\newcommand{\ID}{\textsc{It}\xspace}
\newcommand{\SID}{\textsc{Id}\xspace}
\newcommand{\OOD}{\textsc{Ood}\xspace}
\newcommand{\chem}{\textsc{Chem}\xspace}
\newcommand{\phys}{\textsc{Phys}\xspace}
\newcommand{\MT}{\textsc{Lrm}\xspace}

\newcommand{\SMol}{\textsf{SMol}\xspace}

\newcommand{\qa}{(q,a)}

\newcommand{\dto}[2]{#1$\mathrel{\overset{\scriptscriptstyle d}{\to}}$#2}

\DeclareMathOperator{\FFT}{FFT}
\DeclareMathOperator{\LST}{LST}
\DeclareMathOperator{\ASFT}{ASFT}
\DeclareMathOperator{\LORA}{LoRA}

\DeclareMathOperator{\KL}{KL}
\DeclareMathOperator{\HR}{HR@1}
\newcommand{\given}{\,|\,}
\newcommand{\klgiven}{\,||\,}

\usepackage{collcell,fp}
\usepackage{xcolor,colortbl}
\usepackage{pgfmath}

\NewDocumentCommand{\increase}{m m o}{%
    \edef\ColorP{\fpeval{round(min(100, max(20, 20 + 1.8 * abs(#2))))}}%
    \ensuremath{%
        {#1}_{{\color{ForestGreen!\ColorP} \scriptscriptstyle #2\uparrow}}%
        \IfValueT{#3}{^{\boldsymbol{#3}}}%
    }%
}

\NewDocumentCommand{\decrease}{m m o}{%
    \edef\ColorP{\fpeval{round(min(100, max(20, 20 + 1.8 * abs(#2))))}}%
    \ensuremath{%
        {#1}_{{\color{OrangeRed!\ColorP} \scriptscriptstyle #2\downarrow}}%
        \IfValueT{#3}{^{\boldsymbol{#3}}}%
    }%
}

\usepackage{todonotes}
\usepackage{multirow}
\usepackage{array}
\usepackage{tcolorbox}
\definecolor{tticblue}{RGB}{0, 94, 184}  % a more visually pleasing blue

\definecolor{ETHBlue}{RGB}{33,92,175}	% blue 
\definecolor{ETHGreen}{RGB}{98,115,19}		% green
\definecolor{ETHPurpleDark}{RGB}{140,10,89}	% purple
\definecolor{ETHPurple}{RGB}{163,7,116}	% purple
\definecolor{ETHGray}{RGB}{111,111,111}	% gray
\definecolor{ETHRed}{RGB}{183,53,45}	% red
\definecolor{ETHPetrol}{RGB}{0,120,148}	% green/blue 
\definecolor{ETHBronze}{RGB}{142,103,19}	% bronze
\definecolor{softred}{RGB}{255,155,155}
\definecolor{softblue}{RGB}{100, 149, 237}
\definecolor{softorange}{RGB}{255, 183, 77}
\definecolor{softgreen}{RGB}{111, 194, 118}
\definecolor{wistia}{RGB}{244, 183, 65}

\colorlet{originblue}{ETHBlue!10}
\colorlet{originpetrol}{ETHPetrol!10}
\colorlet{baselineorange}{ETHPurple!5}
\colorlet{originbronze}{ETHBronze!10}

\usepackage{xspace}
\newcommand{\modelfontstyle}{\textsf}
\newcommand{\modellogo}[3][0.8em]{%
    \raisebox{-0.1em}{%
        \includegraphics[height=#1]{#2}%
    }%
    \,%
    {\modelfontstyle{#3}}\xspace%
}
\newcommand{\xqwen}[1][0.9em]{%
    \modellogo[#1]{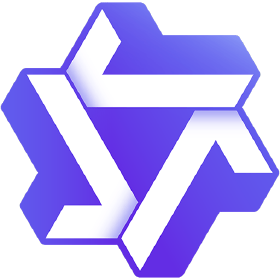}{3-8B}%
}
\newcommand{\xintern}[1][0.9em]{%
    \modellogo[#1]{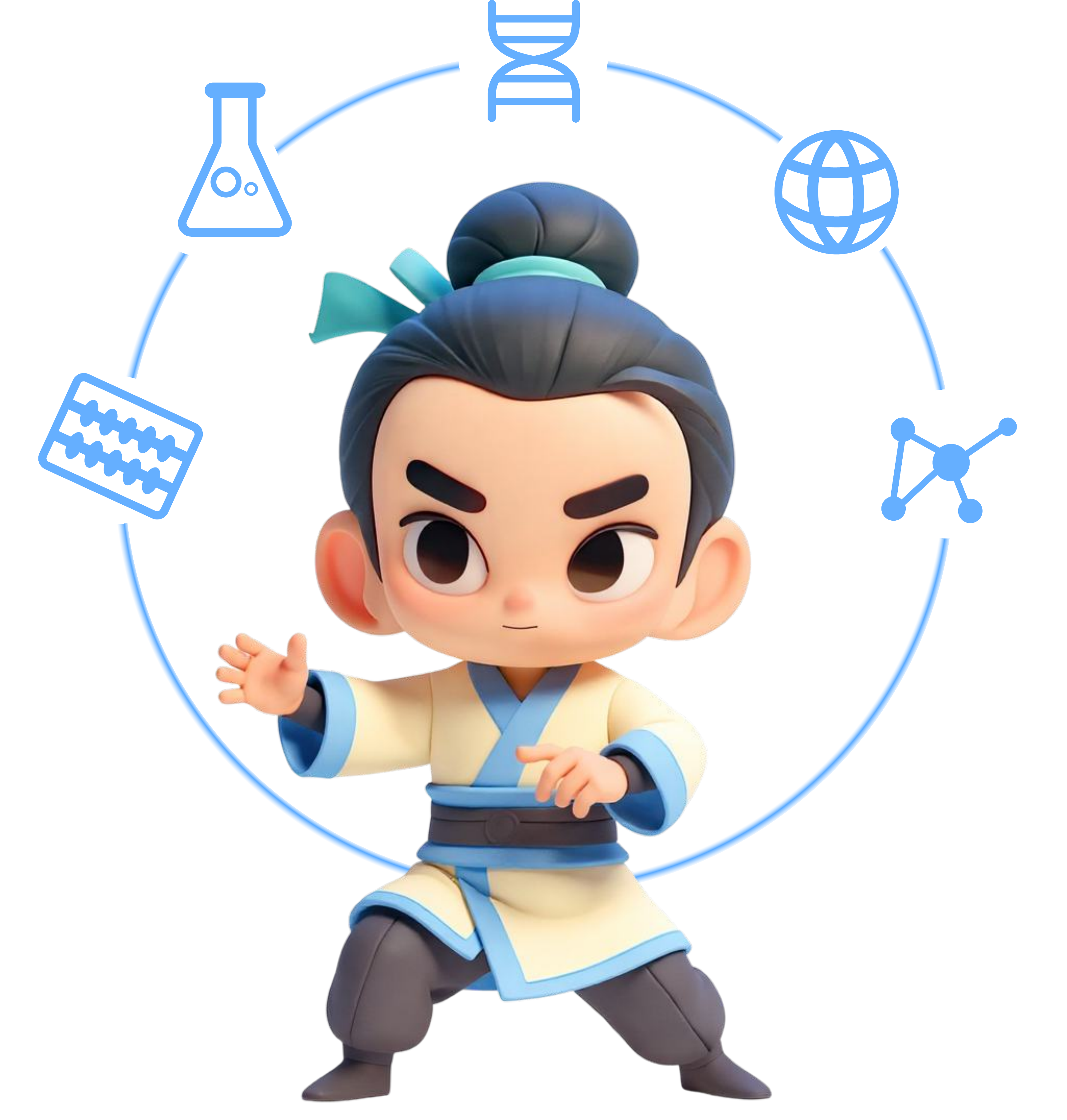}{-S1m}%
}

\newcommand{\subarrow}{%
  \tikz[baseline=-0.6ex]{
    \draw[->, line width=0.5pt, cap=round, join=round] (0, 1.2ex) |- (1.2em, 0);
  }\,
}

\definecolor{blueA}{HTML}{0649B8}
\definecolor{blueLight}{HTML}{EEF5FF}
\definecolor{greenA}{HTML}{008A32}
\definecolor{greenLight}{HTML}{EFF8E9}
\definecolor{purpleA}{HTML}{6A2DB8}
\definecolor{purpleLight}{HTML}{F5EEFF}
\definecolor{panelFill}{HTML}{FCFCFE}
\definecolor{panelLine}{HTML}{D8D8E0}

\newcommand{\numcircle}[4]{%
  \node[circle, fill=#3, inner sep=0pt, minimum size=0.72cm,
        text=white, font=\sffamily\bfseries\large] at (#1,#2) {#4};
}

\newcommand{\dbicon}[4]{%
  \begin{scope}[shift={(#1,#2)}]
    \path[draw=#3!45, fill=#3!13, line width=.45pt]
      (-.62,.32) -- (-.62,-.32) arc (180:360:.62 and .17) -- (.62,.32);
    \path[draw=#3!45, fill=#3!8, line width=.45pt] (0,.32) ellipse (.62 and .17);
    \draw[#3!45, line width=.45pt] (-.62,.32) -- (-.62,-.32);
    \draw[#3!45, line width=.45pt] (.62,.32) -- (.62,-.32);
    \draw[#3!45, line width=.45pt] (-.62,-.32) arc (180:360:.62 and .17);
    \node[font=\sffamily\small] at (0,-0.15) {#4};
  \end{scope}
}

\newcommand{\legenddb}[2]{%
  \begin{scope}[shift={(#1,#2)}, scale=.75]
    \path[draw=black!75, fill=white, line width=.55pt]
      (-.5,.35) -- (-.5,-.35) arc (180:360:.5 and .16) -- (.5,.35);
    \path[draw=black!75, fill=white, line width=.55pt] (0,.35) ellipse (.5 and .16);
    \draw[black!75, line width=.55pt] (-.5,.35) -- (-.5,-.35);
    \draw[black!75, line width=.55pt] (.5,.35) -- (.5,-.35);
  \end{scope}
}

\newcommand{\distOne}[5]{%
\begin{scope}[shift={(#1,#2)}, xscale=#4, yscale=#5]
  \path[fill=#3!16]
    (0,0) .. controls (.25,.03) and (.38,.28) .. (.55,.72)
    .. controls (.68,1.10) and (.88,1.38) .. (1.05,1.39)
    .. controls (1.28,1.39) and (1.30,.70) .. (1.46,.35)
    .. controls (1.68,.03) and (2.15,.07) .. (2.58,.13)
    .. controls (2.95,.18) and (3.20,.08) .. (3.55,0) -- cycle;
  \draw[#3, line width=.82pt]
    (0,0) .. controls (.25,.03) and (.38,.28) .. (.55,.72)
    .. controls (.68,1.10) and (.88,1.38) .. (1.05,1.39)
    .. controls (1.28,1.39) and (1.30,.70) .. (1.46,.35)
    .. controls (1.68,.03) and (2.15,.07) .. (2.58,.13)
    .. controls (2.95,.18) and (3.20,.08) .. (3.55,0);
  \draw[#3!75!black, line width=.55pt] (0,0) -- (3.55,0);
\end{scope}}

\newcommand{\distTwo}[5]{%
\begin{scope}[shift={(#1,#2)}, xscale=#4, yscale=#5]
  \path[fill=#3!16]
    (0,0) .. controls (.25,.04) and (.38,.24) .. (.58,.68)
    .. controls (.76,1.00) and (1.02,1.08) .. (1.22,.94)
    .. controls (1.48,.72) and (1.38,.18) .. (1.78,.14)
    .. controls (2.05,.11) and (2.13,.56) .. (2.45,.72)
    .. controls (2.78,.88) and (3.00,.32) .. (3.55,0) -- cycle;
  \draw[#3, line width=.82pt]
    (0,0) .. controls (.25,.04) and (.38,.24) .. (.58,.68)
    .. controls (.76,1.00) and (1.02,1.08) .. (1.22,.94)
    .. controls (1.48,.72) and (1.38,.18) .. (1.78,.14)
    .. controls (2.05,.11) and (2.13,.56) .. (2.45,.72)
    .. controls (2.78,.88) and (3.00,.32) .. (3.55,0);
  \draw[#3!65, dashed, line width=.65pt]
    (0,0) .. controls (.25,.03) and (.38,.28) .. (.55,.72)
    .. controls (.68,1.10) and (.88,1.38) .. (1.05,1.39)
    .. controls (1.28,1.39) and (1.30,.70) .. (1.46,.35)
    .. controls (1.68,.03) and (2.15,.07) .. (2.58,.13)
    .. controls (2.95,.18) and (3.20,.08) .. (3.55,0);
  \draw[#3!75!black, line width=.55pt] (0,0) -- (3.55,0);
\end{scope}}

\newcommand{\distRight}[5]{%
\begin{scope}[shift={(#1,#2)}, xscale=#4, yscale=#5]
  \path[fill=#3!16]
    (0,0) .. controls (.30,.04) and (.45,.35) .. (.78,.55)
    .. controls (1.08,.72) and (1.25,.18) .. (1.55,.15)
    .. controls (1.93,.12) and (2.08,.80) .. (2.48,1.08)
    .. controls (2.88,1.38) and (3.10,.28) .. (3.55,0) -- cycle;
  \draw[#3, line width=.82pt]
    (0,0) .. controls (.30,.04) and (.45,.35) .. (.78,.55)
    .. controls (1.08,.72) and (1.25,.18) .. (1.55,.15)
    .. controls (1.93,.12) and (2.08,.80) .. (2.48,1.08)
    .. controls (2.88,1.38) and (3.10,.28) .. (3.55,0);
  \draw[#3!65, dashed, line width=.65pt]
    (0,0) .. controls (.25,.03) and (.38,.28) .. (.55,.72)
    .. controls (.68,1.10) and (.88,1.38) .. (1.05,1.39)
    .. controls (1.28,1.39) and (1.30,.70) .. (1.46,.35)
    .. controls (1.68,.03) and (2.15,.07) .. (2.58,.13)
    .. controls (2.95,.18) and (3.20,.08) .. (3.55,0);
  \draw[#3!75!black, line width=.55pt] (0,0) -- (3.55,0);
\end{scope}}

\title{Training Specialist Models without Reasoning Trajectories for Domain Expert Distillation}
\author{
Yilei Tu\textsuperscript{1}\thanks{Work done during an internship at Shanghai Artificial Intelligence Laboratory.} \quad
Zihao Li\textsuperscript{3} \quad
Shaoxiong Ji\textsuperscript{4,5} \quad
Jörg Tiedemann\textsuperscript{3} \quad
Fei Yuan\textsuperscript{2}
\\[3pt]
\textsuperscript{1}University of British Columbia \qquad
\textsuperscript{2}Shanghai Artificial Intelligence Laboratory
\\
\textsuperscript{3}University of Helsinki \qquad
\textsuperscript{4}ELLIS Institute Finland \qquad
\textsuperscript{5}University of Turku
\\[3pt]
\texttt{yileitu@cs.ubc.ca} \quad
\texttt{zihao.li@helsinki.fi} \quad
\texttt{shaoxiong.ji@utu.fi}
\\
\texttt{jorg.tiedemann@helsinki.fi} \quad
\texttt{feiyvan@163.com}
}

\begin{document}

\maketitle

\setcounter{footnote}{0}

\begin{abstract}
Specialist distillation effectively transfers domain expertise to student models via teacher-generated reasoning trajectories. However, when these specialists are trained solely on question--answer pairs without explicit reasoning supervision, what governs the trajectories they generate? In this work, we show that specialist optimization implicitly selects from this latent trajectory space. To isolate and observe this latent distribution, we leverage student distillation not as a downstream goal, but as an agnostic probe---since students inherit no parameterization or optimization constraints from the specialist, inheriting only the sampled trajectories themselves. Through this probe, our empirical analysis unveils a tight governing relationship: across 27 specialist--student pairings, their specialization--generalization profiles correlate exceptionally strongly. Crucially, explicitly controlling the specialist's distributional drift systematically shifts both the teacher and its distilled student along a controllable trade-off between domain precision and general-capability retention. Across chemistry, physics, and multilingual settings, distilled students systematically reflect these specialist-induced profiles, even across divergent model families. Our findings establish a new view of specialist training: when gold reasoning is absent, tuning choices directly control the latent supervision passed to downstream models. The code\footnote{\url{https://github.com/CONE-MT/DCO/tree/main/specialist_distillation}}, models and datasets\footnote{\url{https://huggingface.co/collections/yileitu/qaonly-specialist-distillation}} are publicly available.
\end{abstract}

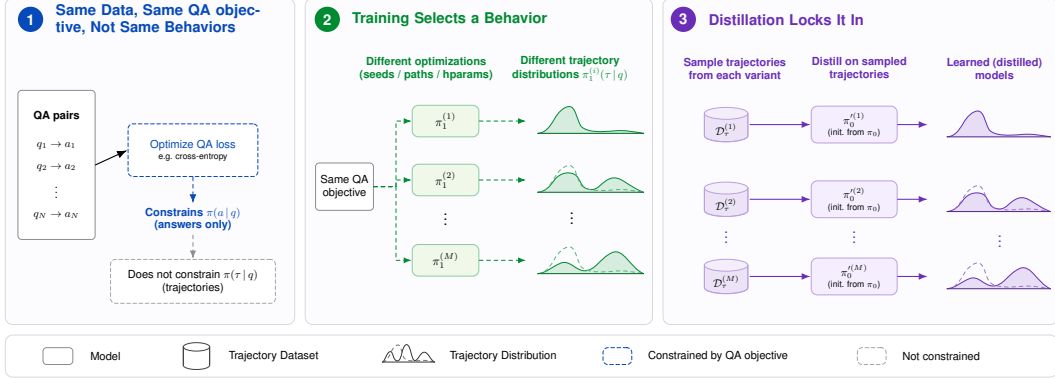
\begin{figure*}[!htp]
\centering
% \vspace{-15pt}

\resizebox{\textwidth}{!}{%
\begin{tikzpicture}[
  font=\sffamily,
  >=Latex,
  line cap=round,
  line join=round,
  panel/.style={draw=panelLine, fill=panelFill, rounded corners=6pt, line width=.5pt},
  box/.style={draw=black!45, fill=white, rounded corners=3pt, line width=.5pt, align=center},
  dashedblue/.style={draw=blueA, dashed, dash pattern=on 3pt off 2pt, rounded corners=4pt, line width=.8pt, fill=white, align=center},
  dashedgray/.style={draw=black!35, dashed, dash pattern=on 3pt off 2pt, rounded corners=4pt, line width=.7pt, fill=white, align=center},
  modelgreen/.style={draw=greenA!45, fill=greenLight, rounded corners=5pt, line width=.7pt, minimum width=1.9cm, minimum height=.9cm, align=center},
  modelpurple/.style={draw=purpleA!45, fill=purpleLight, rounded corners=5pt, line width=.7pt, minimum width=2.45cm, minimum height=1.05cm, align=center}
]
% Top row panels
\draw[panel] (0,10.25) rectangle (8.25,19.55);
\draw[panel] (8.55,10.25) rectangle (18.45,19.55);
\draw[panel] (18.75,10.25) rectangle (30,19.55);

% Panel 1
\numcircle{.72}{18.95}{blueA}{1}
\node[anchor=west, text=blueA, font=\sffamily\bfseries\large, align=left, text width=6.3cm]
  at (1.32,18.95) {Same Data, Same QA objective, Not Same Behaviors};
\node[box, minimum width=2.2cm, minimum height=4.25cm, font=\sffamily\small]
  (qa) at (1.46,14.8)
  {\textbf{QA pairs}\\[.45cm]
   $q_1 \to a_1$\\[.27cm]
   $q_2 \to a_2$\\[.2cm]
   $\vdots$\\[.2cm]
   $q_N \to a_N$};
\node[dashedblue, minimum width=3.7cm, minimum height=1.5cm, font=\sffamily\small]
  (opt) at (5.35,15.22) {\textcolor{blueA}{Optimize QA loss}\\[-1pt]
  {\scriptsize e.g. cross-entropy}};
\draw[-{Latex[length=3mm]}, line width=.9pt] (qa.east) -- (opt.west);
\node[text=blueA, font=\sffamily\bfseries\small, align=center]
  (ans) at (5.35,13.25) {Constrains $\pi(a\given q)$\\(answers only)};
\draw[-{Latex[length=2.6mm]}, blueA, dashed, line width=.8pt] (opt.south) -- (ans.north);
\node[dashedgray, minimum width=4.7cm, minimum height=1.25cm, font=\sffamily\small]
  (notraj) at (5.35,11.48) {Does not constrain $\pi(\tau\given q)$\\(trajectories)};
\draw[-{Latex[length=2.4mm]}, black!45, dashed, line width=.8pt] (ans.south) -- (notraj.north);

% Panel 2
\numcircle{9.18}{18.95}{greenA}{2}
\node[anchor=west, text=greenA, font=\sffamily\bfseries\large, align=left, text width=8.5cm]
  at (9.78,18.95) {Training Selects a Behavior};
\node[text=greenA, font=\sffamily\bfseries\small, align=center]
  at (12.0,17.55) {Different optimizations\\(seeds / paths / hparams)};
\node[text=greenA, font=\sffamily\bfseries\small, align=center]
  at (16.1,17.55) {Different trajectory\\distributions $\pi_1^{(i)}(\tau\given q)$};
\node[box, minimum width=1.65cm, minimum height=1.35cm, font=\sffamily\small]
  (sameqa) at (9.67,14.18) {Same QA\\objective};
\node[modelgreen] (m21) at (12.55,16.05) {$\pi_1^{(1)}$};
\node[modelgreen] (m22) at (12.55,14.37) {$\pi_1^{(2)}$};
\node[font=\sffamily\large] at (12.55,13.35) {$\vdots$};
\node[modelgreen] (m2m) at (12.55,12.08) {$\pi_1^{(M)}$};
\foreach \n in {m21,m22,m2m}{
  \draw[-{Latex[length=2.3mm]}, greenA, dashed, line width=.8pt]
    (sameqa.east) -| ($ (\n.west) + (-.45,0) $) -- (\n.west);
}
\distOne{15.18}{15.70}{greenA}{.85}{.55}
\distTwo{15.18}{14.02}{greenA}{.85}{.55}
\distRight{15.18}{11.70}{greenA}{.85}{.55}
\foreach \n/\yy in {m21/16.05,m22/14.37,m2m/12.08}{
  \draw[-{Latex[length=2.3mm]}, greenA, dashed, line width=.8pt] (\n.east) -- (14.85,\yy);
}
\node[font=\sffamily\large] at (16.15,13.38) {$\vdots$};

% Panel 3
\numcircle{19.33}{18.95}{purpleA}{3}
\node[anchor=west, text=purpleA, font=\sffamily\bfseries\large, align=left, text width=9.8cm]
  at (19.93,18.95) {Distillation Locks It In};
\node[text=purpleA, font=\sffamily\bfseries\small, align=center] at (20.8,17.55)
  {Sample trajectories\\from each variant};
\node[text=purpleA, font=\sffamily\bfseries\small, align=center] at (24.35,17.55)
  {Distill on sampled\\trajectories};
\node[text=purpleA, font=\sffamily\bfseries\small, align=center] at (28.2,17.55)
  {Learned (distilled)\\models};
\dbicon{20.55}{15.95}{purpleA}{$\mathcal{D}_\tau^{(1)}$}
\dbicon{20.55}{13.82}{purpleA}{$\mathcal{D}_\tau^{(2)}$}
\node[font=\sffamily\large, text=purpleA] at (20.55,12.85) {$\vdots$};
\dbicon{20.55}{11.62}{purpleA}{$\mathcal{D}_\tau^{(M)}$}
\node[modelpurple, font=\sffamily\small] (p31) at (24.2,15.95)
  {$\pi_0^{\prime(1)}$\\[-1pt]{\scriptsize (init. from $\pi_0$)}};
\node[modelpurple, font=\sffamily\small] (p32) at (24.2,13.82)
  {$\pi_0^{\prime(2)}$\\[-1pt]{\scriptsize (init. from $\pi_0$)}};
\node[font=\sffamily\large, text=purpleA] at (24.2,12.85) {$\vdots$};
\node[modelpurple, font=\sffamily\small] (p3m) at (24.2,11.62)
  {$\pi_0^{\prime(M)}$\\[-1pt]{\scriptsize (init. from $\pi_0$)}};
\draw[-{Latex[length=2.6mm]}, purpleA, line width=.85pt] (21.23,15.95) -- (p31.west);
\draw[-{Latex[length=2.6mm]}, purpleA, line width=.85pt] (21.23,13.82) -- (p32.west);
\draw[-{Latex[length=2.6mm]}, purpleA, line width=.85pt] (21.23,11.62) -- (p3m.west);
\foreach \model/\y in {p31/15.95,p32/13.82,p3m/11.62}{
  \draw[-{Latex[length=2.6mm]}, purpleA, line width=.85pt] (\model.east) -- (26.30,\y);
}
\distOne{26.85}{15.60}{purpleA}{.83}{.53}
\distTwo{26.85}{13.47}{purpleA}{.83}{.53}
\distRight{26.85}{11.27}{purpleA}{.83}{.53}
\node[font=\sffamily\large, text=purpleA] at (28.35,12.72) {$\vdots$};

% Legend
\draw[draw=panelLine, fill=white, rounded corners=6pt, line width=.5pt] (0,8.75) rectangle (30,9.95);
\node[box, minimum width=.85cm, minimum height=.48cm] at (1.5,9.35) {};
\node[anchor=west, font=\sffamily\small] at (2.3,9.35) {Model};
\legenddb{5.45}{9.35}
\node[anchor=west, font=\sffamily\small] at (6.25,9.35) {Trajectory Dataset};
\begin{scope}[shift={(10.75,9.18)}, xscale=.42, yscale=.35]
  % solid distribution
  \draw[black!80, line width=.7pt]
    (0,0) .. controls (.18,.04) and (.28,.85) .. (.55,.85)
    .. controls (.75,.85) and (.70,.05) .. (1.05,.10)
    .. controls (1.35,.15) and (1.30,1.40) .. (1.65,1.42)
    .. controls (1.95,1.38) and (1.95,.15) .. (2.20,.12)
    .. controls (2.42,.09) and (2.55,.80) .. (2.85,.80)
    .. controls (3.14,.78) and (3.25,.14) .. (3.55,0);
  % dashed reference distribution
  \draw[black!55, dashed, line width=.65pt]
    (0,0) .. controls (.25,.03) and (.38,.28) .. (.55,.72)
    .. controls (.68,1.10) and (.88,1.38) .. (1.05,1.39)
    .. controls (1.28,1.39) and (1.30,.70) .. (1.46,.35)
    .. controls (1.68,.03) and (2.15,.07) .. (2.58,.13)
    .. controls (2.95,.18) and (3.20,.08) .. (3.55,0);
  % baseline
  \draw[black!80, line width=.6pt] (0,0) -- (3.55,0);
\end{scope}
\node[anchor=west, font=\sffamily\small] at (12.55,9.35) {Trajectory Distribution};
\node[dashedblue, minimum width=.82cm, minimum height=.48cm] at (17.45,9.35) {};
\node[anchor=west, font=\sffamily\small] at (18.2,9.35) {Constrained by QA objective};
\node[dashedgray, minimum width=.82cm, minimum height=.48cm] at (24.7,9.35) {};
\node[anchor=west, font=\sffamily\small] at (25.45,9.35) {Not constrained};
\end{tikzpicture}
}
% \caption{
% \textbf{QA-only specialist distillation is underdetermined at the trajectory level.}
% \textbf{(1)} Supervising only question--answer $(q, a)$ pairs constrains the answer marginal $\pi(a \given q)$, but does not identify a unique trajectory distribution $\pi(\tau \given q)$ over latent reasoning paths $\tau$.
% \textbf{(2)} Under the same QA objective, different optimization outcomes (e.g., different optimization paths, or parameterizations) can induce different intermediate specialists $\pi_{\theta_1}^{(i)}$ with distinct trajectory distributions, even though they are all equally compatible with the observed gold answers.
% \textbf{(3)} Once trajectories are sampled from a chosen intermediate model and used for distillation, the final distilled model inherits that variant-specific trajectory distribution.
% % The figure highlights the central claim of the paper: QA-only supervision leaves trajectory learning fundamentally ambiguous, and specialist distillation commits to one implicitly selected solution.
% We highlight the central claim of the paper: QA-only supervision leaves trajectory learning fundamentally underdetermined. Distinct specialist--distillation strategy commits to distinct implicitly selected trajectory distribution.
% }

\caption{
\textbf{QA-only specialist distillation is underdetermined at the trajectory level.}
\textbf{(1)} QA supervision over QA $(q, a)$ pairs constrains the answer distribution $\pi(a \given q)$, but not a unique trajectory distribution $\pi(\tau \given q)$ over latent reasoning paths $\tau$.
\textbf{(2)} Under the same QA objective, different optimization profiles can produce intermediate specialists $\pi_{\theta_1}^{(i)}$ with distinct trajectory distributions, even though they are all equally compatible with the observed gold answers.
\textbf{(3)} Distilling from trajectories sampled from one such specialist causes the final model to inherit that variant-specific trajectory distribution.
Overall, QA-only supervision leaves trajectory learning underdetermined, and specialist distillation implicitly commits to one selected behavior.
}
\label{fig:pull_fig_intro}
\vspace{-15pt}
\end{figure*}
\section{Introduction}\label{sec:intro}

Specialist distillation \citep{hinton2015distilling,fu2023specializing,hsieh2023distilling,llm_reasoning_teachers,teaching_reason,adapt_and_distill,liu2024ddk} transfers domain expertise through an intermediate specialist model. A general-purpose \textit{origin} model~$\pi_{\theta}$, e.g., Qwen3-8B-Instruct \citep{qwen3}, is first adapted to a target domain, and the resulting \textit{specialist} then generates reasoning trajectories that serve as supervision for a downstream student~\citep{orca,wizardlm}. Yet in most specialized domains, the specialist itself is never explicitly taught how to reason. Domain datasets typically provide $(q,a)$ question--answer (QA) pairs~\citep{uesato2022solving,chan2022unirex,lightman2023let,turpin2023language} but no gold trajectories~\citep{star,rest,beyond_human_data}, because expert reasoning is difficult to obtain and verify. Since the specialist is optimized only against final answers, what governs the trajectories it generates?

\pagebreak

Let $\tau$ denote a free-running reasoning trajectory generated for question $q$. We consider representative behaviors including a substantive, answer-consistent path \citep[$\tau_{+}$;][]{faithful_reasoning,turpin2023language}, a shortcut \citep[$\tau_{\sim}$;][]{shortcut_learning}, and an empty trajectory ($\tau_{\emptyset}$). Standard teacher-forced QA training directly optimizes answer likelihood and does not explicitly supervise these trajectories. At generation time, however, the trained model induces a distribution over them. Conceptually,
\begin{equation}
\log \pi_\theta(a \given q)
= \log \sum_{\tau \rightarrow a} \pi_\theta(\tau \given q),
\label{eq:intro_qa_loss}
\end{equation}
where $\tau\rightarrow a$ denotes a trajectory compatible with answer $a$. \Cref{eq:intro_qa_loss} characterizes generation-time behavior rather than the implemented SFT objective. Since answer-level supervision provides no direct preference among $\{\tau_{+},\tau_{\sim},\tau_{\emptyset},\cdots\}$, multiple trajectory distributions may remain compatible with the same supervised answer.

To isolate and observe this latent selection, we repurpose student distillation as an agnostic probe rather than treating it only as a downstream goal. A student inherits neither the specialist's parameters nor the optimization constraints used to obtain them; it receives only the specialist-generated supervision. In our controlled pipeline, students share the same initialization and training configuration and are trained on equal amounts of sampled and filtered data. Differences among students thus expose what these reasoning trajectories actually carry, even though students never receive the specialist's parameters or optimization constraints directly.

Through this diagnostic probe, we uncover a tight link between specialists and their distilled students. Across chemistry, physics, and multilingual benchmarks, the balance between domain specialization and general capability in specialists systematically transfers to their downstream students. Across nine specialist-student pairs covering 27 distinct evaluations, these performance profiles align remarkably well (Spearman $\rho = 0.9573$, permutation test $p-\mathrm{value} = 0.0093$, see \cref{sec:exp:results:mono_corr} for details). This pattern holds even when teacher and student models belong to completely different model families, showing that shared teacher--student parameterization is not necessary for the observed transfer. Distilled students therefore expose how the latent supervision selected by specialist optimization shapes downstream capability profiles.

Crucially, this selection process is both observable and controllable. Limiting how far a specialist drifts from its original base model directly recalibrates its trade-off between domain mastery and general capabilities, guiding the student model along with it. Layer-Selective Tuning~\citep[LST;][]{llamax2} provides an implicit anchor, yielding lower behavioral Kullback-Leibler (KL) divergence and better preserving reasoning structures than standard full fine-tuning. For explicit control, Anchored Supervised Fine-Tuning~\citep[ASFT;][]{asft} provides a complementary explicit intervention: varying its anchoring strength systematically moves both specialist and student along the same trade-off. Together, these results identify distributional drift as a controllable axis of the latent supervision passed downstream.

Our main contributions are:
\begin{itemize}[leftmargin=10pt, topsep=-5pt, itemsep=-1pt]
\item \textbf{Specialist optimization is the key design variable for distillation data.} Under QA-only training, reasoning trajectories remain underdetermined by answer labels; the specialist's optimization procedure selects the trajectory distribution from which downstream supervision is generated.

\item \textbf{Distilled students reveal the supervision selected by their specialists.}
Using student distillation as an agnostic probe, we uncover a strong correspondence between specialist and student specialization--generalization profiles across domains and model families. This inheritance shows that the effects of specialist optimization are encoded in the generated trajectories and transferred downstream, rather than remaining confined to the specialist's parameters.

\item \textbf{We systematically characterize and control the resulting specialization--generalization trade-off.}
Through controlled distillation experiments, trajectory-quality analysis, and comparisons between unconstrained tuning, implicit drift control ($\LST$), and explicit KL anchoring ($\ASFT$), we identify distributional drift as a governing axis of latent supervision. Varying this drift steers both specialists and their students between domain precision and general-capability retention.

\end{itemize}

\section{Related Works}

\paragraph{Specialist Distillation and Domain Adaptation.}
Adapting general-purpose models to specialized domains is challenging due to cost, latency, and data scarcity, motivating specialist distillation and domain adaptation.
Early work compared ``distill-then-adapt'' with ``adapt-then-distill'', showing that adapting both the teacher and student to the target domain before task-agnostic distillation can yield compact models that preserve domain expertise \citep{adapt_and_distill}.
Recent LLM pipelines, including \textsf{DeepSeek-V3.2} \citep{dsv32} and \textsf{Qwen3.5-Omni} \citep{qwen35}, similarly train domain-specific experts and distill their capabilities back into a generalist model; \citet{li2024meteor} further propose a staged expert-growth framework from external supervision toward autonomous improvement.
Other work focuses on constructing and exploiting high-quality domain supervision.
Synthetic query generation has been used to distill lightweight retrieval rerankers \citep{udapdr}, while knowledge hierarchies guide literature data distillation for biomedical QA \citep{KAILIN}.
On the distillation process, \citet{xia2026reasoning} use contrastive self-distillation to transfer LLM reasoning paths into BERT without requiring explicit reasoning at inference, and \citet{liu2024ddk} adapt the composition of distillation data to teacher--student performance gaps across domains.
Unlike these studies, we focus on trajectory-level ambiguity in QA-only specialist distillation, where specialist optimization implicitly shapes the trajectory distribution inherited by the final model.

\paragraph{Self-Distillation in LLMs.}
Self-distillation uses a model's own outputs or internal distributions as supervision.
\citet{self_distillation_bridges} rewrite original responses into the model's own distribution before fine-tuning, mitigating catastrophic forgetting while preserving alignment.
Similarly, \citet{shenfeld2026self} construct a demonstration-conditioned teacher from the same model and distill its predictions via on-policy reverse KL for continual skill acquisition without reward engineering.
For complex reasoning, \citet{zhao2026self} use the same model as a privileged teacher and student, providing dense per-token supervision over the student's own rollouts.
In multilingual settings, \citet{zhang-etal-2024-enhancing-multilingual} distill resource-rich language responses to improve cross-lingual capabilities while preserving source-language performance.
For code generation, \citet{ssd} simply fine-tune on sampled solutions, showing that even minimal self-distillation can improve performance through reshaping the model's output distribution.
Our work complements this literature by studying how QA-only specialist optimization determines the latent reasoning supervision passed downstream.
\section{QA-only Specialist Distillation as Trajectory-Distribution Selection}\label{sec:method}

Specialist distillation typically refers to a two-phase training pipeline. Starting from an origin model $\Mzero$, one first obtains an intermediate model $\Mspec$ that is adapted to a target domain. This intermediate model is then used to generate training data for the final model $\Mdist$, often initialized from the same origin model $\Mzero$. In this section, we study specialist distillation from a trajectory learning perspective (in \cref{sec:reformulation}). Surprisingly, we find that properly controlling specialist optimization can induce high-quality reasoning trajectories under QA-only supervision (in \cref{sec:traj_learnable}).

\subsection{From QA-only supervision to trajectory learning in specialist distillation.}
\label{sec:reformulation}

% \paragraph{QA-only supervision does not uniquely determine the distribution of plausible reasoning trajectories.}
% \label{sec:amplification}
% In many domain-specific tasks, datasets $\mathcal{D}$ contain only question--answer pairs $\qa$ without reasoning trajectories $\tau$. Let $\pi_{\theta_1}(\tau \given q)$ denote the induced distribution over reasoning trajectories under model $\pi_{\theta_1}$. Under QA-only supervision, only the final answer is observed, while reasoning trajectories remain latent. Consequently, optimization constrains only the aggregate probability mass over answer-consistent trajectories, rather than how probability mass is distributed among them. Formally, training maximizes the marginal likelihood of the correct answer: 

\paragraph{QA-only supervision underdetermines the induced trajectory distribution.} \label{sec:amplification}
In many domain-specific tasks, datasets $\mathcal{D}$ contain only question--answer pairs $\qa$ without reasoning trajectories $\tau$. In our implementation, the specialist is trained with standard teacher-forced
SFT on question--answer pairs $(q,a)$; it does not explicitly optimize or
marginalize over latent reasoning trajectories.
We instead use a trajectory-distribution abstraction to characterize the
behavior induced by such answer-level supervision.
Let $\pi_{\theta_1}(\tau\mid q)$ denote the distribution over reasoning
trajectories generated by the resulting specialist.
At this abstraction level, the probability assigned to an answer can be
viewed as aggregating over trajectories compatible with that answer:

\begin{equation}
\log \pi_{\theta_1}(a \given q)
= \log \sum_{\tau \to a} \pi_{\theta_1}(\tau \given q).
\label{eq:qa-loss}
\end{equation}
where $\tau \to a$ denotes trajectories consistent with the correct answer $a$ (or its equivalents). Consequently, multiple trajectory distributions may be equally consistent with the same QA supervision by inducing the same answer likelihood.

\paragraph{Optimization implicitly selects a trajectory distribution among many valid ones.}
\label{sec:extraction}
Although multiple trajectory distributions are valid with the same QA supervision, optimization does not treat them equally. To understand how optimization resolves this ambiguity, we examine the gradient of the objective:
\begin{equation}
\nabla \log \pi_{\theta_1}(a \given q)
=
\sum_{\tau \to a}
p_{\theta}(\tau \given q, a)
\nabla \log \pi_{\theta_1}(\tau \given q),
\quad
p_{\theta_1}(\tau \given q, a)
=
\frac{\pi_{\theta_1}(\tau \given q)}{\pi_{\theta_1}(a \given q)}.
\label{eq:grad}
\end{equation}
Eq.~\ref{eq:grad} shows that trajectories with larger posterior weight $p_{\theta_1}(\tau \given q,a)$ contribute more strongly to the gradient update. Since $p_{\theta_1}(\tau \given q,a)\propto \pi_{\theta_1}(\tau \given q)$ among answer-consistent trajectories, high-probability trajectories dominate the gradient update. Repeated optimization therefore resolves the underdetermination by selecting a particular trajectory distribution $\pi_{\theta_1}(\tau \given q)$ from many valid ones.

\paragraph{Distillation probes the latent trajectory distribution induced by specialist optimization.}
\label{sec:distill_commit}
The latent trajectory distribution induced by specialist optimization is difficult to characterize directly: it spans a vast space of variable-length reasoning sequences, and individual samples reveal only partial information about its structure and value as supervision. Distillation provides an operational probe of this distribution by examining what a student learns from its sampled trajectories. Crucially, the student receives these trajectories without inheriting the specialist's adapted parameters or optimization constraints. Under controlled student training, downstream differences therefore provide evidence of how specialist optimization shapes transferable supervision. Formally, when trajectories sampled from $\pi_{\theta_1}$ are used to train $\pi_{\theta'_0}$, the resulting distillation objective is:
\begin{equation}
\mathcal{L}(\theta'_0)
=
\mathbb{E}_{\tau \sim \pi_{\theta_1}(\cdot \given q)}
\bigl[-\log \pi_{\theta'_0}(\tau \given q)\bigr].
\label{eq:distill_obj}
\end{equation}

This quantity corresponds to the cross-entropy between $\pi_{\theta_1}$ and $\pi_{\theta'_0}$ and admits the decomposition
\begin{equation}
\begin{aligned}
\mathbb{E}_{\tau \sim \pi_{\theta_1}(\cdot \given q)}
\bigl[-\log \pi_{\theta'_0}(\tau \given q)\bigr]
&=
\mathbb{E}_{\tau \sim \pi_{\theta_1}(\cdot \given q)}
\left[
\log \frac{\pi_{\theta_1}(\tau \given q)}{\pi_{\theta'_0}(\tau \given q)}
\right]  -
\mathbb{E}_{\tau \sim \pi_{\theta_1}(\cdot \given q)}
\bigl[\log \pi_{\theta_1}(\tau \given q)\bigr] \\
&=
\mathrm{KL}\bigl(\pi_{\theta_1}(\cdot \given q) \klgiven \pi_{\theta'_0}(\cdot \given q)\bigr) +
\mathcal{H}\bigl(\pi_{\theta_1}(\cdot \given q)\bigr).
\end{aligned}
\end{equation}
where $\mathcal{H}(\pi_{\theta_1}(\cdot\given q))=
-\mathbb{E}_{\tau\sim\pi_{\theta_1}(\cdot\given q)}[\log \pi_{\theta_1}(\tau\given q)]$ denotes the entropy of $\pi_{\theta_1}(\cdot\given q)$ and is independent of $\pi_{\theta'_0}$. Therefore, optimizing w.r.t. $\pi_{\theta'_0}$ is equivalent to minimizing
$\mathrm{KL}\bigl(\pi_{\theta_1}(\cdot \given q)\klgiven \pi_{\theta'_0}(\cdot \given q)\bigr),$ which drives the distilled model $\pi_{\theta'_0}$ to approximate $\pi_{\theta_1}$ in the trajectory space. Consequently, once the trajectory distribution $\pi_{\theta_1}$ is fixed, the behavior of the $\pi_{\theta'_0}$ is largely determined by $\pi_{\theta_1}$. Distillation thus probes the downstream consequences of trajectory selection without requiring an explicit characterization of the full trajectory distribution.

\subsection{Specialist Training as a Key Design Variable for Distillation Data}
\label{sec:traj_learnable}
The preceding analysis connects specialist training to downstream
data design: different training strategies can induce different
trajectory distributions under identical QA supervision, thereby
changing the supervision available for distillation. The specialist's
training procedure is therefore a key design variable for shaping
what the student learns. This raises a practical question: how can
specialist adaptation be controlled to shape the resulting
distillation data?

One approach is to regulate distributional drift from the origin
model. Such control allows domain-specific QA supervision to reshape
the trajectory distribution while constraining its departure from
the origin model's behavior. A canonical formulation of this
principle is a KL-constrained objective:
\begin{equation}
\max_{\pi_{\theta_1}}
\quad
\mathbb{E}_{(q,a)\sim \mathcal{D}}[\log \pi_{\theta_1}(a \given q)]
\quad
\text{s.t.}
\quad
\mathrm{KL}\big(\pi_{\theta_1}(\cdot\given q) \klgiven \pi_{\theta_0}(\cdot\given q) \big) \le \delta,
\label{eq:kl_penalty}
\end{equation}
where a small $\delta > 0$ limits how far $\pi_{\theta_1}$ can drift from the original model $\pi_{\theta_0}$. Under this constraint, the induced trajectory distribution can be viewed as a reweighted version of the $\pi_{\theta_0}$:
\begin{equation}
\pi_{\theta_1}(\tau\given q)
\propto
\pi_{\theta_0}(\tau\given q)\exp\left(\frac{w(\tau,q,a)}{\lambda}\right),
\label{eq:kl_reweight}
\end{equation}
where $w(\tau,q,a)$ is an implicit quantity reflecting how a trajectory contributes to increasing $\log \pi(a \given q)$, and $\lambda > 0$ is a Lagrange multiplier controlling the strength of the constraint. In our setting, the KL--constrained formulation serves only as a characterization of the induced trajectory reweighting behavior. We use standard QA-only supervised fine-tuning without explicit KL regularization, so neither $w(\tau,q,a)$ nor $\lambda$ is explicitly specified during training. The derivation of \cref{eq:kl_reweight} is provided in \appref{app:kl_reweighting}. Substituting \cref{eq:kl_reweight} into the distillation objective \cref{eq:distill_obj} yields:
% \begin{equation}
% \mathcal{L}(\theta)
% =
% \mathbb{E}_{\tau \sim \pi_{\theta_0}}
% \left[
% \underbrace{\exp\left(\frac{w(\tau,x,y)}{\lambda}\right)}_{\text{quality-aware reweighting}}
% \cdot
% (-\log \pi'_{\theta}(\tau \given x))
% \right],
% \end{equation}

\begin{equation}
\mathcal{L}(\theta'_0)
\propto
\mathbb{E}_{\tau \sim \pi_{\theta_0}}
\biggl[
\underbrace{\exp\left(\frac{w(\tau,q,a)}{\lambda}\right)}_{\text{quality-aware reweighting}}
\cdot
(-\log \pi_{\theta'_0}(\tau \given q))
\biggr],
\end{equation}
This form shows that the fine-tuned trajectory distribution $\pi_{\theta_1}(\tau \given q)$ is obtained by reweighting the base distribution $\pi_{\theta_0}(\tau \given q)$. Under this characterization, trajectories that contribute more strongly to
the answer-level objective receive greater relative weight while the overall distribution remains anchored to the origin model.
\section{Experiments}\label{sec:exp}

\begin{table*}[!t]
    \renewcommand{\arraystretch}{1.0}
    \setlength{\tabcolsep}{4.2pt} 
    \footnotesize
    \centering
    \captionsetup{skip=0pt}
    \caption{
    Summary of training setups for our Specialist-Distillation pipeline, 
    additional \colorbox{originbronze}{explicit KL-drift control} analysis with ASFT, and two 
    \colorbox{baselineorange}{\textit{self-training}} baselines.
    }
    \label{tab:settings}
    
    \begin{tabular}{l c l}
        \toprule
        \textbf{Model Setting} &
        \textbf{Tuning Data Format} &
        \textbf{Implementation Brief} \\
        \midrule
        
        $\FFT$ Specialist $\Mspec$ 
        & \multirow{8}{*}{$(q, a^*)$} 
        & \makecell[l]{
        \quad $\FFT$, $\LORA$, and $\LST$ share the same hyperparameters, except \\
        for those specific to $\LORA$ and $\LST$. See 
        \appref{app:exp_details:training} for details.
        } \\

        $\LORA$ Specialist $\Mspec$ 
        & 
        & \makecell[l]{
        \quad Rank $r = 64$, $\alpha = 2r$, 
        $\texttt{lora\_dropout}=0.05$, \texttt{target:all}. \\
        We provide ablation study on LoRA configs in 
        \appref{app:sup_analysis:lora}.
        } \\

        $\LST$ Specialist $\Mspec$ 
        & 
        & \makecell[l]{
        \quad Only update selected layers. We adopt \citet{llamax2}'s best- \\performing 
        config for $\Mzero$: update bottom $4$ and top $16$ layers.
        } \\

        \rowcolor{originbronze}
        \makecell[l]{
        $\ASFT$ Specialist $\Mspec$ \\
        [-0.5ex]\scriptsize explicit KL-drift control
        }
        &
        &
        \makecell[l]{
        \quad Explicitly regularizes $\Mspec$ toward $\Mzero$ with a KL penalty; \\
        $\lambda \in \{0.05, 0.2, 0.5\}$ controls anchoring strength.
        See \cref{sec:analysis:asft} for details.
        } \\

        \midrule
        
        $\FFT$ Distilled $\Mdist$ 
        & $(q, \hat{\tau}, \hat{a})$ 
        & \makecell[l]{
        \quad All $\Mdist$'s are $\FFT$-trained identically from $\Mzero$ \ \textsf{Qwen3-8B}, \\
        varying only in the rationale data generated by their corresponding $\Mspec$.
        } \\

        \rowcolor{baselineorange}
        \makecell[l]{
        Self-Distill $\Mdist$ \\
        [-0.5ex] \scriptsize $q \to \hat{\tau}, \hat{a}$
        } 
        & $(q, \hat{\tau}, \hat{a})$ 
        & \makecell[l]{
        \quad $\Mzero$ is prompted with only the question $q$ to generate both the \\
        reasoning path $\hat{\tau}$ and the prediction answer $\hat{a}$.
        } \\

        \rowcolor{baselineorange}
        \makecell[l]{
        Self-Rationalize $\Mdist$ \\
        [-0.5ex] \scriptsize $q, a^{*} \to \hat{\tau}$
        } 
        & $(q, \hat{\tau}, a^*)$ 
        & \makecell[l]{
        \quad $\Mzero$ is prompted with both the question $q$ and the ground-truth \\
        answer $a^*$, and is tasked to generate the rationale $\hat{\tau}$ that leads to $a^*$.
        } \\
        
        \bottomrule
    \end{tabular}
    \vspace{-10pt}
\end{table*}
% \subsection{Experimental Setup} \label{sec:exp:setup}

% \paragraph{Specialist Fine-tuning Strategies.}
% We deploy \qwenEightB \citep{qwen3} as our origin model $\Mzero$. To investigate how different FT strategies affect the quality of generated rationales, we train the specialist model $\Mspec$ from $\Mzero$ using three different straigies: (1) Full Fine-Tuning ($\FFT$), (2) $\LORA$, and (3) Layer-Selective Tuning ($\LST$).

% \paragraph{Distillation Pipeline.}
% Once trained, each specialist $\Mspec$ variant generates candidate reasoning chains and prediction answers $(\hat{c}, \hat{a})$, which are filtered to construct the valid and correct $(q, \hat{c}, \hat{a})$ datasets. Crucially, to isolate the impact of data quality from the specialist model's learning capacity, all downstream distillation models (\textit{Distill} $\Mdist$) are strictly trained via $\FFT$ on the same amount of (downsampled) data, regardless of the strategy used to train their corresponding specialists $\Mspec$.

\subsection{Experimental Setup} \label{sec:exp:setup}

We summarize in \cref{tab:settings} the overarching experimental setup, including required data formats, training objectives, and key implementation details. Comprehensive training, inference, rationale filtration and evaluation protocols are deferred to \apprefs{app:exp_details}{app:bench_details}. The core components are described below.

\paragraph{Specialist Fine-Tuning Strategies.}
We deploy \textsf{Qwen3-8B} as our origin model $\Mzero$. To investigate how different fine-tuning strategies affect the quality of generated rationales, we train the specialist models $\Mspec$ starting from $\Mzero$ using three methods as in \cref{tab:settings}: (1) Full Fine-Tuning ($\FFT$), (2) $\LORA$ \citep{lora}, and (3) Layer-Selective Tuning \citep[$\LST$,][]{llamax2}. We additionally evaluate Anchored Supervised Fine-Tuning \citep[$\ASFT$;][]{asft}
as an explicit KL-based drift-control mechanism in
\cref{sec:analysis:asft}.

\paragraph{Distillation Pipeline.}
Once trained, each specialist $\Mspec$ variant generates candidate chain-of-thoughts \citep[CoT;][]{cot} and answers $(\hat{\tau}, \hat{a})$, which are \emph{filtered} such that $\hat{a}$ is equivalent to ground-truth answer $a^*$ and \textit{complete} CoT to construct valid $(q, \hat{\tau}, \hat{a})$ pool (see \cref{sec:analysis:traj_quality} and \appref{app:bench_details} for details). Crucially, to isolate the impact of data quality from the \emph{student} model's learning capacity, all \textit{distilled} models $\Mdist$ are $\FFT$-trained on an equal amount of subsampled data and identical configurations, regardless of their corresponding \textit{specialist} model $\Mspec$ tuning strategy.

\paragraph{Baselines.}
We introduce two \emph{self-training} baselines (\cref{tab:settings,tab:main}) derived directly from $\Mzero$: (1) \textbf{Self-Distill}: unconditional rationale generation and (2) \textbf{Self-Rationalize}: answer-conditioned rationale generation. We further report larger models \textsf{Qwen3-\{14,32\}B} for scale-based comparisons.

\paragraph{Training Data and Evaluation Suites.}\label{sec:exp:setup:eval_datasets}
We organize our datasets and benchmarks into four categories (see \appref{app:bench_details} for all benchmarks we evaluate): 
(1) \textbf{Training} data utilized for fine-tuning within each target domain;
% (2) \textbf{In-Task (\ID)} \footnote{We term \emph{in-task} since some datasets lack an official \emph{test set}, so we use alternatives with a closely matched distribution.} benchmarks that share the same domain and task formulation as the training data; 
(2) \textbf{In-Task (\ID)} benchmarks that share the same domain and task formulation as the training data, using the official test split when available, otherwise a distribution-wise closely matched benchmark; 
(3) \textbf{In-Domain (\SID)} benchmarks that remain within the target domain but differ in task distribution and difficulty level, evaluating robustness under domain shift~\citep{domain_adaptation}; 
and (4) \textbf{Out-of-Domain (\OOD)} benchmarks drawn from domains entirely different from the target domain, assessing broader generalization. For \OOD, we use $1$ benchmarks for complex reasoning \citep{bbeh}, $2$ for mathematics \citep{aime26}, and $2$ for coding \citep{lcb}.

\paragraph{Domains and Rationale Filtrations.} We study three \emph{target} domains for Training, \ID, and \SID:
(1) \textbf{Chemistry} (\chem). We train on \textsf{SMol}~\citep{smol}, covering molecular understanding and generation across $14$ subtasks, whose official test split serves as the \ID. For \SID, we evaluated on general chemistry benchmarks. During rationale filtration, we apply subtask-specific criteria (\appref{app:bench_details:chem:smol:stage2}) to accommodate its diverse output formats and evaluation protocols.
(2) \textbf{Physics} (\phys). We train on the university-level physics subset of \textsf{MegaScience}~\citep{megascience}. For \ID, we evaluate on university-level subsets from \textsf{PHYSICS} benchmark~\citep{physics}; for \SID, on high-school physics benchmarks. Rationales are retained only for answers that pass rule-based symbolic and numerical verification.
(3) \textbf{Low-Resource Multilingualism} (\MT). We train on \textsf{OPUS}~\citep{opus} for \emph{bi}-directional English -- $8$ low-resource languages translation. For \ID, we evaluate the same $16$ translation task using \textsf{Flores-101}~\citep{flores101}; for \SID, we assess general reasoning in these $8$ languages, beyond translation. Rationales are ranked by sentence-level \texttt{spBLEU}~\citep{bleu,flores101} and the top-performing $20\%$ subset is retained.

\begin{table*}[!t]
    \renewcommand{\arraystretch}{1.0}
    \setlength{\tabcolsep}{1.5pt}
    % \small
    \footnotesize
    \centering
% \caption{
% Performance comparison across Chemistry, Physics, and Multilingualism. 
% The \textsf{Qwen3-8B} origin model and its corresponding group label are highlighted in light blue, while the two untuned baselines (Self-Distill and Self-Rationalize) are highlighted in light pink. 
% }
% \captionsetup{skip=-1pt}
\caption{
\textbf{Performance comparison across Chemistry, Physics, and Multilingualism.}
The parenthesized, e.g., \SID$(4)$, are the number of benchmarks (subsets) we use for this category.
Chemistry (\SMol) \ID score aggregate its $14$ subtasks, with its computation detailed in \appref{app:bench_details:chem:smol}.
All other metrics are macro--averages over the corresponding categories. The same five \OOD\ benchmarks ($5^{\scriptscriptstyle =}$) are used across all three \textit{target} domain settings. Bold denotes the best result per column among \colorbox{originblue}{models derived from \textsf{Qwen3-8B}}; \textsf{-14/32B} are excluded.
}
    \label{tab:main}
    \begin{tabular}{@{}ll ccc ccc ccc}
        \toprule
        \multicolumn{2}{@{}l}{\multirow{2}{*}{\textbf{Model}}} &
          \multicolumn{3}{c}{\textbf{Chemistry}} &
          \multicolumn{3}{c}{\textbf{Physics}} &
          \multicolumn{3}{c}{\textbf{Multilingualism}} \\
        \cmidrule(lr){3-5} \cmidrule(lr){6-8} \cmidrule(lr){9-11}
        \multicolumn{2}{c}{} &
          \textbf{\ID$(14)$} & \textbf{\SID$(4)$} & \textbf{\OOD$(5^{\scriptscriptstyle =})$} &
          \textbf{\ID$(1)$} & \textbf{\SID$(4)$} & \textbf{\OOD$(5^{\scriptscriptstyle =})$} &
          \textbf{\ID$(16)$} & \textbf{\SID$(16)$} & \textbf{\OOD$(5^{\scriptscriptstyle =})$} \\
        \midrule
        \rowcolor{originblue}
        \multicolumn{2}{@{}l}{\textsf{Qwen3-8B} $\Mzero$} &
          19.34 & 60.19 & 42.22 & 90.21 & 68.61 & 42.22 & 29.48 & 44.99 & \textbf{42.22} \\
        \multicolumn{2}{@{}l}{\textsf{Qwen3-14B}} &
          23.80 & 63.67 & 48.95 & 91.33 & 73.06 & 48.95 & 32.31 & 48.43 & 48.95 \\
        \multicolumn{2}{@{}l}{\textsf{Qwen3-32B}} &
          25.35 & 68.09 & 51.23 & 92.45 & 77.86 & 51.23 & 33.78 & 49.13 & 51.23 \\
        \midrule

        % \multirow{10}{*}{\cellcolor{originblue}\rotatebox[origin=c]{90}{Origin Model $\Mzero$ \textsf{Qwen3-8B}}} &

\multirow{10}{*}{%
  \begingroup
  \setlength{\fboxsep}{0pt}%
  \colorbox{originblue}{%
    \makebox[0.4cm][c]{%
      \rule[-5em]{0pt}{8em}%
      \rotatebox[origin=c]{90}{\footnotesize All tuned on $\Mzero$ \textsf{Qwen3-8B}}%
    }%
  }%
  \endgroup
} &

        \cellcolor{baselineorange} $q \to \hat{\tau}, \hat{a} \ \ $ Self-Distill $\Mdist$ &
        \cellcolor{baselineorange}21.06 & \cellcolor{baselineorange}59.28 & \cellcolor{baselineorange}42.04 &
        \cellcolor{baselineorange}91.11 & \cellcolor{baselineorange}67.97 & \cellcolor{baselineorange}\textbf{43.54} &
        \cellcolor{baselineorange}29.30 & \cellcolor{baselineorange}\textbf{45.94} & \cellcolor{baselineorange}39.66 \\
        &
        \cellcolor{baselineorange} $q, a^{*} \to \hat{\tau}$ Self-Rationalize $\Mdist$ &
        \cellcolor{baselineorange}17.92 & \cellcolor{baselineorange}58.78 & \cellcolor{baselineorange}33.14 &
        \cellcolor{baselineorange}87.89 & \cellcolor{baselineorange}68.64 & \cellcolor{baselineorange}41.78 &
        \cellcolor{baselineorange}6.02 & \cellcolor{baselineorange}27.26 & \cellcolor{baselineorange}27.71 \\
        \cmidrule{2-11}
        & $\FFT$ \textit{Specialist} $\Mspec$ &
          \textbf{59.65} & 57.72 & 33.60 & 90.73 & 68.43 & 16.31 & \textbf{35.93} & 23.86 & 6.90 \\
        & $\ \subarrow \FFT$ \textit{Distilled} $\Mdist$ &
          40.55 & 46.14 & 25.20 & 91.85 & 68.51 & 14.09 & 30.77 & 27.78 & 15.08 \\
        \cmidrule{2-11}
        & $\LORA$ \textit{Specialist} $\Mspec$ &
          28.57 & 60.36 & 42.65 & 92.83 & 68.98 & 38.03 & 20.47 & 43.16 & 41.41 \\
        & $\ \subarrow \FFT$ \textit{Distilled} $\Mdist$ &
          29.80 & 59.31 & 34.89 & 91.93 & 69.87 & 38.26 & 22.97 & 30.87 & 35.89 \\
        \cmidrule{2-11}
        & $\LST$ \textit{Specialist} $\Mspec$ &
          29.61 & 60.20 & \textbf{44.03} & \textbf{93.65} & 70.13 & 38.50 & 35.68 & 43.99 & 39.08 \\
        & $\ \subarrow \FFT$ \textit{Distilled} $\Mdist$ &
          29.54 & \textbf{60.47} & 42.11 & 92.75 & \textbf{70.84} & 40.80 & 34.17 & 43.87 & 41.23
        \\ \bottomrule
    \end{tabular}
    \vspace{-10pt}
\end{table*}
\subsection{Main Results} \label{sec:exp:results}
\Cref{tab:main} compares the origin model, larger same-family models, untuned self-training baselines, and our specialist--distillation pipeline across Chemistry, Physics, and Multilingualism under \ID, \SID, and \OOD evaluation. Benchmark composition and metric computation details are provided in \appref{app:bench_details}.

\paragraph{QA-only specialist distillation consistently improves target-domain performance.} \label{sec:exp:results:qa_improvement}
Across $\FFT, \LORA, \text{and} \LST$ tuning strategies and three target domains, our key observations from \cref{tab:main} demonstrate that even without gold rationales, explicitly training a QA-only \textit{specialist} to induce domain-specific reasoning traces, and subsequently transferring them to a \textit{distilled} model, consistently yields robust and significant target-domain growth:
% \begin{itemize}[leftmargin=10pt, topsep=-5pt, itemsep=-2.5pt]
\begin{itemize}[leftmargin=10pt]
    \item \textbf{Substantial target-domain improvements.} Across all configurations, both the \textit{specialist} models $\Mspec$ and their downstream \textit{distilled} models $\Mdist$ achieve massive \textit{in-task} (\ID) capability gains compared to the origin \textsf{Qwen3-8B} $\Mzero$ and two \textit{self-training} baselines.
    \item \textbf{Bridging a $4\times$ parameter gap.} Our pipeline enables an 8B model to ``punch above its weight class'' without relying on human-annotated rationales. For example, our $\LST$-distilled model ($29.54 / 92.75/ 34.17$ for all \ID) surpass the zero-shot performance of the $4\times$ larger \textsf{Qwen3-32B} ($25.35 / 92.45/ 33.78$). This highlights that extracting latent reasoning paths from a specialist is a highly parameter-efficient paradigm compared to merely scaling up generalist models.
    % \item \textbf{Untrained post-hoc rationalization yields toxic supervision.} A notable failure case is the \textit{Self-Rationalize} baseline, which severely degrades performance across domains—most drastically in Multilingualism, where \ID plummets from $29.48$ to $6.02$. As discussed in \cref{sec:intro}, when a weak model is forced to rationalize an answer it cannot naturally reach, the probability mass concentrates on hallucinatory or shortcut trajectories ($\tau_{\sim}$). Distilling these flawed trajectories acts as poisoned data, actively harming the model's inherent capabilities.
    % \item \textbf{Untrained post-hoc rationalization yields toxic supervision.} A notable failure case is the \textit{Self-Rationalize} baseline, which severely degrades performance across domains—most drastically in Multilingualism, where \ID plummets from $29.48$ to $6.02$. As discussed in \cref{sec:intro}, when a weak model is forced to rationalize an answer it cannot naturally reach, the probability mass concentrates on hallucinatory or shortcut trajectories ($\tau_{\sim}$). Distilling these flawed trajectories acts as poisoned data, actively harming the model's inherent capabilities. Our observation, particularly the severe degradation in Multilingualism, corroborates recent findings that forced post-hoc explanations often induce spurious and unreliable reasoning \citep{key1, key2}.
    \item \textbf{Toxic post-hoc rationalization.} The untrained \textit{Self-Rationalize} baseline severely degrades performance, notably plummeting \MT \ID from $29.48$ to $6.02$, which corroborates findings that post-hoc reasoning on translation is often spurious and unreliable \citep{please_translate_again,test_time_scaling_mt}. Forcing weak models to rationalize answers induces hallucination or shortcuts that poison distillation. 
    
\end{itemize}
% Together, these findings 

% \paragraph{Different QA-only supervision methods implicitly select different trajectory distributions and leads to specialization--generalization trade-offs.}
% The main difference across tuning strategies is not simply their absolute strength, but the trade-off they induce between domain specialization and cross-distribution generalization. $\LST$ tends to be more comparable on \SID/\OOD even when its \ID gains are less aggressive than those of $\FFT$ or $\LORA$. This is consistent with our discussion in \cref{sec:method}: because the latent trajectory distribution is underdetermined, different finetuning strategies implicitly \emph{select} different rationale distributions under the same QA-only supervision. These differences then surface as distinct performance profiles across splits. Therefore, the choice of QA-only supervision method should be made according to the desired operating point on the specialization--generalization spectrum, rather than a single universal notion of optimality.

\paragraph{Different QA-only supervision methods implicitly select different trajectory distributions and lead to distinct specialization-generalization trade-offs.}
Overall, across all $3$ domains, $\LST$ consistently improves \ID performance over the origin, achieves modest gains on \SID, and remains on par on \OOD. While $\FFT$ yields impressive \ID gains in specific domains such as chemistry, it severely sacrifices both the performance of \SID and \OOD, suffering from \textit{catastrophic forgetting} \citep{underperform_ood}. $\LORA$ exhibits a trade-off pattern similar to $\LST$ but generally falls short of $\LST$ across all $9$ splits. These distinct performance profiles confirm that different QA-only tuning strategies implicitly select different rationale distributions. Thus, the optimal tuning method depends on the desired specialization--generalization trade--off rather than a universal optimum.

\paragraph{Distillation reliably inherits and refines specialist capabilities.}\label{sec:exp:results:mono_corr}
\begin{figure}[t]
    \centering

    % ============================================================
    % Left: Rank-preservation figure
    % ============================================================
    \begin{minipage}[t]{0.42\linewidth}
        \centering
        \vspace{0pt}

        \begin{tikzpicture}
        \begin{axis}[
            name=mainaxis,
            width=\linewidth,
            height=\linewidth,
            xmin=0, xmax=100,
            ymin=0, ymax=100,
            xlabel={\scriptsize Specialist $\Mspec$ Performance (\%)},
            ylabel={\scriptsize Distilled $\Mdist$ Performance (\%)},
            grid=both,
            grid style={line width=.1pt, draw=gray!20},
            major grid style={line width=.2pt, draw=gray!40, dashed},
            axis x line*=bottom,
            axis y line*=left,
            tick align=outside,
            tick pos=left,
            tick label style={font=\tiny},
            label style={font=\tiny},
            xlabel style={at={(axis description cs:0.5,-0.08)}},
            ylabel style={at={(axis description cs:-0.10,0.5)}},
        ]

            % Spearman annotation
            \node[
                anchor=north west,
                draw=gray!50,
                fill=none,
                rounded corners=2pt,
                inner sep=2pt,
                align=left,
                font=\scriptsize
            ]
            at (axis cs:0,100) {
                \textbf{Strict Rank Preservation} \\
                Spearman's $\rho = 0.9573$ \\
                Each point is one \\
                $(\Mspec,\Mdist)$ performance \\
                pair as in \cref{tab:main}.
            };

            % y = x
            \addplot[gray, dashed, thick, domain=-5:100] {x};
            \node[
                anchor=south east,
                text=gray,
                rotate=45,
                font=\scriptsize
            ]
            at (axis cs:88,90) {$y=x$ (Perfect Mirror)};

            % ====================================================
            % Chemistry
            % ====================================================
            % FFT
            \addplot[only marks, mark=diamond*, draw=ETHRed, fill=ETHRed,
                     mark size=2pt]
                coordinates {(59.65,40.55)};
            \addplot[only marks, mark=diamond*, draw=ETHRed, fill=ETHRed!40,
                     mark size=2pt]
                coordinates {(57.72,46.14)};
            \addplot[only marks, mark=diamond*, draw=ETHRed, fill=white,
                     mark size=2pt]
                coordinates {(33.60,25.20)};

            % LoRA
            \addplot[only marks, mark=triangle*, draw=ETHRed, fill=ETHRed,
                     mark size=2.5pt]
                coordinates {(28.57,29.80)};
            \addplot[only marks, mark=triangle*, draw=ETHRed, fill=ETHRed!40,
                     mark size=2.5pt]
                coordinates {(60.36,59.31)};
            \addplot[only marks, mark=triangle*, draw=ETHRed, fill=white,
                     mark size=2.5pt]
                coordinates {(42.65,34.89)};

            % LST
            \addplot[only marks, mark=square*, draw=ETHRed, fill=ETHRed,
                     mark size=2pt]
                coordinates {(29.61,29.54)};
            \addplot[only marks, mark=square*, draw=ETHRed, fill=ETHRed!40,
                     mark size=2pt]
                coordinates {(60.20,60.47)};
            \addplot[only marks, mark=square*, draw=ETHRed, fill=white,
                     mark size=2pt]
                coordinates {(44.03,42.11)};

            % ====================================================
            % Physics
            % ====================================================
            % FFT
            \addplot[only marks, mark=diamond*, draw=ETHBlue, fill=ETHBlue,
                     mark size=2pt]
                coordinates {(90.73,91.85)};
            \addplot[only marks, mark=diamond*, draw=ETHBlue, fill=ETHBlue!40,
                     mark size=2pt]
                coordinates {(68.43,68.51)};
            \addplot[only marks, mark=diamond*, draw=ETHBlue, fill=white,
                     mark size=2pt]
                coordinates {(16.31,14.09)};

            % LoRA
            \addplot[only marks, mark=triangle*, draw=ETHBlue, fill=ETHBlue,
                     mark size=2.5pt]
                coordinates {(92.83,91.93)};
            \addplot[only marks, mark=triangle*, draw=ETHBlue, fill=ETHBlue!40,
                     mark size=2.5pt]
                coordinates {(68.98,69.87)};
            \addplot[only marks, mark=triangle*, draw=ETHBlue, fill=white,
                     mark size=2.5pt]
                coordinates {(38.03,38.26)};

            % LST
            \addplot[only marks, mark=square*, draw=ETHBlue, fill=ETHBlue,
                     mark size=2pt]
                coordinates {(93.65,92.75)};
            \addplot[only marks, mark=square*, draw=ETHBlue, fill=ETHBlue!40,
                     mark size=2pt]
                coordinates {(70.13,70.84)};
            \addplot[only marks, mark=square*, draw=ETHBlue, fill=white,
                     mark size=2pt]
                coordinates {(38.50,40.80)};

            % ====================================================
            % Multilingualism
            % ====================================================
            % FFT
            \addplot[only marks, mark=diamond*, draw=ETHGreen, fill=ETHGreen,
                     mark size=2pt]
                coordinates {(35.93,30.77)};
            \addplot[only marks, mark=diamond*, draw=ETHGreen, fill=ETHGreen!40,
                     mark size=2pt]
                coordinates {(23.86,27.78)};
            \addplot[only marks, mark=diamond*, draw=ETHGreen, fill=white,
                     mark size=2pt]
                coordinates {(6.90,15.08)};

            % LoRA
            \addplot[only marks, mark=triangle*, draw=ETHGreen, fill=ETHGreen,
                     mark size=2.5pt]
                coordinates {(20.47,22.97)};
            \addplot[only marks, mark=triangle*, draw=ETHGreen, fill=ETHGreen!40,
                     mark size=2.5pt]
                coordinates {(43.16,30.87)};
            \addplot[only marks, mark=triangle*, draw=ETHGreen, fill=white,
                     mark size=2.5pt]
                coordinates {(41.41,35.89)};

            % LST
            \addplot[only marks, mark=square*, draw=ETHGreen, fill=ETHGreen,
                     mark size=2pt]
                coordinates {(35.68,34.17)};
            \addplot[only marks, mark=square*, draw=ETHGreen, fill=ETHGreen!40,
                     mark size=2pt]
                coordinates {(43.99,43.87)};
            \addplot[only marks, mark=square*, draw=ETHGreen, fill=white,
                     mark size=2pt]
                coordinates {(39.08,41.23)};

        \end{axis}

% ========================================================
% Legend inside plot: lower-right, transparent background
% ========================================================
\node[
    draw=black!30,
    rounded corners=2pt,
    fill=none,
    fill opacity=0,
    text opacity=1,
    anchor=south east,
    inner sep=2pt,
    outer sep=0pt,   % 关键：去掉 node 外边距
    font=\tiny
]
at (rel axis cs:0.98,-0.05) {
    \renewcommand{\arraystretch}{1.1}
    \setlength{\tabcolsep}{2pt}
    \begin{tabular}{
        @{}cl@{\hspace{2mm}}
        cl@{\hspace{2mm}}
        cl@{}
    }
        \multicolumn{2}{@{}l}{\textbf{$\Mspec$--Tuning}}
        &
        \multicolumn{2}{l}{\textbf{Domain}}
        &
        \multicolumn{2}{l@{}}{\textbf{Cat.}}
        \\

        \tikz[baseline=-0.5ex]
            \draw plot[
                mark=diamond*,
                mark options={draw=gray!70,fill=gray!70},
                mark size=2pt
            ] coordinates {(0,0)};
        & $\FFT$
        &
        \tikz[baseline=-0.5ex]
            \draw plot[
                mark=*,
                mark options={draw=ETHRed,fill=ETHRed},
                mark size=2pt
            ] coordinates {(0,0)};
        & \chem
        &
        \tikz[baseline=-0.5ex]
            \draw plot[
                mark=*,
                mark options={draw=gray!70,fill=gray!70},
                mark size=2pt
            ] coordinates {(0,0)};
        & \ID
        \\

        \tikz[baseline=-0.5ex]
            \draw plot[
                mark=triangle*,
                mark options={draw=gray!70,fill=gray!70},
                mark size=2.5pt
            ] coordinates {(0,0)};
        & $\LORA$
        &
        \tikz[baseline=-0.5ex]
            \draw plot[
                mark=*,
                mark options={draw=ETHBlue,fill=ETHBlue},
                mark size=2pt
            ] coordinates {(0,0)};
        & \phys
        &
        \tikz[baseline=-0.5ex]
            \draw plot[
                mark=*,
                mark options={draw=gray!70,fill=gray!40},
                mark size=2pt
            ] coordinates {(0,0)};
        & \SID
        \\

        \tikz[baseline=-0.5ex]
            \draw plot[
                mark=square*,
                mark options={draw=gray!70,fill=gray!70},
                mark size=2pt
            ] coordinates {(0,0)};
        & $\LST$
        &
        \tikz[baseline=-0.5ex]
            \draw plot[
                mark=*,
                mark options={draw=ETHGreen,fill=ETHGreen},
                mark size=2pt
            ] coordinates {(0,0)};
        & \MT
        &
        \tikz[baseline=-0.5ex]
            \draw plot[
                mark=*,
                mark options={draw=gray!70,fill=white,thick},
                mark size=2pt
            ] coordinates {(0,0)};
        & \OOD
    \end{tabular}
};

        \end{tikzpicture}

        \captionsetup{skip=-4pt}
        % \caption{
        %     \textbf{Rank preservation from \textit{Specialist} to
        %     \textit{Distilled} models.}
            % Each point is one $(\Mspec,\Mdist)$ performance pair as in
            % \cref{tab:main}.
        % }
        \caption{\textbf{Rank Preservation} from $\Mspec$ to $\Mdist$.}
        \label{fig:mono_corr}

    \end{minipage}
    \hfill
    % ============================================================
    % Right: Cross-model transferability table
    % ============================================================
    \begin{minipage}[t]{0.5\linewidth}
        \centering
        \vspace{0pt}

        % Make this minipage use TABLE numbering/caption
        \captionsetup{type=table}
        \captionsetup{skip=2pt}

        \footnotesize
        \renewcommand{\arraystretch}{1.00}
        \setlength{\tabcolsep}{1.5pt}

        \caption{
            \textbf{Cross-model Transferability on Chemistry.}
            $\FFT/\LST$-tuned \textsf{Intern-S1-mini}
            ($\IMzero$ \xintern) serves as the specialist $\IMspec$,
            and $\pi_\phi$-generated $(q,\hat{\tau},\hat{a})$
            are distilled into \textsf{Qwen3-8B} (\xqwen).
        }
        \label{tab:cross_model}

        \begin{tabular}{@{}lccc@{}}
            \toprule
            \textbf{Models on Chemistry}
            & \textbf{\ID$(14)$}
            & \textbf{\SID$(4)$}
            & \textbf{\OOD$(5^{\scriptscriptstyle =})$}
            \\
            \midrule

            \rowcolor{originblue}
            Origin $\Mzero$ \xqwen
            & 19.34
            & 60.19
            & \textbf{42.22}
            \\

            \midrule

            \rowcolor{originpetrol}
            Origin $\IMzero$ \xintern
            & 34.80
            & \textbf{62.51}
            & 37.77
            \\

            \rowcolor{baselineorange}
            $\ \subarrow$ $\FFT$-\textit{Distill} to $\Mdist$ \xqwen
            & 31.33
            & 60.25
            & 33.79
            \\

            \midrule

            $\FFT$ \textit{Specialist} $\IMspec$ \xintern
            & \textbf{42.83}
            & 54.86
            & 22.43
            \\

            $\ \subarrow \FFT$-\textit{Distill} to $\Mdist$ \xqwen
            & 36.91
            & 56.99
            & 25.90
            \\

            \midrule

            $\LST$ \textit{Specialist} $\IMspec$ \xintern
            & 33.11
            & 60.31
            & 36.21
            \\

            $\ \subarrow \FFT$-\textit{Distill} to $\Mdist$ \xqwen
            & 32.72
            & 58.79
            & 37.46
            \\

            \bottomrule
        \end{tabular}

    \end{minipage}

    \vspace{-10pt}
\end{figure}
Beyond absolute metrics, an intriguing phenomenon observed in \cref{tab:main} is the strong \textit{rank} correlation between the capabilities of $\Mspec$ and its downstream $\Mdist$: across all $27$ experimental measurements, Spearman's rank correlation reaches $\rho=0.9573$ (\cref{fig:mono_corr}). To account for the dependence among $\ID$, $\SID$, and $\OOD$ measurements from the same specialist--student pair, we additionally conduct a domain-blocked, model-profile-level exact permutation test, which yields $p\mathrm{-value}=2/215\approx0.0093$. The relative performance ranking among tuning strategies (e.g., $\LST$ > $\LORA$ > $\FFT$ in \phys \ID) is therefore strongly reflected in the downstream \emph{distilled} models across the evaluated splits. This consistency supports the effectiveness of our variable-controlled pipeline: because all $\Mdist$ models are fine-tuned using the exact same $\FFT$ configuration, differences in their performance are closely associated with the rationale supervision $\hat{\tau}$ generated by their corresponding $\Mspec$. Consequently, stronger \textit{specialists} generally tend to produce supervision that leads to stronger \textit{distilled} models. Furthermore, in nearly half ($12/27$) of the splits, the \textit{distilled} model even exceeds its corresponding \textit{specialist}, indicating that distillation does not merely copy teacher behavior but can further refine the supervision induced by specialist-generated rationales. More broadly, distilled models consistently outperform the two untuned self-training baselines, suggesting that effective rationale supervision is better obtained from explicitly trained \textit{specialist} models than from the origin model alone or post-hoc self-rationalization. Together, these results support specialist training as an effective approach for producing rationale data useful for downstream distillation within the evaluated Qwen3-8B-based pipeline.

We also demonstrate in \appref{app:sup_analysis:model_scaling} that these specialization--generalization trade-offs and monotonic rank correlation remain consistent when scaling the origin model $\Mzero$ up to \textsf{Qwen3-14B}.
\section{Analysis: Data Scaling, Cross-model, KL Drift, Explicit KL Anchor, and Trajectory Quality}
% \section{Analysis}
\label{sec:analysis}
% \subsection{Data Scaling}
\subsection{Data scaling yields continuous in-task gains while preserving robustness.}
\begin{figure*}[t]
    \centering

    % 第一行
    \begin{subfigure}[t]{0.47\textwidth}
        \centering
    \begin{tikzpicture}
        % --- 第一层坐标轴：右 Y 轴（柱状图 \SID 和 \OOD） ---
        % 注意：为了不挡住底部的 X 轴标签，我们将这个 axis 画在底层，并隐藏它的 X 轴
        \begin{axis}[
            scale only axis,          % 严格固定绘图区尺寸，保证双轴完美对齐
            width=0.8\textwidth,     % 留出两侧写 Y 轴标签的空间
            height=3.3cm,             % 统一高度
            xmin=0, xmax=360,
            enlarge x limits=0.05,
            ymin=30, ymax=70,         % 右侧 Y 轴的数据范围 (适应 \SID 和 \OOD)
            ytick={30, 40, 50, 60, 70},
            axis y line*=right,       % 激活右侧 Y 轴
            axis x line=none,         % 隐藏该层的 X 轴，避免标签重叠
            ylabel={\footnotesize \chem \SID\ \& \OOD Score},
            ylabel near ticks,
            tick label style={font=\tiny},
            ybar=1pt,                 % 柱状图模式，1pt为两根柱子间的间距
            bar width=4.5pt,          % 柱子宽度
        ]
        
        % \SID 柱状图 (右 Y 轴)
        \addplot+[ybar, fill=ETHBlue!30, draw=ETHBlue!80!black, mark=none]
        plot coordinates {(0, 60.19) (50, 60.47) (100, 60.35) (150, 60.06) (200, 58.99) (250, 59.94) (300, 58.26) (360, 59.57)};
        
        % \OOD 柱状图 (右 Y 轴)
        \addplot+[ybar, fill=ETHPetrol!30, draw=ETHPetrol!80!black, mark=none]
        plot coordinates {(0, 65.87) (50, 62.79) (100, 61.03) (150, 62.4) (200, 61.35) (250, 64.28) (300, 56.79) (360, 60)};
        \end{axis}

        % --- 第二层坐标轴：左 Y 轴（折线图 \ID 以及 X 轴、网格） ---
        \begin{axis}[
            scale only axis,          % 必须与上面保持一致
            width=0.8\textwidth,
            height=3.3cm,
            xmin=0, xmax=360,
            enlarge x limits=0.05,
            ymin=15, ymax=35,         % 左侧 Y 轴的数据范围 (适应 \ID)
            ytick={18, 20, 25, 30, 35},
            axis y line*=left,        % 激活左侧 Y 轴
            axis x line*=bottom,      % 激活底部 X 轴
            ylabel={\footnotesize \chem \ID (\SMol) Score},
            xtick={0, 50, 100, 150, 200, 250, 300, 360},
            xticklabels={0, 50K, 100K, 150K, 200K, 250K, 300K, 360K Full},
            x tick label style={rotate=35, anchor=north east, font=\tiny},
            tick label style={font=\tiny},
            xlabel near ticks,
            ylabel near ticks,
            xmajorgrids=true,
            ymajorgrids=true,
            grid style=dashed,
            % 因为右下方有柱状图，建议将图例放在顶部居中，或者你原代码的右上角也行
            legend style={
                at={(0.5, 1.05)}, 
                anchor=north,
                nodes={scale=0.7, transform shape},
                fill=white,
                legend columns=3
            },
        ]
        
        % \ID 曲线 (左 Y 轴)
        \addplot+[line width=0.4mm, mark=star, mark options={scale=1.5}, color=ETHRed]
        plot coordinates {(0, 19.34) (50, 29.54) (100, 30.3) (150, 31.12) (200, 32.29) (250, 31.19) (300, 31.33) (360, 32.47)};
        \addlegendentry{\ID}

        % 为了让柱状图也能在这个图例框里显示，这里添加"虚拟图例 (Dummy Legend)"
        \addlegendimage{ybar, fill=ETHBlue!30, draw=ETHBlue!80!black, area legend}
        \addlegendentry{\SID}
        
        \addlegendimage{ybar, fill=ETHPetrol!30, draw=ETHPetrol!80!black, area legend}
        \addlegendentry{\OOD}
        
        \end{axis}
        
        % X 轴标题统一通过 node 放置在下方，避免错位
        \node at ([yshift=-0.8cm]current axis.south) {\footnotesize $\FFT$ Distillation Data Size};
    \end{tikzpicture}
    \caption{}
    \label{fig:data_scaling}
    \end{subfigure}\hfill
    \begin{subfigure}[t]{0.48\textwidth}
        \centering
        
    \begin{tikzpicture}
        \begin{axis}[
            % ==================== 尺寸调整区 ====================
            % width 控制图表宽度。设为 \linewidth 的 0.5 倍，完美适应半页宽
            width=\linewidth, 
            % height 控制图表高度。调低数值让图片更紧凑，不那么高
            height=5cm, 
            % ====================================================
            % 坐标轴样式
            axis x line=bottom,
            axis y line=left,
            ymin=0, ymax=4.5,
            ytick={0.0,1.0,2.0,3.0,4.0},
            % 将 X 轴左右稍微延伸(0.8到2.2)，给两侧内部的文字留出空间，使其不碰到 Y 轴
            xmin=0.8, xmax=2.2, 
            xtick={1,2},
            xticklabels={\chem $\Mspec$, \chem $\Mdist$},
            xticklabel style={font=\scriptsize, yshift=-1ex},
            ylabel={$\KL\big( \pi_{\bullet} \klgiven \Mzero \big)$},
            ylabel style={font=\scriptsize, yshift=-1ex},
            ymajorgrids=true,
            grid style={dotted, ETHGray!50}, % 网格也换用稍微柔和的 ETHGray
            clip=false,
            set layers,
            axis on top=false,
            legend style={
                at={(0.5,0.90)},
                anchor=south,
                legend columns=3,
                draw=none,
                fill=none,
                font=\tiny,
                on layer=axis background
            },
        ]

        % =========================================================
        % 1. LoRA (ETHGreen)
        % mark options={solid} 修复虚线描边导致的畸变问题
        % =========================================================
        \addplot[color=ETHGreen, mark=triangle*, mark size=2.5pt, mark options={solid}, dashed, line width=1pt] 
            coordinates {(1, 2.0029) (2, 2.0996)}
            % S1: 靠右上方 (south west)，文字内部换行
            node[pos=0, anchor=south west, align=center, font=\scriptsize, text=ETHGreen] {$\LORA \Mspec$\\(2.00)}
            % S2: 靠左下方 (north east)，刚好和 FFT 错开
            node[pos=1, anchor=south west, align=center, font=\scriptsize, text=ETHGreen] {$\Mdist$\\(2.11)};
        \addlegendentry{$\LORA$}

        % =========================================================
        % 2. FFT (ETHBlue)
        % =========================================================
        \addplot[color=ETHBlue, mark=square*, mark size=2pt, mark options={solid}, dashed, line width=1pt] 
            coordinates {(1, 3.0490) (2, 4.3971)}
            % S1: 靠右下方 (north west)
            node[pos=0, anchor=south west, align=center, font=\scriptsize, text=ETHBlue] {$\FFT \Mspec$\\(3.05)}
            % S2: 靠左上方 (south west)，完美避开和 LoRA(4.27) 的重叠
            node[pos=1, anchor=north west, align=center, font=\scriptsize, text=ETHBlue] {$\Mdist$\\(4.40)};
        \addlegendentry{$\FFT$}

        % =========================================================
        % 3. LST (ETHRed, Ours)
        % =========================================================
        \addplot[color=ETHRed, mark=*, mark size=2.5pt, mark options={solid}, dashed, line width=1.5pt] 
            coordinates {(1, 1.4304) (2, 2.1143)}
            % S1 靠右下，S2 靠左上，文字居中加粗
            node[pos=0, anchor=north west, align=center, font=\scriptsize\bfseries, text=ETHRed] {$\LST \Mspec$\\(1.43)}
            node[pos=1, anchor=north west, align=center, font=\scriptsize\bfseries, text=ETHRed] {$\Mdist$\\(2.11)};
        \addlegendentry{$\LST$}

        % =========================================================
        % 4. Self-Distill Baseline (ETHPurple)
        % 为了渲染更规整，将 star 改为了 diamond* (菱形)，或者 pentagon*
        % =========================================================
        \addplot[color=ETHPurple, mark=diamond*, mark size=3pt, mark options={solid}, only marks] 
            coordinates {(2, 0.1063)}
            % 靠左上方，保证不超出图表右侧边缘
            node[pos=0, anchor=south west, align=center, font=\scriptsize, text=ETHPurple] {Self-Distill \\ $\Mdist$ (0.11)};
        \addlegendentry{Self-Distill}
        % 辅助基准虚线
        \draw[ETHPurple!40, dotted, line width=0.8pt] (axis cs:0.8,0.1063) -- (axis cs:2.2,0.1063);

        % =========================================================
        % 5. Self-Rationalize Baseline (ETHPurple)
        % 为了渲染更规整，将 star 改为了 diamond* (菱形)，或者 pentagon*
        % =========================================================
        \addplot[color=ETHBronze, mark=o, mark size=3pt, mark options={solid}, only marks] 
            coordinates {(2, 0.1063)}
            % 靠左上方，保证不超出图表右侧边缘
            node[pos=0, anchor=south east, align=center, font=\scriptsize, text=ETHBronze] {Self-Rationalize \\ $\Mdist$ (0.10)};
        \addlegendentry{Self-Rationalize}
        % 辅助基准虚线
        \draw[ETHBronze!40, dotted, line width=0.8pt] (axis cs:0.8,0.0960) -- (axis cs:2.2,0.0960);
        
        \end{axis}
    \end{tikzpicture}
    \caption{}
    \label{fig:kl_divergence}
    \end{subfigure}

\captionsetup{skip=-13pt}
    \caption{
\textbf{Left (a)}: Data scaling on \SMol using rationales collected from the $\LST$\ \textit{Specialist} $\Mspec$ and $\FFT$-distillation. Increasing the amount of filtered training data improves \ID performance substantially, while leaving \SID largely stable and only minimal degradation on \OOD.
\textbf{Right (b)}: KL divergence of \chem $\Mspec$ and $\Mdist$ models w.r.t $\Mzero$ \textsf{Qwen3-8B}. $\LST$ models exhibit lower KL divergence compared to $\FFT$ / $\LORA$ counterparts. 
% The two self-training $\Mdist$ models serves as baselines.
}
\vspace{-10pt}
 \label{fig:data_scaling_and_kl}
\end{figure*}
We study data scaling on the \chem \SMol using rationales collected from the $\LST$\ \textit{Specialist} $\Mspec$, followed by $\FFT$ distillation. \SMol contains roughly 3M examples in total; after generating $(\hat{\tau},\hat{a})$ pairs and filtering them by answer correctness and completeness of the reasoning traces, we obtain a curated pool of 360K training examples. We scale the training set from 50K to the full 360K in 50K increments (\cref{fig:data_scaling}). Increasing the amount of filtered distillation data leads to clear gains on \ID (from $19.3$ to $32.5$), while \SID remains virtually unchanged and \OOD degrades only modestly overall. These results suggest that distillation from $\Mspec$ rationales has not saturated with a moderate amount of data and exhibits favorable scaling behavior with additional high-quality data.

\subsection{Students inherit their teachers' generalization profiles across model families.}
We evaluate \textit{cross-model transfer} by distilling rationales from \textsf{Intern-S1-mini} \citep[\textsf{S1m};][]{internS1}, a science-specialized 8B model, into a \textsf{Qwen3-8B} student (\cref{tab:cross_model}). Crucially, the distilled student strictly inherits the behavioral profile of its specific teacher. While direct distillation from \textsf{S1m} already improves the student's \ID substantially ($19.34 \to 31.33$), distilling from the $\FFT$-\textsf{S1m}-specialist maximizes student \ID but severely degrades \SID/\OOD, exactly mirroring the $\FFT$-teacher's extreme overfitting. Conversely, an $\LST$-\textsf{S1m}-specialist imparts a balanced profile: the student preserves robustness comparable to \textsf{S1m} while still delivering a sizable \ID gain ($+13.38$) from \textsf{Qwen3-8B}. These results confirm that structural constraints ($\LST$) versus unconstrained tuning ($\FFT$) fundamentally shape the rationale distributions, dictating downstream generalization even across model families. Beyond cross-family transfer, \appref{app:sup_analysis:cross_scale} shows that our 8B specialist actually imparts better performance to an 8B student than a larger 14B teacher, hence specialist supervision surpasses the model scale.
% We further show in \appref{app:sup_analysis:cross_scale} that our 8B specialist outperforms cross-scale distillation from \textsf{Qwen3-14B}, proving that a larger teacher alone cannot substitute for our specialist supervision.

\subsection{Post-training strategies induce different degrees of KL drift.}

To characterize how specialist post-training changes model behavior with respect to the origin model, we measure the token-level $\mathrm{KL}(\pi_{\bullet} \klgiven \pi_{\theta_0}), \forall \pi_{\bullet} \in \{\pi_{\theta_1}, \pi_{\theta'_0}\}$, across different tuning strategies on \chem (\cref{fig:kl_divergence}). We use KL divergence as a behavioral diagnostic of distributional drift rather than as the explicit tuning objective. $\FFT$ exhibits the largest divergence from the origin model for both the specialist $\pi_{\theta_1}$ and its distilled student $\pi_{\theta'_0}$. $\LORA$ produces substantially smaller drift, while $\LST$ yields the lowest KL divergence among the adapted models evaluated. These results show an empirical association between stronger constraints on specialist adaptation and reduced behavioral drift from the origin model. In particular, $\LST$ is consistent with an \emph{implicit anchoring effect}: although it contains no explicit KL regularization, its restricted update space yields a specialist that remains closer to the origin-model distribution while acquiring target-domain capability. Importantly, we do not claim that $\LST$ optimizes a KL-constrained objective, nor that a particular layer-selection configuration corresponds to a specific KL radius. The lower-drift profile induced by $\LST$ is also reflected in its distilled student, motivating the question of whether explicitly controlling KL drift can systematically move the specialization--generalization operating point. We test this directly in \cref{sec:analysis:asft}.

% \subsection{Trajectory control is reflected in KL divergence drift.}
% To empirically validate our theoretical framework (\cref{sec:method}), we analyze the trajectory distribution shift by measuring the token-level $\KL(\pi_{\bullet} \parallel \Mzero), \forall \pi_{\bullet} \in \{ \Mspec, \Mdist\}$, across different optimization strategies on \chem (\cref{fig:kl_divergence}). Unconstrained $\FFT$ exhibits the most severe distribution drift for both the specialist $\Mspec$ and distilled $\Mdist$ models, confirming that standard QA-only supervision arbitrarily overfits the underdetermined trajectory space. In contrast, parameter-efficient architectures impose varying degrees of implicit regularization: $\LORA$ noticeably restricts this divergence, while $\LST$ framework achieves the tightest KL control among the adapted models. This perfectly corroborates our analysis in \cref{sec:traj_learnable}: the structural constraints of $\LST$ implicitly act as the KL-divergence penalty, guiding the optimization to select a trajectory distribution $\pi_{\theta_1}$ that solves the task without destructively departing from $\Mzero$'s intrinsic generation manifold. Consequently, committing to this well-anchored and smoothly reweighted trajectory distribution, $\LST$ enables a fundamentally more stable specialist-distillation, implying that trajectory drift control is a vital factor of downstream behavioral retention.

\subsection{Explicit KL Drift Control with ASFT} \label{sec:analysis:asft}
\begin{figure*}[t]
    \centering

    % ============================================================
    % Shared legend for both subfigures
    % ============================================================
    \makebox[\textwidth][c]{%
        \scriptsize
        % IT
        \tikz[baseline=-0.55ex]{
            \draw[ETHRed, line width=1.5pt] (0,0)--(0.42,0);
        }
        \hspace{0.15em}\ID
        \hspace{1.4em}
        % ID
        \tikz[baseline=-0.55ex]{
            \draw[ETHBlue, line width=1.5pt] (0,0)--(0.42,0);
        }
        \hspace{0.15em}\SID
        \hspace{1.4em}
        % OOD
        \tikz[baseline=-0.55ex]{
            \draw[ETHGreen, line width=1.5pt] (0,0)--(0.42,0);
        }
        \hspace{0.15em}\OOD
        \hspace{2.0em}
        % Specialist
        \tikz[baseline=-0.55ex]{
            \draw[black, line width=0.9pt]
                plot[mark=triangle*, mark size=2.1pt, mark options={solid}]
                coordinates {(0,0) (0.42,0)};
        }
        \hspace{0.15em}\textit{Specialist}
        \hspace{2.0em}
        % Distilled
        \tikz[baseline=-0.55ex]{
            \draw[black, dashed, line width=0.9pt]
                plot[mark=*, mark size=1.8pt, mark options={solid}]
                coordinates {(0,0) (0.42,0)};
        }
        \hspace{0.15em}$\FFT$ \textit{Distilled}
    }

    \vspace{-4pt}

    % ============================================================
    % Chemistry
    % ============================================================
    \begin{subfigure}[t]{0.49\textwidth}
        \centering
        \begin{tikzpicture}
        \begin{axis}[
            width=0.97\linewidth,
            height=4.75cm,
            xmin=0.55, xmax=4.45,
            ymin=20, ymax=62,
            ytick={20,30,40,50,60},
            xtick={1,2,3,4.10},
            xticklabels={
                {\shortstack{$\ASFT$\\$\lambda=0.05$}},
                {\shortstack{$\ASFT$\\$\lambda=0.2$}},
                {\shortstack{$\ASFT$\\$\lambda=0.5$}},
                {\shortstack{$\LST$\\\textit{ref.}}}
            },
            title={Chemistry},
            title style={font=\small\bfseries, yshift=-9pt},
            tick label style={font=\tiny},
            xticklabel style={
                font=\scriptsize,
                align=center
            },
            ylabel={\scriptsize Score},
            ylabel near ticks,
            ymajorgrids=true,
            xmajorgrids=false,
            grid style={dashed, ETHGray!35, line width=0.4pt},
            axis line style={line width=0.6pt},
            tick style={line width=0.5pt},
            clip=false,
            axis on top,
        ]

        % ------------------------------------------------------------
        % LST reference band
        % ------------------------------------------------------------
        \path[fill=black!4, draw=none]
            (axis cs:3.72,20)
            rectangle
            (axis cs:4.45,62);

        \draw[ETHGray!60, dashed, line width=0.6pt]
            (axis cs:3.65,20) --
            (axis cs:3.65,62);

        % ============================================================
        % ASFT: IT
        % ============================================================
        \addplot+[
            color=ETHRed,
            line width=1.1pt,
            mark=triangle*,
            mark size=2.2pt,
            mark options={solid}
        ]
        coordinates {
            (1,39.43)
            (2,32.83)
            (3,22.42)
        };

        \addplot+[
            color=ETHRed,
            dashed,
            line width=1.1pt,
            mark=*,
            mark size=1.9pt,
            mark options={solid}
        ]
        coordinates {
            (1,37.63)
            (2,30.27)
            (3,24.74)
        };

        % ============================================================
        % ASFT: ID
        % ============================================================
        \addplot+[
            color=ETHBlue,
            line width=1.1pt,
            mark=triangle*,
            mark size=2.2pt,
            mark options={solid}
        ]
        coordinates {
            (1,55.53)
            (2,55.56)
            (3,57.36)
        };

        \addplot+[
            color=ETHBlue,
            dashed,
            line width=1.1pt,
            mark=*,
            mark size=1.9pt,
            mark options={solid}
        ]
        coordinates {
            (1,55.40)
            (2,56.90)
            (3,58.11)
        };

        % ============================================================
        % ASFT: OOD
        % ============================================================
        \addplot+[
            color=ETHGreen,
            line width=1.1pt,
            mark=triangle*,
            mark size=2.2pt,
            mark options={solid}
        ]
        coordinates {
            (1,29.97)
            (2,33.98)
            (3,36.10)
        };

        \addplot+[
            color=ETHGreen,
            dashed,
            line width=1.1pt,
            mark=*,
            mark size=1.9pt,
            mark options={solid}
        ]
        coordinates {
            (1,28.55)
            (2,33.50)
            (3,36.24)
        };

        % ============================================================
        % LST reference points
        % ============================================================
        % IT
        \addplot+[
            color=ETHRed,
            only marks,
            mark=triangle*,
            mark size=2.7pt,
            mark options={solid}
        ]
        coordinates {(4.00,29.61)};

        \addplot+[
            color=ETHRed,
            only marks,
            mark=*,
            mark size=2.4pt,
            mark options={solid}
        ]
        coordinates {(4.20,29.54)};

        % ID
        \addplot+[
            color=ETHBlue,
            only marks,
            mark=triangle*,
            mark size=2.7pt,
            mark options={solid}
        ]
        coordinates {(4.00,60.20)};

        \addplot+[
            color=ETHBlue,
            only marks,
            mark=*,
            mark size=2.4pt,
            mark options={solid}
        ]
        coordinates {(4.20,60.47)};

        % OOD
        \addplot+[
            color=ETHGreen,
            only marks,
            mark=triangle*,
            mark size=2.7pt,
            mark options={solid}
        ]
        coordinates {(4.00,44.03)};

        \addplot+[
            color=ETHGreen,
            only marks,
            mark=*,
            mark size=2.4pt,
            mark options={solid}
        ]
        coordinates {(4.20,42.11)};
        \end{axis}
        \end{tikzpicture}
        \label{fig:asft:chem}
    \end{subfigure}
    \hfill
    % ============================================================
    % Multilingualism
    % ============================================================
    \begin{subfigure}[t]{0.49\textwidth}
        \centering
        \begin{tikzpicture}
        \begin{axis}[
            width=0.97\linewidth,
            height=4.75cm,
            xmin=0.55, xmax=4.45,
            ymin=0, ymax=50,
            ytick={0,10,20,30,40,50},
            xtick={1,2,3,4.10},
            xticklabels={
                {\shortstack{$\ASFT$\\$\lambda=0.05$}},
                {\shortstack{$\ASFT$\\$\lambda=0.2$}},
                {\shortstack{$\ASFT$\\$\lambda=0.5$}},
                {\shortstack{$\LST$\\\textit{ref.}}}
            },
            title={Multilingualism},
            title style={font=\small\bfseries, yshift=-9pt},
            tick label style={font=\tiny},
            xticklabel style={
                font=\scriptsize,
                align=center
            },
            ylabel={\scriptsize Score},
            ylabel near ticks,
            ymajorgrids=true,
            xmajorgrids=false,
            grid style={dashed, ETHGray!35, line width=0.4pt},
            axis line style={line width=0.6pt},
            tick style={line width=0.5pt},
            clip=false,
            axis on top,
        ]

        % ------------------------------------------------------------
        % LST reference band
        % ------------------------------------------------------------
        \path[fill=black!4, draw=none]
            (axis cs:3.72,0)
            rectangle
            (axis cs:4.45,50);

        \draw[ETHGray!60, dashed, line width=0.6pt]
            (axis cs:3.65,0) --
            (axis cs:3.65,50);

        % ============================================================
        % ASFT: IT
        % ============================================================
        \addplot+[
            color=ETHRed,
            line width=1.1pt,
            mark=triangle*,
            mark size=2.2pt,
            mark options={solid}
        ]
        coordinates {
            (1,30.47)
            (2,30.26)
            (3,30.74)
        };

        \addplot+[
            color=ETHRed,
            dashed,
            line width=1.1pt,
            mark=*,
            mark size=1.9pt,
            mark options={solid}
        ]
        coordinates {
            (1,30.47)
            (2,29.27)
            (3,30.74)
        };

        % ============================================================
        % ASFT: ID
        % ============================================================
        \addplot+[
            color=ETHBlue,
            line width=1.1pt,
            mark=triangle*,
            mark size=2.2pt,
            mark options={solid}
        ]
        coordinates {
            (1,26.65)
            (2,36.48)
            (3,45.42)
        };

        \addplot+[
            color=ETHBlue,
            dashed,
            line width=1.1pt,
            mark=*,
            mark size=1.9pt,
            mark options={solid}
        ]
        coordinates {
            (1,9.15)
            (2,44.98)
            (3,44.79)
        };

        % ============================================================
        % ASFT: OOD
        % ============================================================
        \addplot+[
            color=ETHGreen,
            line width=1.1pt,
            mark=triangle*,
            mark size=2.2pt,
            mark options={solid}
        ]
        coordinates {
            (1,8.70)
            (2,20.14)
            (3,32.99)
        };

        \addplot+[
            color=ETHGreen,
            dashed,
            line width=1.1pt,
            mark=*,
            mark size=1.9pt,
            mark options={solid}
        ]
        coordinates {
            (1,0.51)
            (2,42.94)
            (3,43.71)
        };

        % ============================================================
        % LST reference points
        % ============================================================
        % IT
        \addplot+[
            color=ETHRed,
            only marks,
            mark=triangle*,
            mark size=2.7pt,
            mark options={solid}
        ]
        coordinates {(4.00,35.68)};

        \addplot+[
            color=ETHRed,
            only marks,
            mark=*,
            mark size=2.4pt,
            mark options={solid}
        ]
        coordinates {(4.20,34.17)};

        % ID
        \addplot+[
            color=ETHBlue,
            only marks,
            mark=triangle*,
            mark size=2.7pt,
            mark options={solid}
        ]
        coordinates {(4.00,43.99)};

        \addplot+[
            color=ETHBlue,
            only marks,
            mark=*,
            mark size=2.4pt,
            mark options={solid}
        ]
        coordinates {(4.20,43.87)};

        % OOD
        \addplot+[
            color=ETHGreen,
            only marks,
            mark=triangle*,
            mark size=2.7pt,
            mark options={solid}
        ]
        coordinates {(4.00,39.08)};

        \addplot+[
            color=ETHGreen,
            only marks,
            mark=*,
            mark size=2.4pt,
            mark options={solid}
        ]
        coordinates {(4.20,41.23)};

        \end{axis}
        \label{fig:asft:multiling}
        \end{tikzpicture}
    \end{subfigure}

    \vspace{-8pt}

    \caption{
    \textbf{Explicit KL anchoring with ASFT.}
    Increasing the KL coefficient $\lambda$ strengthens anchoring to the origin model, shifting both specialists and their distilled students from stronger task specialization toward better \SID/\OOD retention.
    The first three $x$-axis settings correspond to $\ASFT$ with $\lambda\in\{0.05,0.2,0.5\}$.
    The shaded $\LST$ \textit{ref.} column shows the corresponding $\LST$ specialist and distilled student as an implicit drift-control reference (\cref{tab:main}) rather than an $\ASFT$ setting. \textbf{Left (a)}: Chemistry. \textbf{Right (b)}: Multilingualism.
    }
    \label{fig:asft}
    % \vspace{-15pt}
\end{figure*}
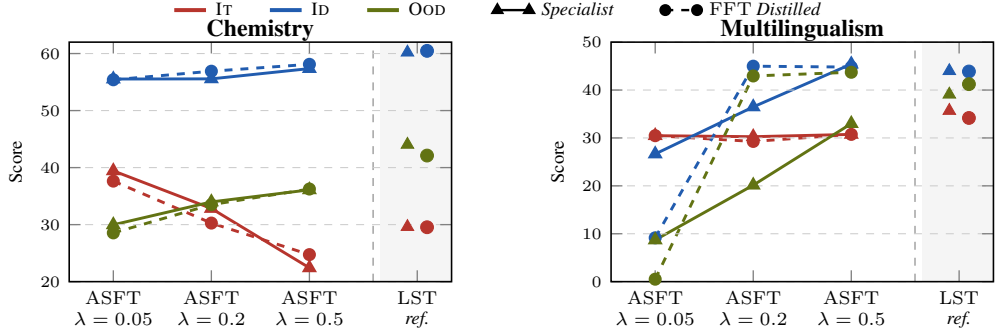

Our analysis suggests that the specialization--generalization trade-off is governed more broadly by how strongly specialist optimization controls distributional drift, rather than by $\LST$ specifically. To test this interpretation beyond structural constraints, we additionally evaluate Anchored Supervised Fine-Tuning \cite[$\ASFT$;][]{asft}, which explicitly regularizes the specialist toward the origin model through a KL penalty. We repeat the specialist generation and $\FFT$-distillation pipeline with the coefficient $\lambda\in\{0.05,0.2,0.5\}$, where larger $\lambda$ imposes stronger anchoring.

As shown in \cref{fig:asft}, increasing the anchoring strength systematically shifts both specialists and their distilled students from stronger task specialization toward better $\SID / \OOD$ retention. In Chemistry, increasing $\alpha$ from $0.05$ to $0.5$ reduces specialist $\ID$ performance from $39.43$ to $22.42$, while improving $\OOD$ performance from $29.97$ to $36.10$; the corresponding distilled models exhibit the same shift, from $37.63/28.55$ to $24.74/36.24$ in $\ID / \OOD$. The trend is even more pronounced in Multilingualism, where specialist $\SID / \OOD$ increases from $26.65/8.70$ to $45.42/32.99$, accompanied by substantial gains in the distilled models from $9.15/0.51$ to $44.79/43.71$. These results show that the observed trade-off is not unique to $\LST$: explicit KL regularization provides a complementary and tunable mechanism for controlling specialist drift, and the resulting specialization--generalization profile is subsequently inherited through distillation. $\LST$ should therefore be viewed as one practical implicit realization of this broader drift-control principle, rather than as a uniquely optimal tuning strategy.

\subsection{Structural constraint improves answer prediction without collapsing the trajectory.} 
\label{sec:analysis:hr}

\begin{figure}[t]
    \centering

    %---------------- left: figure ----------------
    \begin{minipage}[t]{0.48\linewidth}
        \centering
        \vspace{0pt}

        \begin{tikzpicture}
        \pgfplotsset{
            every axis/.style={
                width=\linewidth,
                height=4.7cm,
                enlarge x limits=0.18,
                symbolic x coords={Baseline, LST, FFT, LoRA},
                xticklabels={Baseline, $\LST$, $\FFT$, $\LORA$},
                xtick=data,
                xtick pos=bottom,
                ytick pos=left,
                ybar=2pt,
                bar width=12pt,
                ymajorgrids=true,
                grid style={dashed, gray!30},
                tick label style={font=\tiny},
                label style={font=\footnotesize},
                x tick label style={font=\footnotesize, align=center, rotate=0},
                nodes near coords,
                nodes near coords style={font=\footnotesize, anchor=south},
            },
        }
        \begin{axis}[
            name=hr,
            ylabel={Top-1 Hit Rate ($\HR$) $\uparrow$},
            ymin=0.55, ymax=1.05,
            ytick={0.6,0.7,0.8,0.9,1.0},
            nodes near coords={\pgfplotspointmeta},
            point meta=explicit symbolic,
        ]
        \addplot[fill=ETHPurple!35, draw=ETHPurple!70, bar shift=-7pt]
            coordinates {(Baseline, 0.69) [.69] (LST, 0.85) [.85] (FFT, 0.94) [.94] (LoRA, 0.86) [.86]};
        \addplot[fill=ETHPurple, draw=ETHPurple!90, bar shift=7pt]
            coordinates {(Baseline, 0.69) [.69] (LST, 0.72) [.72] (FFT, 0.83) [.83] (LoRA, 0.74) [.74]};

        \addplot[
            fill=none, draw=ETHGray!70, bar shift=-7pt,
            postaction={pattern=north east lines, pattern color=ETHGray!80},
            nodes near coords=\empty, forget plot,
        ] coordinates {(Baseline, 0.69)};
        \addplot[
            fill=none, draw=ETHGray!70, bar shift=7pt,
            postaction={pattern=north east lines, pattern color=ETHGray!80},
            nodes near coords=\empty, forget plot,
        ] coordinates {(Baseline, 0.69)};
        \end{axis}

        \coordinate (legendTop) at ($(hr.north) + (0, 0.5cm)$);
        \fill[ETHPurple!35, draw=ETHPurple!70] ($(legendTop) + (-3.4, 0)$) rectangle ++(0.25, 0.25);
        \node[anchor=west, font=\small] at ($(legendTop) + (-3.1, 0.125)$) {\footnotesize Specialist $\Mspec$};
        \fill[ETHPurple, draw=ETHPurple!90] ($(legendTop) + (0.4, 0)$) rectangle ++(0.25, 0.25);
        \node[anchor=west, font=\small] at ($(legendTop) + (0.7, 0.125)$) {\footnotesize Distilled $\Mdist$};

        \coordinate (legendBot) at ($(legendTop) + (0, -0.3cm)$);
        \fill[ETHGray!30, draw=ETHGray!70, postaction={pattern=north east lines, pattern color=ETHGray!80}] ($(legendBot) + (-3.4, 0)$) rectangle ++(0.25, 0.25);
        \node[anchor=west, font=\small] at ($(legendBot) + (-3.1, 0.0)$) {\footnotesize \textsf{Qwen3-8B} $\Mzero$ \& Self-Distill $\Mdist$ Baselines};
        \end{tikzpicture}
\captionsetup{skip=0pt}
        \caption{$\HR$ under teacher-forcing on unseen \chem \SMol queries. Higher ($\uparrow$) is better. Hatched are the origin model $\Mzero$ and \textit{self-distill} baselines.}
        \label{fig:hr}
    \end{minipage}
    \hfill
    %---------------- right: table ----------------
    \begin{minipage}[t]{0.5\linewidth}
        \centering
        \vspace{0pt}
        \captionsetup{type=table}

        \footnotesize
        \renewcommand{\arraystretch}{1.1}
        \setlength{\tabcolsep}{2pt}
\captionsetup{skip=0pt}
        \caption{\textbf{Rationale quality on Chemistry}. Statistics over $150$K generated trajectories from $50$K \SMol queries with $3$ samples each, covering two \textit{self-training} modes of $\Mzero$ and three specialists ($\FFT$, $\LORA$, and $\LST$). ``Comp.'' is the percentage of structurally \emph{complete} trajectories. ``Empty'' is the permille of empty rationales among \emph{complete} trajectories. ``Words'' denotes average word count $\pm$ std within \textit{complete} rationales.}
        \label{tab:rationale_quality}

        % \begin{tabular}{@{}lccl@{}}
        %     \toprule
        %     \textbf{Model Generation} & \textbf{Comp}($\%$) & \textbf{Empty}(\textperthousand) & \textbf{Words($\mu \pm \sigma$)} \\
        %     \midrule
        %     \rowcolor{baselineorange}
        %     $\Mzero$ $q \rightarrow \hat{\tau},\hat{a}$ & 87.93 & 0.00 & $2476 \pm 1624$ \\
        %     \rowcolor{baselineorange}
        %     $\Mzero$ $q,a^* \rightarrow \hat{\tau}$ & 16.18 & 0.08 & $2049 \pm 1443$ \\
        %     \midrule
        %     $\FFT$ Spec. $\Mspec$  & 8.90 & 1.10 & $233 \pm 450$ \\
        %     $\LORA$\ Spec. $\Mspec$ & 41.39 & 0.00 & $1378 \pm 902$ \\
        %     $\LST$\ Spec. $\Mspec$ & 92.50 & 0.02 & $2334 \pm 1435$ \\
        %     \bottomrule
        % \end{tabular}

\begin{tabular}{@{}l l c c c@{}}
    \toprule
    \multicolumn{2}{l}{\textbf{Model Generation}} & \textbf{Comp.}(\%) & \textbf{Empty}(\textperthousand) & \textbf{Words}($\mu \pm \sigma$) \\
    \midrule

    \rowcolor{baselineorange}
    \multicolumn{2}{l}{$\Mzero$ $q \rightarrow \hat{\tau},\hat{a}$} & 87.93 & 0.00 & $2476 \pm 1624$ \\

    \rowcolor{baselineorange}
    \multicolumn{2}{l}{$\Mzero$ $q,a^* \rightarrow \hat{\tau}$} & 16.18 & 0.08 & $2049 \pm 1443$ \\

    \cmidrule{1-5}

    \multirow{3}{*}{%
      \begingroup
      \setlength{\fboxsep}{0pt}%
      \colorbox{originblue}{%
        \makebox[0.3cm][c]{%
          \rule[-2.em]{0pt}{2.em}%
          \rotatebox[origin=c]{90}{\tiny $q \to \hat{\tau}, \hat{a}$}%
        }%
      }%
      \endgroup
    }
    & $\FFT$ Spec. $\Mspec$ & 8.90 & 1.10 & $233 \pm 450$ \\
    & $\LORA$ Spec. $\Mspec$ & 41.39 & 0.00 & $1378 \pm 902$ \\
    & $\LST$ Spec. $\Mspec$ & 92.50 & 0.02 & $2334 \pm 1435$ \\
    \bottomrule
\end{tabular}

    \end{minipage}
    \vspace{-7pt}
\end{figure}
To analyze how different fine-tuning strategies distribute probability mass over valid trajectories (\cref{sec:extraction}), we evaluate the next-token top-1 hit rate ($\HR$) via \textit{teacher-forcing} on $2,000$ unseen \SMol samples (\appref{app:bench_details:chem:smol:data_scale}). Suppressing intermediate CoT steps isolates the intrinsic capability to sequentially predict ground-truth answers. While all tuning methods substantially improve $\HR$ over the baselines (\cref{fig:hr}), unconstrained $\FFT$ induces extreme confidence in the specialist ($\HR = 0.94$), indicating a drastic collapse of the trajectory distribution to overfit the answer likelihood (Eq.~\ref{eq:qa-loss}). In contrast, parameter-efficient architectures ($\LST$ and $\LORA$) provide calibrated enhancements ($\HR \approx 0.85$). Aligning with our theoretical framework (\cref{sec:traj_learnable}), these structural constraints enforce a smoother reweighting of the base policy (Eq.~\ref{eq:kl_reweight}) rather than arbitrarily distorting the generation manifold. By preventing rote memorization, they facilitate robust and generalizable transfer to the distilled students. Next-token probability and rank exhibit consistent trends (\appref{app:prob_rank}).

\subsection{Structural constraint preserves the integrity of reasoning trajectories.} \label{sec:analysis:traj_quality}
To assess rationale quality, we analyze $150$K Chemistry trajectories ($3$ candidates per $50$K \SMol queries). We define a ``\textit{complete}'' trajectory as having exactly one valid \texttt{<think> </think>} pair. As \cref{tab:rationale_quality} shows, unconstrained $\FFT$ suffers severe structural collapse: a dismal $8.90\%$ completion rate, the highest empty rationale proportion ($1.1$\textperthousand, $\tau_{\emptyset}$ as in \cref{sec:intro}), and heavily degenerated trace lengths ($233$ vs. origin's $2476$ words). This indicates a collapse into shortcut reasoning ($\tau_{\sim}$ as in \cref{sec:intro}), explaining its poor-quality distillation supervision. $\LORA$ partially recovers completion ($41.39\%$), whereas $\LST$--specialist $\Mspec$ robustly preserves both structural integrity ($92.50\%$) and trace lengths comparable to the origin $\Mzero$. Furthermore, the \textit{self-rationalize} baseline ($q, a^* \to \hat{\tau}$) drops to $16.18\%$ completion, largely due to the generation of multiple reasoning blocks. This empirically reinforces our claim in \cref{sec:exp:results:qa_improvement} that forcing weak models into post-hoc rationalization fundamentally destabilizes the generation manifold and acts as toxic supervision.
\section{Conclusion}
\label{sec:conclusion}

In this work, we frame QA-only specialist distillation as a trajectory-distribution selection problem: answer supervision leaves reasoning trajectories underdetermined, while specialist optimization determines the latent supervision passed downstream. Across chemistry, physics, and low-resource multilingual tasks, distilled students closely inherit their specialists' specialization--generalization profiles, including across model families. We further identify distributional drift as a controllable axis of this transfer. Unconstrained $\FFT$ induces larger drift and degraded generalization, whereas $\LST$ provides effective implicit drift control; explicit KL anchoring with $\ASFT$ systematically moves both specialists and students along the same trade-off. These results establish specialist optimization as a key design variable for distillation data, enabling more predictable control over domain specialization and general-capability retention when gold trajectories are unavailable.

\newpage
\bibliographystyle{acl_natbib}
\bibliography{reference}

%%%%%%%%%%%%%%%%%%%%%%%%%%%%%%%%%%%%%%%%%%%%%%%%%%%%%%%%%%%%

\clearpage
\appendix
\section{Limitations} \label{app:lim}

% While our approach demonstrates promising results, we acknowledge several limitations in our current work. Primarily, due to constraints in computational resources, we were unable to scale our empirical experiments across a broader range of domains, larger-scale foundation models, or alternative fine-tuning paradigms. To mitigate this limitation and ensure the generalizability of our claims, we made our best effort to carefully select the most representative and widely-adopted models, domains, and tuning strategies for our evaluations. As with many empirical studies, our results may implicitly depend on specific data distribution assumptions. The robustness of our approach when these assumptions are violated, such as in highly noisy real-world settings or under severe distribution shifts, remains to be fully explored. Additionally, while our method is efficient in its current scope, its computational scaling behavior on massive, web-scale datasets requires further investigation. Future work equipped with more abundant compute should focus on stress-testing the approach across more diverse tasks, while also exploring potential fairness and privacy implications when adapting these models to sensitive applications.

While our framework provides a novel perspective on QA-only distillation, it entails several limitations that present opportunities for future work. 
(1) \textbf{Scale and Emergent Reasoning:} Our primary experiments focus on the 8B parameter scale to rigorously isolate variables. How implicit trajectory selection interacts with the self-correction capabilities of massive-scale models (e.g., $>70$B) or explicitly reasoning-optimized models remains an open question. 
(2) \textbf{Spurious Correlations:} Our rationale filtration strictly relies on exact answer equivalence. While highly effective in deterministic domains like physics and chemistry, it cannot entirely eradicate the ``right answer, wrong reasoning'' phenomenon (spurious shortcuts), particularly in open-ended generative tasks like multilingual translation.
\section{KL constraints enable controlled reweighting.}
\label{app:kl_reweighting}

To prevent uncontrolled trajectory shift, we formulate learning as a drift-constrained optimization problem:
\begin{equation}
\max_{\pi}
\quad
\mathbb{E}_{(q,a)}[\log \pi(a \given q)]
\qquad
\text{s.t.}
\qquad
\mathbb{E}_{q}\left[
\mathrm{KL}\bigl(\pi(\cdot\given q)\klgiven \pi_{\theta_0}(\cdot\given q)\bigr)
\right] \le \delta,
\label{eq:drift_problem}
\end{equation}
where $\delta > 0$ controls the average allowed deviation from the base policy $\pi_{\theta_0}$.

The objective is defined at the answer level:
\begin{equation}
    \log \pi(a \given q)=\log \sum_{\tau\to a}\pi(\tau\given q),
\end{equation}
\
which depends on a marginalization over latent trajectories and does not explicitly specify how probability mass is distributed across trajectories.

To make this structure explicit, we introduce a trajectory-level view.
For any distribution $p(\tau\given q,a)$ supported on trajectories satisfying $\tau \to a$, Jensen's inequality gives
\begin{equation}
\log \pi(a\given q)
\ge
\mathbb{E}_{\tau\sim p(\cdot\given q,a)}
\left[
\log \pi(\tau\given q)-\log p(\tau\given q,a)
\right].
\end{equation}

This lower bound shows that improving $\log \pi(a\given q)$ corresponds to increasing probability mass on trajectories that produce the correct answer.

Under this trajectory-level perspective, the KL constraint restricts how much the trajectory distribution can deviate from the base policy.
As a result, updates are realized through a controlled reweighting of trajectories, favoring those that support the correct answer while remaining close to $\pi_{\theta_0}$.

Introducing a Lagrange multiplier $\lambda > 0$ for the KL constraint in~\cref{eq:drift_problem}, the resulting solution takes the form
\begin{equation}
\pi(\tau\given q)
\propto
\pi_{\theta_0}(\tau\given q)\exp\left(\frac{w(\tau,q,a)}{\lambda}\right),
\end{equation}
where $w(\tau,q,a)$ is an implicit advantage-like quantity induced by the objective, reflecting the relative contribution of trajectory $\tau$ to increasing $\log \pi(a \given q)$.

Thus, KL-constrained updates achieve improvement through controlled reweighting of trajectories, rather than unconstrained redistribution of probability mass.
\section{Supplementary Analysis}\label{app:sup_analysis}

\subsection{Ablation Study on LoRA Configurations} \label{app:sup_analysis:lora}
We also analyze the effect of LoRA-specific design choices in the \MT setting. 
All experiments are evaluated on the same \ID, \SID, and \OOD benchmark groups as in the main experiments.

\paragraph{Rank sensitivity.}
\cref{tab:lora_rank_sweep_lrm} reports the \MT performance of LoRA
Specialist models trained with ranks $r \in \{16,32,64,128\}$.
The results show that LoRA performance is relatively stable for ranks 16, 32, and 64 on \SID and \OOD evaluation, while increasing the rank to 128 leads to a noticeable \OOD drop. 
Rank 32 obtains the highest \ID score, whereas rank 64 obtains the best \SID score and remains nearly tied with ranks 16 and 32 on \OOD. 
We therefore use rank 64 in the main experiments as a capacity-balanced
default rather than tuning the rank to maximize a single \ID score.

\begin{table*}[!htbp]
\centering
\caption{Rank sensitivity of LoRA Specialist $\pi_{\theta_1}$ on \MT using 7M \MT QA examples.}
\label{tab:lora_rank_sweep_lrm}
% \caption{Rank sensitivity of LoRA Specialist $\pi_{\theta_1}$ on \MT using 7M QA examples.}
\small
\begin{tabular}{lccc}
\toprule
\textbf{\chem Model} & \textbf{\ID} & \textbf{\SID} & \textbf{\OOD} \\
\midrule
LoRA Specialist $\pi_{\theta_1}$, $r=16$  & 23.02 & 42.84 & \textbf{41.45} \\
LoRA Specialist $\pi_{\theta_1}$, $r=32$  & \textbf{23.47} & 43.03 & \textbf{41.45} \\
LoRA Specialist $\pi_{\theta_1}$, $r=64$  & 20.47 & \textbf{43.16} & 41.41 \\
LoRA Specialist $\pi_{\theta_1}$, $r=128$ & 21.77 & 42.07 & 38.90 \\
\bottomrule
\end{tabular}
\end{table*}

\paragraph{Standard LoRA vs. RsLoRA.}
We also compare standard LoRA with rank-stabilized LoRA \citep[RsLoRA,][]{rslora} at rank 64.
As shown in \cref{tab:rslora_lora_lrm}, RsLoRA performs poorly at the Specialist stage, especially on \ID and \OOD.
Although FFT distillation from the RsLoRA Specialist partially recovers \ID performance, its \OOD score remains close to zero, indicating that the generated rationales do not provide transferable supervision.
In contrast, standard LoRA yields a much more balanced Specialist model and produces rationales that lead to a substantially stronger FFT Distilled model on \SID and \OOD. 
This suggests that, in our \MT setting, RsLoRA changes the optimization dynamics in a way that is harmful to trajectory selection and downstream distillation quality.

\begin{table*}[!htbp]
\centering
\caption{Comparison between standard LoRA and RsLoRA at rank 64 on \MT.}
\label{tab:rslora_lora_lrm}
\small
\begin{tabular}{lccc}
\toprule
\textbf{Model} & \textbf{\ID} & \textbf{\SID} & \textbf{\OOD} \\
\midrule
RsLoRA Specialist $\pi_{\theta_1}$, $r=64$ & 0.51  & 19.36 & 0.66  \\
LoRA Specialist $\pi_{\theta_1}$, $r=64$   & \textbf{20.47} & \textbf{43.16} & \textbf{41.41} \\
\midrule
FFT Distilled $\pi'_{\theta_0}$ from RsLoRA & \textbf{27.23} & 21.94 & 0.83  \\
FFT Distilled $\pi'_{\theta_0}$ from LoRA   & 22.97 & \textbf{30.87} & \textbf{35.89} \\
\bottomrule
\end{tabular}
\end{table*}

\subsection{Model Scaling: Using \textsf{Qwen3-14B} as Origin Model} \label{app:sup_analysis:model_scaling}
\begin{table}[!htbp]
    \renewcommand{\arraystretch}{1.2}
    \setlength{\tabcolsep}{5pt} % 稍微放宽一点列距，因为附录里通常有足够空间
    % \footnotesize
    \small
    \centering
    \caption{
    \textbf{Model Scaling on Chemistry.}
    Performance of specialist models and their corresponding distilled models using \textsf{Qwen3-14B} as the origin model $\Mzero$.
    }
    \label{tab:scaling_14b}
    \begin{tabular}{ll ccc}
        \toprule
        \multicolumn{2}{l}{\textbf{Models on Chemistry}} & \textbf{\ID$(14)$} & \textbf{\SID$(4)$} & \textbf{\OOD$(5^{\scriptscriptstyle =})$} \\
        \midrule
        \rowcolor{originblue}
        \multicolumn{2}{l}{Origin $\Mzero$ \textsf{Qwen3-14B}}  & 23.80 & 63.94 & 48.95 \\
        \midrule

    \multirow{4}{*}{%
      \begingroup
      \setlength{\fboxsep}{0pt}%
      \colorbox{originblue}{%
        \makebox[0.3cm][c]{%
          \rule[-2.em]{0pt}{2.em}%
          \rotatebox[origin=c]{90}{\scriptsize Tuned on $\Mzero$ \textsf{14B}}%
        }%
      }%
      \endgroup
    }
        & $\FFT$ \textit{Specialist} $\Mspec$  & 59.10 & 37.50 & 37.14 \\
        & $\ \subarrow \FFT$-\textit{Distilled} $\Mdist$  & 59.06 & 27.53 & 3.46 \\
        \cmidrule{2-5}
        & $\LST$ \textit{Specialist} $\Mspec$  & 35.76 & 63.80 & 48.94 \\
        & $\ \subarrow \FFT$-\textit{Distilled} $\Mdist$  & 33.40 & 60.68 & 29.54 \\
        \bottomrule
    \end{tabular}
\end{table}

To investigate whether the phenomena observed in our main experiments hold for larger models, we scale our origin model $\Mzero$ from \textsf{Qwen3-8B} to \textsf{Qwen3-14B} and replicate the QA-only specialist distillation pipeline on the \textbf{Chemistry} domain. We compare the $\FFT$ and $\LST$ tuning strategies. As shown in \cref{tab:scaling_14b}, scaling up the origin model reveals trends that are perfectly consistent with our findings in \cref{sec:exp:results}:

\begin{itemize}[leftmargin=12pt]
    \item \textbf{Consistent Target-Domain Improvement:} Both $\FFT$ and $\LST$ specialists ($\Mspec$), as well as their downstream distilled models ($\Mdist$), significantly outperform the zero-shot \textsf{Qwen3-14B} origin model on the \ID benchmark. This reinforces that QA-only specialist distillation remains highly effective for larger-scale models.
    \item \textbf{Specialization--Generalization Trade-offs:} The distinct trade-off profiles induced by different tuning strategies persist. $\FFT$ achieves extreme \ID gains ($23.80 \to 59.10$) but suffers from severe catastrophic forgetting on \SID and \OOD. Alarmingly, the $\FFT$-distilled model inherits and amplifies this toxicity, plummeting to $3.46$ on \OOD. In contrast, $\LST$ strikes a much healthier balance, providing robust \ID improvements ($23.80 \to 35.76$) while tightly preserving both \SID and \OOD capabilities in the specialist, which translates safely into the distilled model. 
    \item \textbf{Monotonic Rank Correlation:} The relative capabilities of the 14B-based specialists are strictly mirrored in their distilled counterparts. $\FFT$ yields a higher \ID but lower \OOD than $\LST$ in $\Mspec$, and exactly the same ranking applies to $\Mdist$. This further corroborates that the distilled model's behavior is deterministically governed by the trajectory distribution selected by the specialist tuning strategy.
\end{itemize}

\subsection{Cross-scale Teacher--Student Gap} \label{app:sup_analysis:cross_scale}

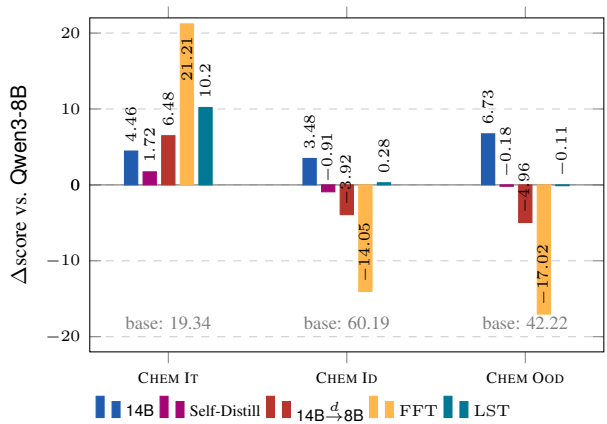
\begin{figure}[ht]
        \centering
        % 你的第二个 tikzpicture
    \begin{tikzpicture}
    \begin{axis}[
        ybar,
        width=0.6\textwidth,
        height=6cm,
        bar width=5pt,
        ymin=-22, ymax=22,
        ylabel={$\Delta$score vs. \textsf{Qwen3-8B}},
        symbolic x coords={ID,SID,OOD},
        xtick=data,
        xticklabels={\chem \ID,\chem \SID,\chem \OOD},
        xticklabel style={font=\tiny},
        yticklabel style={font=\tiny},
        ylabel style={font=\footnotesize},
        enlarge x limits=0.22,
        grid=major,
        ymajorgrids=true,
        xmajorgrids=false,
        grid style={dashed, gray!40},
        axis on top,
        extra y ticks={0},
        extra y tick style={grid=major, major grid style={black!55, solid}},
        ytick={-20,-10,0,10,20,25},
        set layers,
        axis on top=false,
        legend style={
            at={(0.42,-0.11)},
            anchor=north,
            legend columns=5,
            draw=none,
            fill=none,
            font=\tiny,
            on layer=axis background
        },
        nodes near coords,
        every node near coord/.append style={
            font=\tiny,
            rotate=90,
            anchor=west,
            color=black
        },
    ]

    % 14B
    \addplot+[fill=ETHBlue, draw=ETHBlue]
        coordinates {(ID,4.46) (SID,3.48) (OOD,6.73)};
    \addlegendentry{\textsf{14B}}

    % Self-Distill
    \addplot+[fill=ETHPurple, draw=ETHPurple]
        coordinates {(ID,1.72) (SID,-0.91) (OOD,-0.18)};
    \addlegendentry{Self-Distill}

    % 14B -> 8B
    \addplot+[fill=ETHRed, draw=ETHRed]
        coordinates {(ID,6.48) (SID,-3.92) (OOD,-4.96)};
    \addlegendentry{\textsf{14B}$\dto$\textsf{8B}}

    % FFT Distill
    \addplot+[fill=softorange, draw=softorange,
        % y>20时向下平移(负值)，数值大小(-25)可以根据你需要伸进柱子的深度微调
        visualization depends on={y>20 ? -25 : 0 \as \myshift},
        nodes near coords style={xshift=\myshift pt} % 这里改成了 xshift
    ]
        coordinates {(ID,21.21) (SID,-14.05) (OOD,-17.02)};
    \addlegendentry{$\FFT$}

    % LST Distill
    \addplot+[fill=ETHPetrol, draw=ETHPetrol]
        coordinates {(ID,10.20) (SID,0.28) (OOD,-0.11)};
    \addlegendentry{$\LST$ }

    % baseline annotations
    \node[gray, font=\scriptsize] at (axis cs:ID,-18.2) {base: 19.34};
    \node[gray, font=\scriptsize] at (axis cs:SID,-18.2) {base: 60.19};
    \node[gray, font=\scriptsize] at (axis cs:OOD,-18.2) {base: 42.22};

    \end{axis}
    \end{tikzpicture}
    \caption{Delta performance on Chemistry relative to the \textsf{Qwen3-8B} origin model $\Mzero$. The gray labels the absolute baseline score (as in \cref{tab:main}) of $\Mzero$ for each category. \textsf{14B}$\dto$\textsf{8B} denotes the \textit{cross-scale distillation} in contrast to \emph{self-distill} baseline.}
    \label{fig:cross_scale}
    
\end{figure}
We test whether a stronger teacher alone is sufficient for effective distillation. For \chem, we use \textsf{Qwen3-14B} to generate $(q,\hat{\tau},\hat{a})$ data, apply the same filtering procedure, and subsample to the same training size as \textit{Self-Distill}, before continuing $\FFT$ on \textsf{Qwen3-8B}. This cross-scale baseline achieves $25.82/\allowbreak56.27/\allowbreak37.26$ on \chem \ID/\SID/\OOD. \Cref{fig:cross_scale} shows that cross-scale distillation (\textsf{14B}$\dto$\textsf{8B}) improves \ID by $+6.48$, but reduces \SID and \OOD by $-3.92$ and $-4.96$. In contrast, specialist distillation from the same \textsf{8B} base yields better overall trade-offs, suggesting that teacher strength alone is insufficient and echoing that effective distillation also depends on the compatibility between generated supervision and the student learner \citep{speculative_knowledge_distillation}.

\subsection{Additional Teacher-forcing Next-token Prediction Metrics: Probability and Rank}
\label{app:prob_rank}

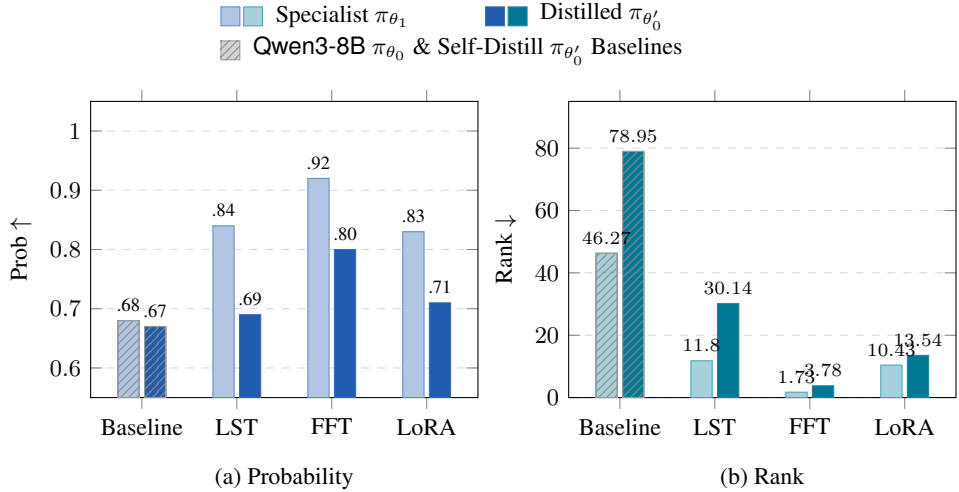
\begin{figure}[ht]
\centering
\usetikzlibrary{patterns, calc}
\begin{tikzpicture}
\pgfplotsset{
    every axis/.style={
        width=0.48\linewidth,
        height=5.5cm,
        enlarge x limits=0.18,
        symbolic x coords={Baseline, LST, FFT, LoRA},
        xtick=data,
        ybar=2pt,
        bar width=8pt,
        ymajorgrids=true,
        grid style={dashed, gray!30},
        tick label style={font=\small},
        label style={font=\small},
        x tick label style={font=\small, align=center, rotate=0},
        nodes near coords,
        nodes near coords style={font=\scriptsize, anchor=south},
    },
}

% ---- Prob subplot ----
\begin{axis}[
    name=prob,
    ylabel={Prob $\uparrow$},
    ymin=0.55, ymax=1.05,
    ytick={0.6,0.7,0.8,0.9,1.0},
    nodes near coords={\pgfplotspointmeta},
    point meta=explicit symbolic,
]
% Stage 1
\addplot[fill=ETHBlue!35, draw=ETHBlue!70, bar shift=-5pt]
    coordinates {(Baseline, 0.68) [.68] (LST, 0.84) [.84] (FFT, 0.92) [.92] (LoRA, 0.83) [.83]};
% Stage 2
\addplot[fill=ETHBlue, draw=ETHBlue!90, bar shift=5pt]
    coordinates {(Baseline, 0.67) [.67] (LST, 0.69) [.69] (FFT, 0.80) [.80] (LoRA, 0.71) [.71]};
% Hatch overlay
\addplot[fill=none, draw=ETHGray!70, bar shift=-5pt, postaction={pattern=north east lines, pattern color=ETHGray!80}, nodes near coords=\empty, forget plot] coordinates {(Baseline, 0.68)};
\addplot[fill=none, draw=ETHGray!70, bar shift=5pt, postaction={pattern=north east lines, pattern color=ETHGray!80}, nodes near coords=\empty, forget plot] coordinates {(Baseline, 0.67)};
\end{axis}
\node[anchor=north, font=\small] at ($(prob.south) + (0, -0.8cm)$) {(a) Probability};

% ---- Rank subplot ----
\begin{axis}[
    name=rank,
    at={(prob.east)},
    anchor=west,
    xshift=1.2cm,
    ylabel={Rank $\downarrow$},
    ymin=0, ymax=95,
    ytick={0,20,40,60,80},
]
% Stage 1
\addplot[fill=ETHPetrol!35, draw=ETHPetrol!70, bar shift=-5pt]
    coordinates {(Baseline, 46.27) (LST, 11.80) (FFT, 1.73) (LoRA, 10.43)};
% Stage 2
\addplot[fill=ETHPetrol, draw=ETHPetrol!90, bar shift=5pt]
    coordinates {(Baseline, 78.95) (LST, 30.14) (FFT, 3.78) (LoRA, 13.54)};
% Hatch overlay
\addplot[fill=none, draw=ETHGray!70, bar shift=-5pt, postaction={pattern=north east lines, pattern color=ETHGray!80}, nodes near coords=\empty, forget plot] coordinates {(Baseline, 46.27)};
\addplot[fill=none, draw=ETHGray!70, bar shift=5pt, postaction={pattern=north east lines, pattern color=ETHGray!80}, nodes near coords=\empty, forget plot] coordinates {(Baseline, 78.95)};
\end{axis}
\node[anchor=north, font=\small] at ($(rank.south) + (0, -0.8cm)$) {(b) Rank};

% ---- Shared legend (manually drawn) ----
\coordinate (legendTop) at ($(prob.north east) + (0.6cm, 1.0cm)$);

\fill[ETHBlue!35, draw=ETHBlue!70] ($(legendTop) + (-4.0, 0)$) rectangle ++(0.25, 0.25);
\fill[ETHPetrol!35, draw=ETHPetrol!70] ($(legendTop) + (-3.7, 0)$) rectangle ++(0.25, 0.25);
\node[anchor=west, font=\small] at ($(legendTop) + (-3.4, 0.125)$) {Specialist $\Mspec$};

\fill[ETHBlue, draw=ETHBlue!90] ($(legendTop) + (-0.5, 0)$) rectangle ++(0.25, 0.25);
\fill[ETHPetrol, draw=ETHPetrol!90] ($(legendTop) + (-0.2, 0)$) rectangle ++(0.25, 0.25);
\node[anchor=west, font=\small] at ($(legendTop) + (0.1, 0.125)$) {Distilled $\Mdist$};

\coordinate (legendBot) at ($(legendTop) + (0, -0.45cm)$);
\fill[ETHGray!30, draw=ETHGray!70, postaction={pattern=north east lines, pattern color=ETHGray!80}] ($(legendBot) + (-4.0, 0)$) rectangle ++(0.25, 0.25);
\node[anchor=west, font=\small] at ($(legendBot) + (-3.7, 0.125)$) {\textsf{Qwen3-8B} $\Mzero$ \& Self-Distill $\Mdist$ Baselines};

\end{tikzpicture}
\caption{Comparison of Chemistry specialist $\Mspec$ and distilled $\Mdist$ models across \textit{teacher-forcing} next-token Probability (Higher $\uparrow$ is better) and Rank (Lower $\downarrow$ is better) metrics. Baseline groups ($\Mzero$\,/\,$\Mdist$) are shown with hatched overlay.}
\label{fig:prob_rank_appx}
\end{figure}

In addition to the Top-1 Hit Rate discussed in \cref{sec:analysis:hr}, we also evaluate the next-token probability and rank using \textit{teacher-forcing} on the same $2,000$ unseen samples.

As shown in \cref{fig:prob_rank_appx}, both Probability and Rank exhibit behaviors highly consistent with the Top-1 Hit Rate. Specifically, the unconstrained $\FFT$ specialist achieves near-perfect next-token probability ($0.92$) and an extremely low rank ($1.73$). This extreme over-confidence highlights a severe collapse of its trajectory distribution, indicating that the model arbitrarily distorts its generation manifold to overfit the answer marginal likelihood rather than learning a generalizable reasoning process. 

Conversely, parameter-efficient architectures, namely $\LST$ and $\LORA$, demonstrate a much more controlled and calibrated improvement in both metrics. For instance, their specialist probabilities are anchored around $0.83\text{--}0.84$, and their ranks are maintained at approximately $10.4\text{--}11.8$. This empirical observation aligns well with the implicit structural regularization provided by parameter-efficient tuning. By restricting the optimization space, $\LST$ and $\LORA$ prevent the trajectory distribution from degenerating into rote memorization. Consequently, these structural constraints enable a smoother, more robust transfer of rationale quality to the distilled student models ($\Mdist$), effectively avoiding the downstream generalizability degradation typically caused by extreme over-specialization.

\section{Experimental Setup Details}\label{app:exp_details}

\subsection{Training}\label{app:exp_details:training}
All fine-tuning experiments are implemented using \texttt{LlamaFactory} \citep{llamafactory} on clusters of NVIDIA \& AMD high-end GPUs. We uniformly employ \texttt{bfloat16} precision to accelerate training. Unless otherwise specified or when multiple runs with different seeds are required for averaging, we fix the random seed to $42$ across all experiments to ensure reproducibility. We maintain a consistent effective batch size of $128$ across all runs. Implementation-wise, we constrain 
\begin{equation}
    \texttt{per\_device\_train\_batch\_size} \times \texttt{gradient\_accumulation\_steps} \times \texttt{world\_size} \equiv 128.
\end{equation}

The exact combination varies dynamically to accommodate different computational requirements of the training data. The cutoff lengths are $16,384$ for \chem, $15,384$ for \phys, and $8,192$ for \MT. The specific configurations for each FT strategy are detailed below:
\begin{itemize}[leftmargin=10pt]
    \item $\FFT$. We train models for $1$ epoch by \texttt{AdamW} \citep{adamw} with a $1\%$ weight decay. The learning rate ($\mathrm{lr}$) follows a cosine decay schedule, ascending to a peak of $\mathrm{lr}_{\max} = 1 \times 10^{-5}$ with a warmup ratio of $3\%$ and subsequently decaying to a minimum of $\mathrm{lr}_{\min}=2 \times 10^{-6}$.
    \item $\LST$. Following \citet{llamax2}, we adopt the configuration reported to achieve the best performance on \textsf{Qwen3-8B}. We restrict trainable parameters to the bottom $4$ and top $16$ transformer layers with all other layers frozen. All other hyperparameters remain identical to the $\FFT$ setting.
    \item $\LORA$. We set the intrinsic rank $r = 64$, the scaling parameter $\alpha = 2r$, and a dropout rate of $5\%$. The peak $\mathrm{lr}_{\max}$ increases to $2 \times 10^{-4}$ without a $\mathrm{lr}_{\min}$ constraint. The remaining align with the $\FFT$ setup.
\end{itemize}

\subsection{Test-time Decoding}\label{app:exp_details:decoding}
We use \texttt{vLLM} \citep{vllm} and \texttt{HuggingFace Transformers} \citep{hf_transformers} for CoT generation and \ID evaluation, enabling thinking mode with \texttt{max\_seq\_len} matching the training cutoff lengths (see \appref{app:exp_details:training}) and sampling parameters $T=0.6, \texttt{TopP}=0.95, \texttt{TopK}=20, \texttt{MinP}=0$ in accordance with the \emph{Best Practices}\footnote{\url{https://huggingface.co/Qwen/Qwen3-8B\#best-practices}} of \textsf{Qwen3} series. For a sufficient pool of high-quality CoT rationales for distilled models, $\texttt{n}=3$ candidate responses are generated for each query in \chem and \phys domains, and $\texttt{n}=1$ for \MT.

\section{Details of Training Data and Evaluation Suites}\label{app:bench_details}
\begin{table*}[!t]
    \setlength{\tabcolsep}{1.2pt}
    
    \renewcommand{\arraystretch}{1.5}
    \setcellgapes{3pt}
    \makegapedcells
    % \scriptsize
    \footnotesize
    \centering
    \caption{\textbf{Overview of evaluation suites} across the three studied \emph{target} domains (\chem, \phys and \MT) under \textit{Training}, \ID and \SID settings, and three \OOD benchmarks (complex reasoning for \textsf{BBEH}, math for \textsf{AIME} and coding for \textsf{LCB}). Bracketed terms (e.g., \textsf{SuperGPQA[Chemistry, $\cdots$]}) indicate that only these domain-relevant subset(s) are evaluated, rather than the entire benchmark. The numbers in parentheses in the table headers denote the number of evaluated subsets (or variants) for each \emph{target} domain under the respective benchmark category. Note that the \OOD benchmarks are shared across all three \emph{target} domains. Notably, \textsf{SMol} officially consists of 14 subtasks (see \appref{app:bench_details:chem:smol}). For the \MT domain, we select 8 low-resource languages (see \appref{app:bench_details:mt}) and evaluate bidirectional translation with English (\texttt{en}$\rightleftarrows$\texttt{xx}), yielding 16 subsets in total.}  
    
\begin{threeparttable}
\begin{tabular}{l|l|l|l|l}

\toprule\hline

\multicolumn{1}{l|}{\bfseries Domain} &
  \multicolumn{1}{l|}{\bfseries Training ($14/1/16$)} &
  \multicolumn{1}{l|}{\bfseries \ID ($14/1/16$)} &
  \bfseries \SID $(4/4/(2\times8))$ &
  \multicolumn{1}{l}{\bfseries \OOD ($5^{=}$)} \\
\hline \hline
\multirow{6}{*}{\makecell[l]{Chemistry\\(\chem)}} &
  \multirow{6}{*}{\makecell[l]{\textsf{SMolInstruct} \\ \textsf{[Train] (SMol)} \\ \citep{smol}}} &
  \multirow{6}{*}{\textsf{SMol[Test]}} &
  \makecell[l]{\textsf{ChemBench} \\ \citep{chembench} } &
  \multirow{14}{*}{\makecell[l]{\textsf{BIG-Bench} \\ \textsf{Extra Hard} \\ (\textsf{BBEH}) \\ \citep{bbeh}\\ \\ \textsf{AIME} \\ \textsf{2025\tnote{a} \ \& 2026\tnote{b}} \\ \\ \textsf{LCB} \textsf{V5}\tnote{d} \ \& \textsf{V6}\tnote{e} }} \\
\cline{4-4}
&
   &
   &
  \makecell[l]{\textsf{MMLU-Pro[Chemistry]} \\  \citep{mmlupro}} &
   \\
\cline{4-4}
&
   &
   &
  \makecell[l]{ \textsf{SuperGPQA[Chemistry,} \\ \textsf{Chemical Engineering}  \\  \textsf{and Technology]} \\ \citep{supergpqa} } &
   \\
\cline{1-4}
\multirow{6}{*}{\makecell[l]{Physics\\(\phys)}} &
  \multirow{6}{*}{\makecell[l]{\textsf{MegaScience} \\ \textsf{[Physics]} \\ \citep{megascience}} } &
  \multirow{6}{*}{\makecell[l]{\textsf{PHYSICS[} \\ \textsf{Undergraduate} \\ \textsf{/Postgraduate} \\ \textsf{(Physics Major)]} \\ \citep{physics}}} &
  \textsf{PIQA} \citep{piqa} &
   \\
\cline{4-4}
&
   &
   &
  \makecell[l]{\textsf{AGIEval[Gaokao Physics]} \\   \citep{agieval}} &
   \\
\cline{4-4}
&
   &
   &
  \makecell[l]{\textsf{MMLU[High School Physics]} \\ \citep{mmlu} } &
   \\
\cline{4-4}
&
   &
   &
  \textsf{MMLU-Pro[Physics]} &
   \\
\cline{1-4}
\multirow{2}{*}{\makecell[l]{Low-Resource \\ Multilingualism \\ (\MT)}} &
  \multirow{2}{*}{\makecell[l]{\textsf{OPUS[en$\rightleftarrows$} \\ \textsf{\{bn,cs,hu,sr,sw,te,th,vi\}]} \\ \citep{opus} }} &
  \multirow{2}{*}{\makecell[l]{\textsf{Flores-101[en$\rightleftarrows$} \\ \textsf{\{bn,cs,hu,sr,sw,te,th,vi\}]}  \\ \citep{flores101} }} &
  \textsf{IFEval}\tnote{c} \ \citep{ifeval} &
   \\
\cline{4-4}
&
   &
   &
  \textsf{GPQA}\tnote{c} \ \citep{gpqa} &
   \\
% \cline{4-4}
% &
%    &
%    &
%   \makecell[l]{\textsf{LiveCodeBench}\tnote{c} \ \\ (\textsf{LCB})  \textsf{V4}\tnote{f} \citep{lcb}} & \\

\hline
\bottomrule
\end{tabular}

    \begin{tablenotes}
        \footnotesize
        \item[a] The American Invitational Mathematics Examination (AIME). \url{https://huggingface.co/datasets/math-ai/aime25}
        \item[b] \url{https://huggingface.co/datasets/math-ai/aime26}
        \item[c] These two \SID benchmarks in the \MT domain are originally in English. We use their multilingual translations provided by \texttt{BenchMAX} \citep{benchmax}, and evaluate on $8$ non-English low-resources languages: \{bn, cs, hu, sr, sw, te, th, vi\} (see \cref{tab:app:low_langs} in \appref{app:bench_details:mt} for their names and properties), rather than the original English versions.
        \item[d] Problems released between May 2023 and January 2025, containing $880$ problems. See \url{https://github.com/livecodebench/livecodebench\#dataset-versions} for details.
        \item[e] Problems released between May 2023 and April 2025, containing $1,055$ problems.
        % \item[f] Problems released between May 2023 and Sep 2024.
    \end{tablenotes}
    \end{threeparttable}

    \vspace{-5pt}
    \label{tab:app:eval_suites}
\end{table*}

\cref{tab:app:eval_suites} provides a comprehensive overview of all evaluation suites used across the three target domains (\chem, \phys, and \MT) under the Training, \ID, and \SID\ settings, as well as the shared \OOD\ benchmarks spanning complex reasoning, mathematics, and coding. 

We employ the widely adopted frameworks \texttt{OpenCompass} \citep{opencompass}, \texttt{LM-Eval-Harness} \citep{lm_eval_harness}, and \texttt{BenchMAX} \citep{benchmax} to ensure a standardized, reproducible, and fair evaluation for \SID and \OOD, with the maximum sequence lengths $\texttt{max\_seq\_len} = 40,960$, i.e., the \texttt{max\_position\_embeddings} of \textsf{Qwen3} series, and temperature $T=0.6$. 
\subsection{Chemistry}\label{app:bench_details:chem}
\subsubsection{Chemistry Training and \ID\ Data --- \textsf{SMolInstruct}}\label{app:bench_details:chem:smol}
\textsf{SMolInstruct} is a large-scale instruction tuning dataset crafted for \chem domain. 
\paragraph{Subtasks Composition.}
\textsf{SMolInstruct} comprises $14$ instruction-following molecular (sub-)tasks: \textsf{forward synthesis (FS), retrosynthesis (RS), molecule captioning (MC), molecule generation (MG), name conversion-i2f (I2F), name conversion-i2s (I2S), name conversion-s2f (S2F), name conversion-s2i (S2I), property prediction-esol (ESOL), property prediction-lipo (Lipo), property prediction-bbbp (BBBP), property prediction-clintox (ClinTox), property prediction-hiv (HIV)}, and \textsf{property prediction-sider (SIDER)}. These tasks cover reaction prediction, molecular generation and understanding, conversion of molecular representations, and property prediction, providing a broad testbed in \chem.

\paragraph{Subtask-Specific Evaluation Metrics.}
We adopt subtask-specific evaluation metrics according to \citet{smol}:
\begin{enumerate}[leftmargin=16pt,label=(\arabic*)]
    \item \textbf{\textsf{FS, RS} and \textsf{MG}}
    These are molecular generation tasks. We evaluate them using \emph{Morgan Fingerprint Tanimoto Similarity} \citep[$\mathrm{Morgan}$ FTS,][]{morgan}, which measures the similarity between the ground-truth molecule and the generated molecule. The score ranges from $0$ to $1$, where a larger value indicates higher structural similarity.
    
    \item \textbf{\textsf{MC}.}
    This task requires generating a textual description for a molecule. We evaluate the semantic similarity between the generated caption and the reference caption using $\mathrm{METEOR}$ score \citep{meteor}, whose value lies in $[0.0,1.0]$, the higher the more similar.

    \item \textbf{\textsf{BBBP, ClinTox, HIV} and \textsf{SIDER}.}
    These are binary classification tasks. We report na\"ive \emph{accuracy} ($\mathrm{Acc}$) as the evaluation metric.

    \item \textbf{\textsf{I2F, I2S, S2F} and \textsf{S2I}.}
    These four tasks require converting one molecular representation from another. We adopt variants of \emph{exact match} ($\mathrm{EM}$) accuracy, including \emph{element match} and \emph{split match}, to evaluate both element-order-independent exactness and partial structural consistency.

    \item \textbf{\textsf{ESOL} and \textsf{Lipo}.}
    These are numerical regression tasks. We use \emph{Root Mean Square Error} ($\mathrm{RSME}$) to measure the deviation between the predicted values and the ground-truth values. $\mathrm{RSME}$ ranges from $0$ to $+\infty$, where a smaller value indicates better performance.
\end{enumerate}

\paragraph{Overall Score.}
In \cref{tab:main}, we report a unified overall score in the range $[0.00\%,100.00\%]$ to summarize model performance across all $14$ tasks. Since the raw metrics are heterogeneous and have different scales and optimization directions, we compute the \textsf{SMol} overall score as follows:

\begin{equation}\label{eq:smol_score}
\begin{aligned}
\mathrm{Score}_{\textsf{SMol}} = 100\% \times \frac{1}{14} \Biggl\{&
\mathrm{Morgan}_{\textsf{FS}} + \mathrm{Morgan}_{\textsf{RS}} + \mathrm{Morgan}_{\textsf{MG}} \\
&+ \mathrm{METEOR}_{\textsf{MC}} \\
&+ \mathrm{Acc}_{\textsf{BBBP}} + \mathrm{Acc}_{\textsf{ClinTox}} + \mathrm{Acc}_{\textsf{HIV}} + \mathrm{Acc}_{\textsf{SIDER}} \\
&+ \mathrm{EM}_{\textsf{I2F}} + \mathrm{EM}_{\textsf{I2S}} + \mathrm{EM}_{\textsf{S2F}} + \mathrm{EM}_{\textsf{S2I}} \\
&+ \max\left(\frac{2 - \mathrm{RMSE}_{\textsf{ESOL}}}{2},\, 0\right)
+ \max\left(\frac{1.2 - \mathrm{RMSE}_{\textsf{Lipo}}}{1.2},\, 0\right)
\Biggr\}.
\end{aligned}
\end{equation}

\paragraph{Stage-2 CoT Rationale Filtration Strategies.}\label{app:bench_details:chem:smol:stage2}
Unlike the testing evaluation protocols described above, which often allow for partial credit (e.g., structural similarity via $\mathrm{Morgan}$ FTS even if the generated molecule is not exactly identical to the ground-truth), our rationale filtration process \emph{strictly} enforces equivalence to the ground-truth (GT) answers, if feasible. The guiding principle is to ensure the absolute high quality and correctness of the generated Chain-of-Thought (CoT) rationales. Based on this stringent equivalence principle and the distinct output formats, we re-categorize the $14$ subtasks into $6$ groups and apply tailored filtration rules:

\begin{enumerate}[leftmargin=20pt,label=\Roman*.]
    \item \textbf{SMILES-based Tasks (\textsf{FS, RS, MG, I2S}).} 
    The output format is \textit{Simplified Molecular-Input Line-Entry System} \citep[SMILES,][]{smiles}, which represents molecular graphs as ASCII strings (multiple molecules are delimited by ``\texttt{.}''). Since a single molecule can be legally represented by multiple valid SMILES strings, we utilize the standard \texttt{RDKit} \citep{rdkit} Python library to convert both the generated and GT SMILES into their \emph{canonical} forms. A rationale is retained only if its canonicalized SMILES perfectly matches the GT, ensuring absolute structural equivalence.
    
    \item \textbf{Molecular Formula Tasks (\textsf{S2F, I2F}).} 
    The output is a molecular formula (e.g., \texttt{C6H12O6}). We apply an \emph{Element Match} rule: we extract the constituent elemental symbols and their corresponding counts from the response and compare them to the GT. The rationale is kept if the elements and their quantities are identical, completely disregarding the order in which the elements appear.
    
    \item \textbf{IUPAC Naming Task (\textsf{S2I}).} 
    The output is an \textit{International Union of Pure and Applied Chemistry} \citep[IUPAC,][]{iupac_blue_book} name (e.g., \texttt{2-methylpropane}). We employ a \emph{Split Match} criterion. Both the generated name and the GT are tokenized by splitting at the hyphen (\texttt{-}). The rationale is preserved if the set of generated splits perfectly matches the set of GT splits, disregarding their relative order.
    
    \item \textbf{Binary Property Prediction (\textsf{BBBP, ClinTox, HIV, SIDER}).} 
    The outputs are binary choices (e.g., \texttt{yes}/\texttt{no} or \texttt{True}/\texttt{False}). We parse the final predicted label via heuristic keyword matching and retain the rationale only if the parsed boolean value strictly aligns with the GT annotation.
    
    \item \textbf{Numerical Property Prediction (\textsf{ESOL, Lipo}).} 
    The outputs are continuous numerical values. We calculate the absolute difference between the parsed predicted value and the GT. Rationales are retained if and only if its absolute error is $\le 1.0$.
    
    \item \textbf{Text Generation (\textsf{MC}).} 
    The output is a natural language description of a molecule. We calculate the $\mathrm{METEOR}$ score between the generated caption and the GT reference. A rationale is considered valid and thus retained if its $\mathrm{METEOR}$ score is $\ge 0.25$.
\end{enumerate}

\paragraph{Data Scale and Preparation.} \label{app:bench_details:chem:smol:data_scale}
\begin{table*}[!htbp]
    \small
    \centering
    \alternaterowcolors
    \renewcommand{\arraystretch}{1.5}
    \setlength{\tabcolsep}{7pt}

    \caption{\textbf{Statistics of \textsf{SMolInstruct} across subtasks}, including the \textit{downsampled} training split used for rapid experimentation, the \textit{full} training split, and the official in-task (\ID) test split. For the downsampled setting, each subtask is capped at $50$K training instances; subtasks with fewer than $50$K original examples are upsampled accordingly. For the full training setting, additional upsampling is applied to selected low-resource subtasks to mitigate data imbalance, resulting in a final rebalanced training set of \textasciitilde$3.6$M instances.}

    \begin{tabular}{l l l l}
        \toprule
        \bfseries \SMol Subtasks & \bfseries \textit{Downsampled} Training Split & \bfseries \textit{Full} Training Split & \bfseries Test (\ID) Split \\
        \midrule
        \textit{Total} & $534,805$ & $3,288,855 \xrightarrow{\text{upsampled}} 3,675,404$ & $33,061$ \\
        \midrule
        \textsf{FS} & $50,000$ & $971,809$ & $4,062$ \\
        \textsf{RS} & $50,000$ & $941,735$ & $4,156$ \\
        \textsf{MG} & $50,000$ & $56,498$ & $2,493$ \\
        \textsf{MC} & $50,000$ & $56,498$ & $2,538$ \\
        \textsf{BBBP} & $1,569 \xrightarrow{5\times} 7,845$ & $1,569 \xrightarrow{50\times} 78,450$ & $197$ \\
        \textsf{ClinTox} & $1,144 \xrightarrow{5\times} 5,720$ & $1,144 \xrightarrow{50\times} 57,200$ & $144$ \\
        \textsf{HIV} & $32,864 \xrightarrow{\text{up to } 50\text{K}} 50,000$ & $32,864$ & $4,107$ \\
        \textsf{SIDER} & $22,820 \xrightarrow{\text{up to } 50\text{K}} 50,000$ & $22,820 \xrightarrow{3\times} 68,460$ & $2,860$ \\
        \textsf{I2F} & $50,000$ & $300,000$ & $2,993$ \\
        \textsf{I2S} & $50,000$ & $299,890$ & $2,993$ \\
        \textsf{S2F} & $50,000$ & $299,890$ & $2,993$ \\
        \textsf{S2I} & $50,000$ & $299,890$ & $2,993$ \\
        \textsf{ESOL} & $888 \xrightarrow{5\times} 4,440$ & $888 \xrightarrow{50\times} 44,400$ & $112$ \\
        \textsf{Lipo} & $3,360 \xrightarrow{5\times} 16,800$ & $3,360 \xrightarrow{50\times} 168,000$ & $420$ \\
        \bottomrule
    \end{tabular}

    \vspace{-.2cm}
    \label{tab:app:smol_data_amount}
\end{table*}
\textsf{SMolInstruct} is constructed from a large-scale training corpus with substantial variation in data volume across subtasks. In its original form, the training split contains \textasciitilde$3.2$M instances, while the test split contains \textasciitilde$33$K instances. The exact number of examples for each subtask is reported in the third and fourth columns of \cref{tab:app:smol_data_amount}.

To accelerate early-stage experimentation, ablation studies, and hyperparameter exploration, we additionally curate a downsampled training set, whose per-subtask sizes are summarized in the second column of \cref{tab:app:smol_data_amount}. Specifically, for each subtask, we randomly sample up to $50$K training instances. For six subtasks whose original training sets contain fewer than $50$K examples, we retain all available instances and expand the corresponding subtask data to $50$K through repeated replication (i.e., copying the full set $5$ times at most) or, more generally, by upsampling to the target size.

We also note that the original training split exhibits data imbalance among subtasks. To alleviate this issue, we perform additional upsampling for five subtasks, as detailed in \cref{tab:app:smol_data_amount}. After this rebalancing procedure, the final full training split used in our main experiments contains \textasciitilde$3.6$M instances.

\begin{figure}[!htbp]
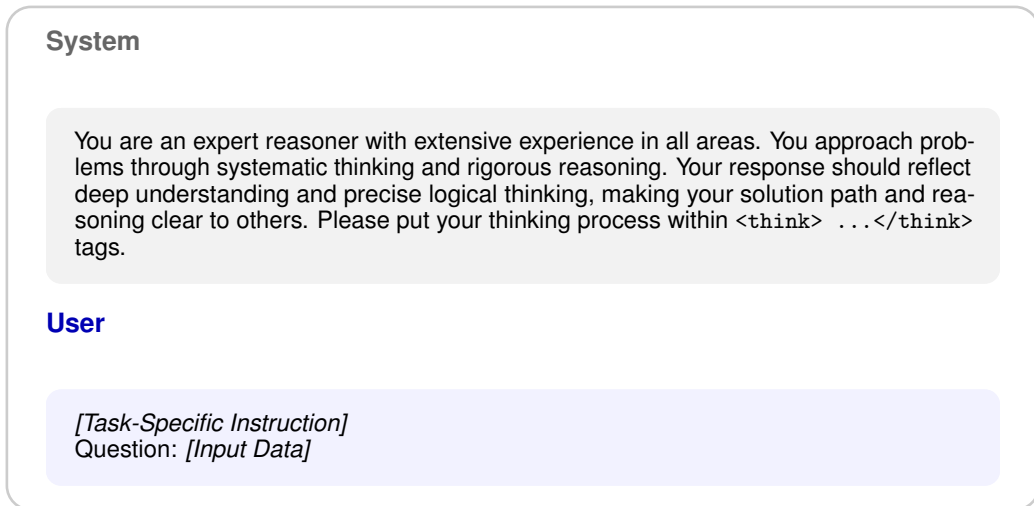

\centering
\begin{tcolorbox}[
  width=0.98\linewidth,
  colback=white,
  colframe=gray!40,
  arc=3mm,
  boxrule=1pt,
]
% System Message
\noindent\textbf{\textsf{\color{gray!80!black} System}}\\[2pt]
\begin{tcolorbox}[
  colback=gray!10,
  colframe=gray!10,
  arc=2mm,
  boxrule=0pt,
  left=8pt, right=8pt, top=6pt, bottom=6pt
]
\small\sffamily
You are an expert reasoner with extensive experience in all areas. You approach problems through systematic thinking and rigorous reasoning. Your response should reflect deep understanding and precise logical thinking, making your solution path and reasoning clear to others. Please put your thinking process within \texttt{<think> \dots </think>} tags.
\end{tcolorbox}

\vspace{0.5em}
% User Message
\noindent\textbf{\textsf{\color{blue!70!black} User}}\\[2pt]
\begin{tcolorbox}[
  colback=blue!5,
  colframe=blue!5,
  arc=2mm,
  boxrule=0pt,
  left=8pt, right=8pt, top=6pt, bottom=6pt
]
\small\sffamily
\textit{[\textsf{Task-Specific Instruction}]} \\
Question: \textit{[\textsf{Input Data}]}
\end{tcolorbox}
\end{tcolorbox}
\caption{\textbf{The unified ShareGPT-style prompt template for \textsf{SMolInstruct}.} The model is prompted with a constant system message to encourage rigorous CoT reasoning. The user message provides the task-specific instruction alongside the input data.}
\label{fig:smol_prompt_template}
\end{figure}

\paragraph{Prompt Template}
We formulate all tasks into a unified conversational format following the \texttt{ShareGPT} scheme. To elicit the rigorous systematic reasoning capabilities of the model, we employ a consistent system prompt across all tasks. This system prompt explicitly instructs the model to act as an expert reasoner and encapsulate its thinking process within \texttt{<think> \dots </think>} tags. The user prompt consists of a task-specific instruction followed by the input question. The visual representation of our \texttt{ShareGPT}-style prompt template is illustrated in \cref{fig:smol_prompt_template}.

% The \textit{[\textsf{Task-Specific Instruction}]} in \cref{fig:smol_prompt_template} is specifically customized for each of the $14$ subtasks to define the chemical context, specify the input format, and strictly constrain the desired output format (e.g., using specific XML tags or \LaTeX\ \texttt{\textbackslash boxed\{\}} commands). The exact text for each task's instruction is listed as follows:

The \textit{[\textsf{Task-Specific Instruction}]} in \cref{fig:smol_prompt_template} is specifically customized for each of the $14$ subtasks to define the chemical context, specify the input format, and strictly constrain the desired output format (e.g., using specific XML tags or \LaTeX\ \texttt{\textbackslash boxed\{\}} commands). The exact text for each task's instruction is provided in \cref{fig:smol_task_instructions}.

\subsubsection{Chemistry \SID\ Benchmarks}\label{app:bench_details:chem:sid}
To comprehensively evaluate the robustness and generalization capabilities of our model under distribution shifts within the chemical domain (i.e., the \SID setting in \cref{tab:app:eval_suites}), we employ three challenging, expert-level benchmarks. All evaluations in this section are conducted in a \textbf{zero-shot} setting using the \texttt{OpenCompass} \citep{opencompass}. We report \textbf{accuracy ($\mathrm{Acc}$)} as the unified evaluation metric across all datasets. The detailed descriptions of these benchmarks are as follows:

\begin{itemize}[leftmargin=12pt, itemsep=4pt]
    \item \textbf{\textsf{MMLU-Pro [Chemistry]}:} 
    This is the chemistry-specific subset of MMLU-Pro \citep{mmlupro}. As an enhanced and more robust version of the original MMLU, MMLU-Pro significantly increases the task difficulty by expanding the number of distractor options and filtering out trivial questions, thereby providing a rigorous testbed for college- and professional-level chemical knowledge.
    
    \item \textbf{\textsf{SuperGPQA [Chemistry, Chemical Engineering and Technology]}:} 
    This comprises two domain-relevant subsets derived from the highly challenging SuperGPQA benchmark suite \citep{supergpqa}. The \textit{Chemistry} subset specifically evaluates PhD-level theoretical understanding, experimental logic, and complex chemical problem-solving abilities that are difficult to bypass via simple search engines. The \textit{Chemical Engineering and Technology} subset shifts the focus to practical engineering, assessing the model's capability to reason about industrial scaling, thermodynamics, applied materials science, and chemical processes.
    
    \item \textbf{\textsf{ChemBench}:} 
    ChemBench \citep{chembench} is a comprehensive, domain-specific evaluation framework tailored for large language models in chemistry. It spans a wide range of sub-disciplines (e.g., physical, organic, inorganic, and analytical chemistry) and evaluates models on their ability to understand chemical properties, follow reaction rules, and perform multi-step scientific reasoning.
\end{itemize}

\subsection{Physics}\label{app:bench_details:phys}
\subsubsection{Physics Training Data --- \textsf{MegaScience}}\label{app:bench_details:phys:megascience}
\textsf{MegaScience} \citep{megascience} is a large-scale, high-quality scientific reasoning dataset designed to enhance the specialized knowledge and logical reasoning capabilities of large language models across multiple disciplines, including physics, chemistry, and biology.
During the construction of our training set, we extracted physics problems from the full corpus.

\paragraph{Data Generation Strategy.}
To enhance the diversity of the model's reasoning paths and improve the robustness of CoT generation, we executed three independent inferences for each physics problem using three different random seeds.
The conversational prompt is illustrated as \cref{fig:megascience_prompt_template}.

\begin{figure}[!htbp]
\centering
\begin{tcolorbox}[
  width=0.98\linewidth,
  colback=white,
  colframe=gray!40,
  arc=3mm,
  boxrule=1pt,
]
% System Message
\noindent\textbf{\textsf{\color{gray!80!black} System}}\\[2pt]
\begin{tcolorbox}[
  colback=gray!10,
  colframe=gray!10,
  arc=2mm,
  boxrule=0pt,
  left=8pt, right=8pt, top=6pt, bottom=6pt
]
\small\sffamily
You are an expert physicist. Solve the following problem step by step.\\
After your reasoning, clearly state your final answer.\\
If the answer is a number, give the numeric value.\\
If the answer is an expression, write it in LaTeX. \\
Wrap your final answer in \textbackslash\textbackslash boxed\{\}. \\
\end{tcolorbox}

\vspace{0.5em}
% User Message
\noindent\textbf{\textsf{\color{blue!70!black} User}}\\[2pt]
\begin{tcolorbox}[
  colback=blue!5,
  colframe=blue!5,
  arc=2mm,
  boxrule=0pt,
  left=8pt, right=8pt, top=6pt, bottom=6pt
]
\small\sffamily
\textit{[\textsf{Question}]}
\end{tcolorbox}
\end{tcolorbox}
\caption{\textbf{Prompt template for \textsf{MegaScience}.}}
\label{fig:megascience_prompt_template}
\end{figure}

\paragraph{Rule-Based Verification.}
To robustly compare the generated answer against the reference ground truth, the verifier sequentially applies five deterministic matching strategies:
\begin{enumerate}[leftmargin=16pt,label=(\arabic*)]
    \item \textbf{Exact Match}: A direct, case-insensitive string comparison after basic \LaTeX\ and whitespace normalization.
    \item \textbf{Numeric Evaluation}: Both answers are parsed as numerical values (seamlessly handling \LaTeX\ scientific notation and fractions). They are deemed equivalent if they match within a $5\%$ relative tolerance or a $10^{-8}$ absolute tolerance.
    \item \textbf{Symbolic Equivalence}: The framework utilizes \texttt{SymPy} to symbolically subtract the parsed \LaTeX\ expressions and check for mathematical equivalence (i.e., simplifying to zero).
    \item \textbf{Normalized Matching}: If symbolic parsing fails, all formatting and non-alphanumeric characters (except basic operators) are stripped for a rigid structural comparison.
    \item \textbf{Substring Inclusion}: For remarkably short reference answers (e.g., under $60$ characters), the prediction is accepted if it fully contains the reference string.
\end{enumerate}

\subsubsection{Physics \ID\ Benchmark --- \textsf{PHYSICS[Undergraduate/Postgraduate(Physics Major)]}}
\label{app:bench_details:phys:physics}

\textsf{PHYSICS} \citep{physics} is a comprehensive, large-scale, and bilingual (English and Chinese) dataset tailored to evaluate and enhance the physical reasoning capabilities of large language models.
For this evaluation, we specifically focus on its most advanced subset to rigorously test expert-level physical reasoning.

\paragraph{Subtasks and Difficulty Composition.}
While the full \textsf{PHYSICS} dataset encompasses various educational stages, in this setting, we strictly isolate the \textsf{Undergraduate/Postgraduate (Physics Major)} difficulty level.
This subset shifts the focus away from foundational concepts to highly complex, expert-level problem-solving.
The evaluated problems span 5 major physics domains: Mechanics, Electromagnetism, Thermodynamics, Optics, and Modern Physics.

\paragraph{Answer-Type-Specific Evaluation Metrics (Rule+Model Framework).}
Given the diversity and complexity of physics answers, standard text matching alone is insufficient.
We therefore adopt a hybrid \textbf{Rule+Model} evaluation framework consisting of a deterministic rule-based verifier followed, when necessary, by an LLM-as-a-judge fallback.
The deterministic verifier applies answer-type-specific rules, including normalized exact matching, numerical comparison, symbolic equivalence checking, MCQ letter-set matching, and bilingual True/False normalization:

\begin{enumerate}[leftmargin=16pt,label=(\arabic*)]
    \item \textbf{\textsf{Numerical}.}
    Numerical answers are extracted while accounting for scientific notation and fractions.
    They are considered equivalent when the prediction matches the reference within a $5\%$ relative tolerance or a $10^{-8}$ absolute tolerance.

    \item \textbf{\textsf{Expression} and \textsf{Equation}.}
    Mathematical expressions and equations are first normalized to remove superficial \LaTeX\ and formatting differences.
    We then attempt symbolic equivalence checking via \texttt{SymPy}, e.g., by simplifying the difference between the predicted and reference expressions.
    Normalized exact matching is also used when the two expressions can be directly matched after canonicalization.

    \item \textbf{\textsf{Multiple Choice (MCQ)} and \textsf{True/False (T/F)}.}
    MCQ answers are evaluated through exact matching of the extracted option-letter set (e.g., sorting letters A--F before comparison).
    For T/F questions, case-insensitive normalization maps bilingual and abbreviated variants (e.g., ``True'', ``\cn{正确}'', ``yes'', and ``T'') to a unified Boolean representation.

    \item \textbf{\textsf{Interval}.}
    Interval-valued answers (e.g., $[-1,1]$) are evaluated using normalized string matching so that both boundary values and inclusion/exclusion symbols must agree with the reference.

    \item \textbf{\textsf{Open-ended and Uncertain (LLM-as-a-Judge)}.}
    For open-ended answers, or whenever deterministic verification cannot confidently establish equivalence, the framework falls back to an LLM judge, specifically \texttt{gemini-3.1-flash-lite-preview}.
    Such cases commonly involve complex multi-variable \LaTeX\ expressions, integrals, vectors, tensors, or other symbolic forms that cannot be reliably converted into standard \texttt{SymPy} representations without task-specific parsing rules.
    The judge determines whether the predicted and reference answers are mathematically or physically equivalent.
\end{enumerate}

\paragraph{Fallback Frequency and Judge Configuration.}
The LLM fallback is frequently invoked because the Undergraduate/Postgraduate subset contains a large proportion of advanced symbolic and open-ended answers that cannot be robustly resolved by generic string or \texttt{SymPy}-based verification.
Across the 2,000 \textsf{PHYSICS} test problems (approximately 3,100--3,300 sub-answers per model), the fallback rates are highly consistent across evaluated model variants: $84.8\%$ for the Qwen3-8B origin model (2,674 sub-answers), $84.9\%$ for the FFT specialist (2,677), $85.9\%$ for the LST specialist (2,707), $85.6\%$ for the LoRA specialist (2,697), and $83.4\%$ for Qwen3-14B (2,628).
The similar fallback frequencies across model variants indicate that use of the LLM judge is primarily determined by the answer structure rather than by a particular model family.

For reproducibility, the LLM judge is queried with \texttt{temperature=0.0} using greedy decoding.
Its system prompt restricts the output to a single binary token-level decision, either \texttt{CORRECT} or \texttt{INCORRECT}, as shown in \cref{fig:llm_judge_prompt_template}.
The same zero-shot judge, prompt, decoding configuration, and verification procedure are applied uniformly to all origin, baseline, specialist, and distilled models to ensure a consistent comparison.

\paragraph{Overall Score and Multi-Part Logic.}
Unlike standard QA benchmarks, many questions in \textsf{PHYSICS} are multi-part, requiring the model to generate a sequence of answers.
We prompt the model to encapsulate each sub-answer in its own separate \texttt{\textbackslash boxed\{\}}.
We apply a strict \textbf{AND logic}: an item is considered fully correct (yielding an Accuracy of $1$) if and only if \emph{all} of its sub-answers are judged correct against their corresponding ground-truth references.
Partial credits are tracked during evaluation, but the primary reported metric remains the strict overall Accuracy ($\mathrm{Acc}$).

\paragraph{Prompt Template.}
We unify the evaluation under a consistent conversational format.
As illustrated in \cref{fig:physics_prompt_template}, the system prompt explicitly establishes the identity of an expert physicist, guiding the model to reason step-by-step before finalizing answers in the required \LaTeX\ \texttt{\textbackslash boxed\{\}} format.
For open-ended or uncertain cases requiring model-based verification, we use the dedicated LLM-as-a-judge template in \cref{fig:llm_judge_prompt_template}, which instructs the judge to assess mathematical and physical equivalence and return only a binary verdict (\texttt{CORRECT} or \texttt{INCORRECT}).

\begin{figure}[!htbp]
\centering
\begin{tcolorbox}[
  width=0.98\linewidth,
  colback=white,
  colframe=gray!40,
  arc=3mm,
  boxrule=1pt,
]
% System Message
\noindent\textbf{\textsf{\color{gray!80!black} System}}\\[2pt]
\begin{tcolorbox}[
  colback=gray!10,
  colframe=gray!10,
  arc=2mm,
  boxrule=0pt,
  left=8pt, right=8pt, top=6pt, bottom=6pt
]
\small\sffamily
You are an expert physicist. Solve the following problem step by step.\\
After your reasoning, clearly state your final answer(s).\\
- If the answer is a number, give the numeric value.\\
- If the answer is a mathematical expression, write it in LaTeX.\\
- If the question is multiple-choice (MCQ), state the correct option letter(s).\\
- If the question is True/False, state True or False.\\
Wrap EACH final answer in \textbackslash\textbackslash boxed\{\}. \\
If there are multiple sub-answers, put each one in its own separate \textbackslash\textbackslash boxed\{\}.
\end{tcolorbox}

\vspace{0.5em}
% User Message
\noindent\textbf{\textsf{\color{blue!70!black} User}}\\[2pt]
\begin{tcolorbox}[
  colback=blue!5,
  colframe=blue!5,
  arc=2mm,
  boxrule=0pt,
  left=8pt, right=8pt, top=6pt, bottom=6pt
]
\small\sffamily
\textit{[\textsf{Question}]}
\end{tcolorbox}
\end{tcolorbox}
\caption{\textbf{The unified ShareGPT-style prompt template for \textsf{PHYSICS}.}}
\label{fig:physics_prompt_template}
\end{figure}

\begin{figure}[!htbp]
\centering
\begin{tcolorbox}[
  width=0.98\linewidth,
  colback=white,
  colframe=gray!40,
  arc=3mm,
  boxrule=1pt,
]
% System Message
\noindent\textbf{\textsf{\color{gray!80!black} System}}\\[2pt]
\begin{tcolorbox}[
  colback=gray!10,
  colframe=gray!10,
  arc=2mm,
  boxrule=0pt,
  left=8pt, right=8pt, top=6pt, bottom=6pt
]
\small\sffamily
You are a strict answer-verification assistant for physics problems.\\
Determine whether the student's answer is mathematically/physically equivalent to the reference answer. Consider different notations, units, rounding, and equivalent expressions. 
Respond with ONLY the single word CORRECT or INCORRECT.
\end{tcolorbox}

\vspace{0.5em}
% User Message
\noindent\textbf{\textsf{\color{blue!70!black} User}}\\[2pt]
\begin{tcolorbox}[
  colback=blue!5,
  colframe=blue!5,
  arc=2mm,
  boxrule=0pt,
  left=8pt, right=8pt, top=6pt, bottom=6pt
]
\small\sffamily
Reference Answer:\textit{[\textsf{ref}]}\\
Student Answer: \textit{[\textsf{model\_ans}]}\\
Verdict:
\end{tcolorbox}
\end{tcolorbox}
\caption{\textbf{The LLM-as-a-judge prompt template for \textsf{PHYSICS}.}}
\label{fig:llm_judge_prompt_template}
\end{figure}

\subsubsection{Physics \SID\ Benchmarks}\label{app:bench_details:phy:sid}
In order to thoroughly assess the model's proficiency in the physical sciences and its resilience to domain-specific distribution shifts, we select five rigorous benchmarks that span various dimensions of physical reasoning. 
All evaluations are implemented utilizing the \texttt{LM-Evaluation-Harness} framework \citep{lm_eval_harness}. 
We adopt specific prompting strategies (i.e., zero-shot or 5-shot) and metrics tailored to the standard evaluation protocol of each respective dataset. 
The details of the selected benchmarks and their specific configurations are outlined below:
\begin{itemize}[leftmargin=12pt, itemsep=4pt]
    \item \textbf{\textsf{PIQA}:} 
    Physical Interaction: Question Answering (PIQA) \citep{piqa} evaluates a model's physical commonsense reasoning. 
    Instead of focusing on abstract academic formulas, it requires an understanding of the affordances of everyday objects and basic physical mechanics to deduce logical outcomes in real-world scenarios. 
    We evaluate this dataset in a \textbf{zero-shot} setting and report the length-normalized accuracy (\textbf{$\mathrm{Acc\_norm}$}).
    
    \item \textbf{\textsf{AGIEval[Gaokao Physics]}:} 
    This subset from the human-centric AGIEval benchmark \citep{agieval} consists of highly challenging, standardized physics questions derived from the Chinese National College Entrance Examination (Gaokao). 
    It assesses the model's ability to tackle complex, multi-step physical problem-solving that requires rigorous mathematical reasoning. 
    This is evaluated in a \textbf{zero-shot} setting, with performance measured by length-normalized accuracy (\textbf{$\mathrm{Acc\_norm}$}).
    
    \item \textbf{\textsf{MMLU[High School Physics]}:} 
    Derived from the Massive Multitask Language Understanding (MMLU) benchmark \citep{mmlu}, the task covers standard curriculum topics ranging from basic kinematics to advanced theoretical concepts. 
    We evaluate in a \textbf{zero-shot} setting, using standard accuracy (\textbf{$\mathrm{Acc}$}) as the metric.

    \item \textbf{\textsf{MMLU-Pro[Physics]:}}
    As the physics-specific subset of MMLU-Pro \citep{mmlupro}, this benchmark significantly increases the evaluation difficulty by expanding the number of distractor options to ten and systematically filtering out trivial questions. 
    We evaluate this subset in a \textbf{5-shot} setting and report the Exact Match (\textbf{$\mathrm{EM}$}) metric.

    % \item \textbf{\textsf{MMLU-Pro+[Physics]:}} 
    % This subset belongs to MMLU-Pro+ \citep{mmluproplus}, an enhanced benchmark designed to assess shortcut learning and higher-order reasoning. 
    % By introducing complex physics questions with multiple correct answers, it rigorously tests the model's ability to resist simplistic problem-solving strategies. 
    % We conduct tests in a \textbf{5-shot} setting and measure performance using the Exact Match (\textbf{$\mathrm{EM}$}) metric.

\end{itemize}

\subsection{Low-Resource Multilingualism}\label{app:bench_details:mt}
\subsubsection{Language Selection}
\begin{table*}[!htbp]
    \small
    \centering
    \alternaterowcolors
    \renewcommand{\arraystretch}{1.3}
    \setlength{\tabcolsep}{9pt} 

    \caption{\textbf{Properties of $8$ low-resource languages} we experiment for \MT domain and English language, in ascending order of ISO 639-1 codes \citep{iso639}.}

    \begin{tabular}{lllll}
        \toprule
        \bfseries ISO Code & \bfseries ISO Language Name & \bfseries Endonym & \bfseries Writing System & \bfseries Language Family\\
        \midrule
        bn & Bengali & B\=a\textipa{N}l\=a & Bengali--Assamese  & Indo--European \\
        cs & Czech & \v{C}e\v{s}tina & Latin & Indo--European \\
        en & English & English & Latin & Indo--European \\
        hu & Hungarian & Magyar nyelv & Latin & Uralic \\
        sr & Serbian & Srpski & Cyrillic & Uralic \\
        sw & Swahili & Kiswahili & Latin & Niger--Congo \\
        te & Telugu & Telugu & Telugu & Dravidian \\
        th & Thai & Phasa Thai & Thai & Kra--Dai \\
        vi & Vietnamese & ti\'{e}ng Vi\^{e}t & Latin & Austroasiatic \\
        
        \bottomrule
    \end{tabular}
    \vspace{-.2cm}
    \label{tab:app:low_langs}
\end{table*}
We focus on all $8$ low-resource languages \cref{tab:app:low_langs} among the $17$ languages supported by \texttt{BenchMAX} \citep{benchmax}. Our language selection is motivated by the observation that the \textsf{Qwen3} series are relatively undertrained in low-resource languages, which makes them a suitable test bed to evaluate the effectiveness of our pipeline in challenging \MT settings. To determine whether a language is categorized as high- or low-resource, we follow the classification logic introduced in \citet{blessing_multilinguality}. The selected $8$ languages span $5$ writing systems and $6$ language families, providing substantial typological and orthographic diversity. Thus we consider this set sufficiently representative for assessing the performance of our pipeline on low-resource multilingualism.

\subsubsection{\MT Training Data--- \textsf{OPUS}}

Our \MT training data are directly sourced from \textsf{Lego-MT} \citep{legomt}, who curated and cleaned large-scale parallel corpora from \textsf{OPUS} \citep{opus}, the largest publicly available collection of translated texts aggregating data from diverse domains including legislative proceedings, subtitles, localization files, and web-crawled content. We use their preprocessed English$\leftrightarrow${target} parallel sentence pairs for each of our 8 low-resource target languages. The resulting training set sizes vary considerably across languages, reflecting the inherent data availability imbalance in the low-resource regime. For the rationale filtration, we apply a quality-based filtration strategy: for each translation direction (i.e., $\texttt{en}\rightarrow\texttt{xx}$ and $\texttt{xx}\rightarrow\texttt{en}$), we retain only the top $20\%$ of parallel pairs ranked by \texttt{spBLEU} \citep{bleu,flores101} scores, ensuring that the curated training signal is of sufficiently high quality for distilled students.

% \subsubsection{\MT \ID\ Benchmark --- \textsf{Flores-101}}
% We evaluate \ID\ \MT performance using \textsf{Flores-101} \citep{flores101}, a many-to-many multilingual translation benchmark covering $101$ languages with \textasciitilde$1$K professionally translated sentences in the \texttt{devtest} split. We report results on both English-to-{target} and {target}-to-English directions. Notably, the source texts of \textsf{Flores-101} are predominantly drawn from English Wikipedia, which is also a major contributing source to several \textsf{OPUS} sub-corpora (e.g., WikiMatrix). This domain overlap justifies treating \textsf{Flores-101} as an \ID\ benchmark, in contrast to the \SID\ multilingual generation tasks in \texttt{BenchMAX} and \OOD tasks.

\subsubsection{\MT \ID\ Benchmark --- \textsf{Flores-101}}
We evaluate \ID\ \MT performance on \textsf{Flores-101} \citep{flores101}, a standardized multilingual translation benchmark covering 101 languages with approximately 1K professionally translated sentences in the \texttt{devtest} split. We report results for both English$\rightarrow${target} and {target}$\rightarrow$English directions.

Although our training data come from \textsf{OPUS}, we use \textsf{Flores-101} for evaluation because it provides a clean, high-quality, and widely adopted test set for multilingual machine translation. We treat it as an \ID\ benchmark in the sense that it matches our training setup at the task level: both involve sentence-level translation in the same language directions, even though the test set is not drawn from the same corpus collection as \textsf{OPUS}.

\subsection{\MT \SID Benchmarks}

\paragraph{\textsf{IFEval}.} To evaluate the models in a shifted-in-domain (\SID) setting, we first assess their capability in \textbf{Rule-based Instruction Following} using the IFEval dataset \citep{ifeval}. \textsf{IFEval} is designed to test whether LLMs can strictly adhere to specific formatting requirements and verifiable constraints (e.g., word count limits and specific output formats). Rather than using the original English dataset, we adopt the high-quality multilingual translated version curated by \texttt{BenchMAX} \citep{benchmax}. This allows us to measure how well instruction-following capabilities transfer to low-resource scenarios. We evaluate the models on the $8$ selected low-resource languages detailed in \cref{tab:app:low_langs} and report the average score across these languages. Following \citet{llama3}, the reported \textsf{IFEval} accuracy is computed as the average of four metrics: \textit{prompt-strict}, \textit{prompt-loose}, \textit{inst-strict}, and \textit{inst-loose} accuracies.

\paragraph{\textsf{GPQA}.} Furthermore, we evaluate the models' proficiency in general \textbf{Science Reasoning} using the \textsf{GPQA} benchmark \citep{gpqa}. \textsf{GPQA} consists of challenging, graduate-level questions spanning various scientific domains such as physics, biology, and chemistry, requiring deep logical reasoning and domain knowledge. Similar to our IFEval setup, we do not evaluate on the original English version of \textsf{GPQA}. Instead, we utilize the translated datasets provided by \texttt{BenchMAX} \citep{benchmax} to investigate whether complex reasoning skills can be effectively elicited in undertrained languages. Consistent with our methodology, the reported results represent the average performance across the same $8$ low-resource languages (\cref{tab:app:low_langs}).

% \subsection{\OOD\ Benchmarks}\label{app:bench_details:ood}
% To evaluate the generalization capabilities of our model beyond the \emph{target} training domain and to verify whether our domain-specific reasoning enhancements translate to general contexts, we assess the model on three highly rigorous out-of-Domain (\OOD) benchmarks covering general science (\sci), mathematics (\mat), and coding (\coding). All experiments in this section are conducted in a \textbf{zero-shot} setting using the \texttt{OpenCompass} framework \citep{opencompass}. The specific metrics and configurations for each benchmark are detailed below.

% \subsubsection{\sci\ -- \textsf{GPQA-Diamond}}
% \textsf{GPQA} \citep[Graduate-Level Google-Proof Q\&A,][]{gpqa} is a highly challenging benchmark comprising expert-crafted questions across physics, biology, chemistry, and other scientific disciplines. In our evaluation, we use the \textsf{GPQA-Diamond} split, which contains the highest-quality questions rigorously vetted by multiple independent domain experts. Given the extreme difficulty of this dataset, we perform 8 independent evaluation runs and report the \textbf{average accuracy} across these $8$ runs as the final performance metric.

\subsection{\OOD\ Benchmarks}\label{app:bench_details:ood}
To evaluate the generalization capabilities of our model beyond the \emph{target} training domain and to verify whether our domain-specific reasoning enhancements translate to general contexts, we assess the model on three highly rigorous out-of-Domain (\OOD) benchmarks covering complex reasoning, mathematics, and coding. All experiments in this section are conducted in a \textbf{zero-shot} setting using the \texttt{OpenCompass} framework \citep{opencompass}. The specific metrics and configurations for each benchmark are detailed below.

\paragraph{Aggregated OOD Score.}
The three \OOD\ domains adopt different official evaluation protocols (harmonic mean for \textsf{BBEH}, average accuracy for \textsf{AIME}, and $\mathrm{Pass@1}$ for \textsf{LCB}) and contain an unequal number of constituent benchmarks (1 for complex reasoning, 2 for mathematics, and 2 for coding). To provide a single, balanced indicator that faithfully reflects cross-domain generalization, we instead report a \textbf{macro-average} across the three \OOD\ domains in \cref{tab:main}, giving each domain equal weight regardless of the number of datasets it contains. Formally, let $S_{\text{reason}}$, $S_{\text{math}}$, and $S_{\text{code}}$ denote the per-domain scores, defined as
\begin{equation}
\label{eq:ood_macro_avg}
\begin{aligned}
s_{\text{reason}} &= \mathrm{HarmonicMean}_{\textsf{BBEH}}, \\
s_{\text{math}}   &= \frac{1}{2}\left( \mathrm{Acc}_{\textsf{AIME2025}} + \mathrm{Acc}_{\textsf{AIME2026}} \right), \\
s_{\text{code}}   &= \frac{1}{2}\left( \mathrm{Pass@1}_{\textsf{LCB-V5}} + \mathrm{Pass@1}_{\textsf{LCB-V6}} \right), \\
s_{\OOD}          &= \frac{1}{3}\left( s_{\text{reason}} + s_{\text{math}} + s_{\text{code}} \right).
\end{aligned}
\end{equation}
The \OOD\ column reported in \cref{tab:main} is $s_{\OOD}$, which is computed according to \cref{eq:ood_macro_avg}, rather than as a flat arithmetic mean over the $5$ individual benchmark scores.

\subsubsection{Complex Reasoning --- \textsf{BIG-Bench Extra Hard}}
\textsf{BIG-Bench Extra Hard} \citep[\textsf{BBEH},][]{bbeh} is explicitly designed to push the frontier of general-purpose reasoning evaluation for frontier LLMs. It is constructed as a successor to \textsf{BIG-Bench Hard} \citep[\textsf{BBH},][]{bbh}, where each of the original \textsf{BBH} tasks is replaced by a substantially more challenging variant that probes a similar underlying reasoning skill but at a markedly increased level of difficulty. The resulting suite spans a diverse spectrum of reasoning competencies, including formal and informal deduction, multi-hop and counterfactual reasoning, long-context understanding, causal analysis, error identification, and the manipulation of complex symbolic, linguistic, and commonsense structures. Owing to the extreme difficulty and broad coverage of \textsf{BBEH}, even the strongest contemporary reasoning models attain only modest scores, making it a particularly stringent testbed for assessing whether reasoning capabilities acquired in our target domain transfer to genuinely out-of-distribution and cognitively demanding scenarios. We report the \emph{(adjusted) harmonic mean}\footnote{To deal with zero values, \citet{bbeh} add a value of $1$ to all accuracy numbers.} of per-task accuracies across all \textsf{BBEH} subtasks as the final metric, following the official evaluation protocol.

\subsubsection{Mathematics --- \textsf{AIME 2025 and 2026}}
The American Invitational Mathematics Examination \citep[\textsf{AIME},][]{aime25,aime26} is a prestigious, highly competitive mathematical Olympiad benchmark that demands exceptionally deep multi-step logical reasoning and advanced problem-solving skills. To rigorously test the model's mathematical reasoning capabilities on the most up-to-date problems, we utilize the latest \textsf{AIME 2025} and \textsf{AIME 2026} test suite. To account for the variance in generating complex mathematical CoT derivations, we evaluate the model across 32 independent runs and report the \emph{average accuracy} over these $32$ runs.

\subsubsection{Coding --- \textsf{LiveCodeBench}}
\textsf{LiveCodeBench} \citep[\textsf{LCB},][]{lcb} is a dynamic, continuously updated evaluation framework designed to rigorously assess the code generation capabilities of LLMs while inherently preventing data contamination. It collects newly published algorithmic problems from competitive programming platforms (e.g., LeetCode, Codeforces, and AtCoder). In our evaluation, we utilize two recent temporal splits to ensure absolute \OOD rigorousness: \textbf{\textsf{V5}} (from August 2024 to February 2025) and \textbf{\textsf{V6}} (from February 2025 to May 2025). The performance on both splits is measured using the standard $\mathrm{Pass@1}$ metric.

% ===== Full-page figure with all 14 task-specific instructions =====
\clearpage
\begin{figure}[p]
\centering
\begin{tcolorbox}[
  width=\linewidth,
  colback=white,
  colframe=gray!40,
  arc=3mm,
  boxrule=1pt,
  left=6pt, right=6pt, top=6pt, bottom=6pt,
]
\scriptsize\sffamily
\begin{description}[leftmargin=14pt, itemsep=2pt, parsep=1pt, topsep=0pt, font=\scriptsize\sffamily\bfseries]
    \item[Forward Synthesis (FS):] You are an expert chemist. Given the SMILES representation of reactants and reagents, your task is to predict the potential product using your chemical reaction knowledge. The input contains both reactants and reagents, and different reactants and reagents are separated by ``\texttt{.}''. Your reply should contain the SMILES representation of the predicted product wrapped in \texttt{<SMILES>} and \texttt{</SMILES>} tags. Your reply must be valid and chemically reasonable.
    \item[Retrosynthesis (RS):] You are an expert chemist. Given the SMILES representation of the product, your task is to predict the potential reactants and reagents using your chemical reaction knowledge. The input contains the SMILES representation of the product. Your reply should contain the SMILES representation of both reactants and reagents, and all reactants and reagents should be enclosed \textbf{together} within a single pair of \texttt{<SMILES>} and \texttt{</SMILES>} tags, separated by ``\texttt{.}''. Your reply must be valid and chemically reasonable.
    \item[Molecule Captioning (MC):] You are an expert chemist. Given the SMILES representation of a molecule, your task is to describe the molecule in natural language. The input contains the SMILES representation of the molecule. Your reply should contain a natural language description of the molecule. Your reply must be valid and chemically reasonable.
    \item[Molecule Generation (MG):] You are an expert chemist. Given the description of a molecule, your task is to generate the potential SMILES representation of the molecule. The input contains the description of the molecule. Your reply should contain the potential SMILES representation of the molecule wrapped in \texttt{<SMILES>} and \texttt{</SMILES>} tags. Your reply must be valid and chemically reasonable.
    \item[Name Conversion (I2F):] You are an expert chemist. Given the IUPAC representation of compounds, your task is to predict the molecular formula of the compound. The input contains the IUPAC representation of the compound. Your reply should contain only the molecular formula of the compound wrapped in \texttt{<MOLFORMULA>} and \texttt{</MOLFORMULA>} tags and no other text. Your reply must be valid and chemically reasonable.
    \item[Name Conversion (I2S):] You are an expert chemist. Given the IUPAC representation of compounds, your task is to predict the SMILES representation of the compound. The input contains the IUPAC representation of the compound. Your reply should contain only the SMILES representation of the compound wrapped in \texttt{<SMILES>} and \texttt{</SMILES>} tags and no other text. Your reply must be valid and chemically reasonable.
    \item[Name Conversion (S2F):] You are an expert chemist. Given the SMILES representation of compounds, your task is to predict the molecular formula of the compound. The input contains the SMILES representation of the compound. Your reply should contain only the molecular formula of the compound wrapped in \texttt{<MOLFORMULA>} and \texttt{</MOLFORMULA>} tags and no other text. Your reply must be valid and chemically reasonable.
    \item[Name Conversion (S2I):] You are an expert chemist. Given the SMILES representation of compounds, your task is to predict the IUPAC representation of the compound. The input contains the SMILES representation of the compound. Your reply should contain only the IUPAC representation of the compound wrapped in \texttt{<IUPAC>} and \texttt{</IUPAC>} tags and no other text. Your reply must be valid and chemically reasonable.
    \item[Property Prediction (ESOL):] You are an expert chemist. Given the SMILES representation of compounds, your task is to predict the log solubility of the compound. The input contains the SMILES representation of the compound. Your reply should contain the log solubility of the compound wrapped in \texttt{\textbackslash boxed\{\}}. Your reply must be valid and chemically reasonable.
    \item[Property Prediction (Lipo):] You are an expert chemist. Given the SMILES representation of compounds, your task is to predict the octanol/water partition coefficient of the compound. The input contains the SMILES representation of the compound. Your reply should contain the octanol/water partition coefficient of the compound wrapped in \texttt{\textbackslash boxed\{\}}. Your reply must be valid and chemically reasonable.
    \item[Property Prediction (BBBP):] You are an expert chemist. Given the smiles representation of the compound, your task is to predict whether blood-brain barrier permeability (BBBP) is a property of the compound. The input contains the compound. Your reply should only contain Yes or No. Your reply must be valid and chemically reasonable.
    \item[Property Prediction (ClinTox):] You are an expert chemist. Given the smiles representation of the compound, your task is to predict whether the compound is toxic. The input contains the compound. Your reply should contain only Yes or No. Your reply must be valid and chemically reasonable.
    \item[Property Prediction (HIV):] You are an expert chemist. Given the smiles representation of the compound, your task is to predict whether the compound serve as an inhibitor of HIV replication. The input contains the compound. Your reply should contain only Yes or No. Your reply must be valid and chemically reasonable.
    \item[Property Prediction (SIDER):] You are an expert chemist. Given the smiles representation of the compound, your task is to predict whether the compound has any side effects. The input contains the compound. Your reply should contain only Yes or No. Your reply must be valid and chemically reasonable.
\end{description}
\end{tcolorbox}
\caption{\textbf{Task-specific instructions for the $14$ subtasks of \textsf{SMolInstruct}.} Each instruction defines the chemical context, specifies the input format, and constrains the output format.}
\label{fig:smol_task_instructions}
\end{figure}
\clearpage

%%%%%%%%%%%%%%%%%%%%%%%%%%%%%%%%%%%%%%%%%%%%%%%%%%%%%%%%%%%%

\newpage
\section*{NeurIPS Paper Checklist}

\begin{enumerate}

\item {\bf Claims}
    \item[] Question: Do the main claims made in the abstract and introduction accurately reflect the paper's contributions and scope?
    \item[] Answer: \answerYes{} % Replace by \answerYes{}, \answerNo{}, or \answerNA{}.
    \item[] Justification: The abstract and \cref{sec:intro} explicitly state our main claims—that QA-only supervision underdetermines trajectory learning, and that optimization strategies (e.g., LST vs. FFT) implicitly resolve this ambiguity to dictate generalization. These claims are strictly scoped and supported by the theoretical formulation in \cref{sec:method} and multi-domain empirical evaluations in \cref{sec:exp,sec:analysis}.


\item {\bf Limitations}
    \item[] Question: Does the paper discuss the limitations of the work performed by the authors?
    \item[] Answer: \answerYes{} % Replace by \answerYes{}, \answerNo{}, or \answerNA{}.
    \item[] Justification: in \appref{app:lim}.

\item {\bf Theory assumptions and proofs}
    \item[] Question: For each theoretical result, does the paper provide the full set of assumptions and a complete (and correct) proof?
    \item[] Answer: \answerYes{} % Replace by \answerYes{}, \answerNo{}, or \answerNA{}.
    \item[] Justification: in \cref{sec:method}.

    \item {\bf Experimental result reproducibility}
    \item[] Question: Does the paper fully disclose all the information needed to reproduce the main experimental results of the paper to the extent that it affects the main claims and/or conclusions of the paper (regardless of whether the code and data are provided or not)?
    \item[] Answer: \answerYes{} % Replace by \answerYes{}, \answerNo{}, or \answerNA{}.
    \item[] Justification: in \cref{sec:exp:setup}, \apprefs{app:exp_details}{app:bench_details}.

\item {\bf Open access to data and code}
    \item[] Question: Does the paper provide open access to the data and code, with sufficient instructions to faithfully reproduce the main experimental results, as described in supplemental material?
    \item[] Answer: \answerYes{} % Replace by \answerYes{}, \answerNo{}, or \answerNA{}.
    \item[] Justification: in \cref{sec:exp:setup}, \apprefs{app:exp_details}{app:bench_details}.

\item {\bf Experimental setting/details}
    \item[] Question: Does the paper specify all the training and test details (e.g., data splits, hyperparameters, how they were chosen, type of optimizer) necessary to understand the results?
    \item[] Answer: \answerYes{} % Replace by \answerYes{}, \answerNo{}, or \answerNA{}.
    \item[] Justification: in \cref{sec:exp:setup,sec:exp:results,sec:analysis}, \apprefs{app:exp_details}{app:bench_details}.

\item {\bf Experiment statistical significance}
    \item[] Question: Does the paper report error bars suitably and correctly defined or other appropriate information about the statistical significance of the experiments?
    \item[] Answer: \answerYes{}{} % Replace by \answerYes{}, \answerNo{}, or \answerNA{}.
    \item[] Justification: We did statistical tests for \cref{sec:exp:results:mono_corr}. Other reported values (basically they are accuracies) intrinsically do not support the calculation of error or statistical significance.

\item {\bf Experiments compute resources}
    \item[] Question: For each experiment, does the paper provide sufficient information on the computer resources (type of compute workers, memory, time of execution) needed to reproduce the experiments?
    \item[] Answer: \answerYes{} % Replace by \answerYes{}, \answerNo{}, or \answerNA{}.
    \item[] Justification: in \apprefs{app:exp_details:training}{app:exp_details:decoding}.
    
\item {\bf Code of ethics}
    \item[] Question: Does the research conducted in the paper conform, in every respect, with the NeurIPS Code of Ethics \url{https://neurips.cc/public/EthicsGuidelines}?
    \item[] Answer: \answerYes{} % Replace by \answerYes{}, \answerNo{}, or \answerNA{}.
    \item[] Justification: Yes, we hereby confirm that we always respect and obey the the NeurIPS Code of Ethics.

\item {\bf Broader impacts}
    \item[] Question: Does the paper discuss both potential positive societal impacts and negative societal impacts of the work performed?
    \item[] Answer: \answerNA{} % Replace by \answerYes{}, \answerNo{}, or \answerNA{}.
    \item[] Justification: This paper focuses on the theoretical analysis and experimental validation of established supervised learning methods. Its contributions are limited to theoretical results, and empirical performance evaluation, without discussing concrete real-world deployment. Therefore, the paper does not directly involve or analyze potential positive or negative societal impacts.
    
\item {\bf Safeguards}
    \item[] Question: Does the paper describe safeguards that have been put in place for responsible release of data or models that have a high risk for misuse (e.g., pre-trained language models, image generators, or scraped datasets)?
    \item[] Answer: \answerNA{} % Replace by \answerYes{}, \answerNo{}, or \answerNA{}.
    \item[] Justification: We claim that our paper poses no such risks.

\item {\bf Licenses for existing assets}
    \item[] Question: Are the creators or original owners of assets (e.g., code, data, models), used in the paper, properly credited and are the license and terms of use explicitly mentioned and properly respected?
    \item[] Answer: \answerYes{} % Replace by \answerYes{}, \answerNo{}, or \answerNA{}.
    \item[] Justification: All datasets, benchmarks, and evaluation toolkits used in this paper are obtained from publicly accessible Hugging Face or GitHub repositories. The original creators or maintainers are properly credited through citations and repository references. These assets are released under licenses such as MIT or CC BY, which permit academic research use without requiring prior approval. We therefore use them in compliance with the applicable licenses and terms of use.

\item {\bf New assets}
    \item[] Question: Are new assets introduced in the paper well documented and is the documentation provided alongside the assets?
    \item[] Answer: \answerNA{} % Replace by \answerYes{}, \answerNo{}, or \answerNA{}.
    \item[] Justification: The paper does not release new assets.

\item {\bf Crowdsourcing and research with human subjects}
    \item[] Question: For crowdsourcing experiments and research with human subjects, does the paper include the full text of instructions given to participants and screenshots, if applicable, as well as details about compensation (if any)? 
    \item[] Answer: \answerNA{} % Replace by \answerYes{}, \answerNo{}, or \answerNA{}.
    \item[] Justification: We claim that the paper does not involve crowdsourcing nor research with human subjects.

\item {\bf Institutional review board (IRB) approvals or equivalent for research with human subjects}
    \item[] Question: Does the paper describe potential risks incurred by study participants, whether such risks were disclosed to the subjects, and whether Institutional Review Board (IRB) approvals (or an equivalent approval/review based on the requirements of your country or institution) were obtained?
    \item[] Answer: \answerNA{} % Replace by \answerYes{}, \answerNo{}, or \answerNA{}.
    \item[] Justification: We claim that the paper does not involve crowdsourcing nor research with human subjects.

\item {\bf Declaration of LLM usage}
    \item[] Question: Does the paper describe the usage of LLMs if it is an important, original, or non-standard component of the core methods in this research? Note that if the LLM is used only for writing, editing, or formatting purposes and does \emph{not} impact the core methodology, scientific rigor, or originality of the research, declaration is not required.
    %this research? 
    \item[] Answer: \answerNA{} % Replace by \answerYes{}, \answerNo{}, or \answerNA{}.
    \item[] Justification: We claim that the core method development in this research does not involve LLMs as any important, original, or non-standard components.

\end{enumerate}

\end{document}